\documentclass[10pt, letter, onecolumn]{arxiv}

\usepackage{kantlipsum, lipsum}
\usepackage{dm-colors}
\usepackage{amsmath}
\usepackage{verbatim}
\usepackage{multirow}
\usepackage{array}
\usepackage{longtable}
\usepackage{pdflscape}
\usepackage{scalerel}
\usepackage{booktabs}
\usepackage{enumitem}
\usepackage{xspace}
\usepackage{bm}
\usepackage{bbm}
\usepackage{mathtools}
\usepackage{soul}
\usepackage{epsfig}
\usepackage{graphicx}
\usepackage{amssymb}
\usepackage{colortbl}
\usepackage{csquotes}
\usepackage{setspace}
\usepackage{colortbl}
\usepackage{bbding}
\usepackage{threeparttable}
\usepackage{tabularx,ragged2e}
\usepackage{placeins}
\usepackage{bbding}
\definecolor{darkpastelgreen}{rgb}{0.13, 0.55, 0.13}
\definecolor{darkpastelred}{rgb}{0.55, 0.13, 0.13}
\definecolor{mygray}{rgb}{1, 1, 1}
\usepackage{lmodern}
\usepackage[hang,flushmargin]{footmisc}
\usepackage{flafter}
\usepackage{nameref}
\usepackage{varioref}
\usepackage{amssymb}
\usepackage{pifont}
\usepackage{rotating}
\usepackage{graphicx}
\usepackage{tcolorbox}
\tcbuselibrary{breakable}
\usepackage[pagebackref=false,breaklinks=false,%
            colorlinks=true,bookmarks=true,citecolor=ourdarkblue,%
            urlcolor=ourdarkblue,linkcolor=ourdarkblue]{hyperref}
\usepackage[noabbrev,capitalize]{cleveref}
\usepackage{etoc}
\usepackage{lineno}
\usepackage{algorithm}
\usepackage{makecell} 
\usepackage{algpseudocode}

\usepackage{amsmath}
\usepackage{thmtools}
\usepackage{lmodern}

\declaretheoremstyle[
    spaceabove=6pt, spacebelow=6pt,
    headfont=\bfseries, headpunct={.}, headformat={\NAME\ \NUMBER},
    bodyfont=\normalfont,
    postheadspace=0.5em
]{promptstyle}

\tcolorboxenvironment{prompt}{
    colback=gray!10!,
    colframe=gray!75!,
    fonttitle=\bfseries,
    title=Prompt,
    boxrule=0.5pt,
    sharp corners
}

\graphicspath{{figures/}}
\definecolor{mygray}{rgb}{0.85, 0.85, 0.85}

\usepackage[normalem]{ulem}

\newcommand{\benchname}{ICUWorld\xspace}
\newcommand{\agentname}{ICUGuardian\xspace}

\title{\Large{A Proactive Longitudinal Clinical Agent for ICU Monitoring}}

\author[1]{Qian Yang}
\author[1]{Pengcheng Qiu}
\author[1]{Qiaoyu Zheng}
\author[2]{Kai Wang}
\author[2]{Shanhua Tang}
\author[3]{Yue Qiu}
\author[1]{Weidi Xie}
\author[1,$\dag$]{\\Ya Zhang}
\author[1,$\dag$]{Chaoyi Wu}

\affil[1]{\normalsize Shanghai Jiao Tong University, Shanghai, China \authorcr \vspace{0.1cm}}
\affil[2]{\normalsize Division of Hepatobiliopancreatic Surgery, Department of General Surgery, Nanfang Hospital, The First School of Clinical Medicine, Southern Medical University, Guangzhou, Guangdong, China \authorcr \vspace{0.1cm}}
\affil[3]{\normalsize The Chinese University of Hong Kong, Hong kong, China \authorcr \vspace{0.1cm}}
\affil[$\dag$]{\normalsize Corresponding authors\authorcr ya\_zhang@sjtu.edu.cn; wtzxxxwcy02@sjtu.edu.cn}

\begin{document}

\begin{abstract}
Intensive care unit~(ICU) monitoring requires continuous integration of evolving patient records, yet conventional alarm systems lack comprehensive contextual explanations and patient-specific adaptability, generating redundant alerts and still burdening clinicians with repeated review. 
We developed \agentname{}, a longitudinal clinical agentic assistant powered by large-language models. It is a multi-agent system with chronological clinical memory that coordinates ICU-specific tools and draws on clinical guidelines to support proactive monitoring and free-text interaction with clinicians.
In time-gated replay evaluation with 317.5 patient-days (7,742 hourly decision points) from MIMIC-IV, eICU-CRD and NWICU, \agentname{} achieved the highest F1 for intervention anticipation and deterioration detection among rule-based, machine-learning and five single-agent LLM baselines.
%
%
Compared with the GPT-5.5 single-agent baseline with basic memory support, Qwen3.5-based \agentname{} increased intervention recall from 54.1\% to 78.7\% with 43.9\% fewer alerts and reduced question-answering hard failures from 16.9\% to 7.6\%, alongside consistent improvements over the same-backbone Qwen3.5 baseline.
In source-blinded manual review, ICU physicians judged 82.6\% of alert times appropriate, typically hours before treatment, and in interactive use rated 93.7\% of answers usable and 97.6\% free of safety concerns.
%
These results support language-model agents with persistent memory as clinician-supervised assistants for ICU monitoring.

\end{abstract}

\maketitle


\section{Introduction}

Effective intensive care unit~(ICU) monitoring is essential for detecting early signs of clinical deterioration~\cite{callcut2021external}.
Although bedside monitors continuously capture physiological measurements, translating these data into timely, targeted interventions remains challenging.
Clinicians must integrate physiological trends with laboratory results, bedside assessments, and the patient's evolving response to treatment, while accounting for relevant clinical guidelines~\cite{fackler2009critical,monnet2025esicm}.
Repeatedly synthesizing this information across extensive longitudinal electronic health records~(EHRs) places substantial demands on clinical attention~\cite{moorman2021principles,lai2022integration}.

Current bedside alarms, severity-of-illness and organ dysfunction scores, and deterioration prediction models support this assessment~\cite{drew2014insights,knaus1985apache,vincent1996sofa,hyland2020early}.
However, these tools struggle to support continuous longitudinal surveillance in two key areas: (i) \textbf{contextual interpretability}: because alarms, scores, and risk estimates operate largely as stateless, isolated evaluations, they cannot interpret new physiological changes against a patient's prior clinical course, evolving treatment responses, and previously issued alerts—nor provide transparent rationales for their clinical significance; and (ii) \textbf{adaptive personalization}: 
predefined inputs and static thresholds cannot tailor monitoring plans to individual patients and clinical tasks, or flexibly incorporate patient-specific evidence as conditions and therapies evolve.
Without maintaining evolving patient states across assessment cycles, existing systems must repeatedly re-evaluate the patient from scratch and frequently trigger redundant alerts for the same unresolved episode, leaving clinicians to manually distinguish genuinely new or worsening deterioration from already recognized issues~\cite{lai2022integration}.

\begin{figure}[!t]
\vspace{-15mm}
\centering
\includegraphics[width=1\linewidth]{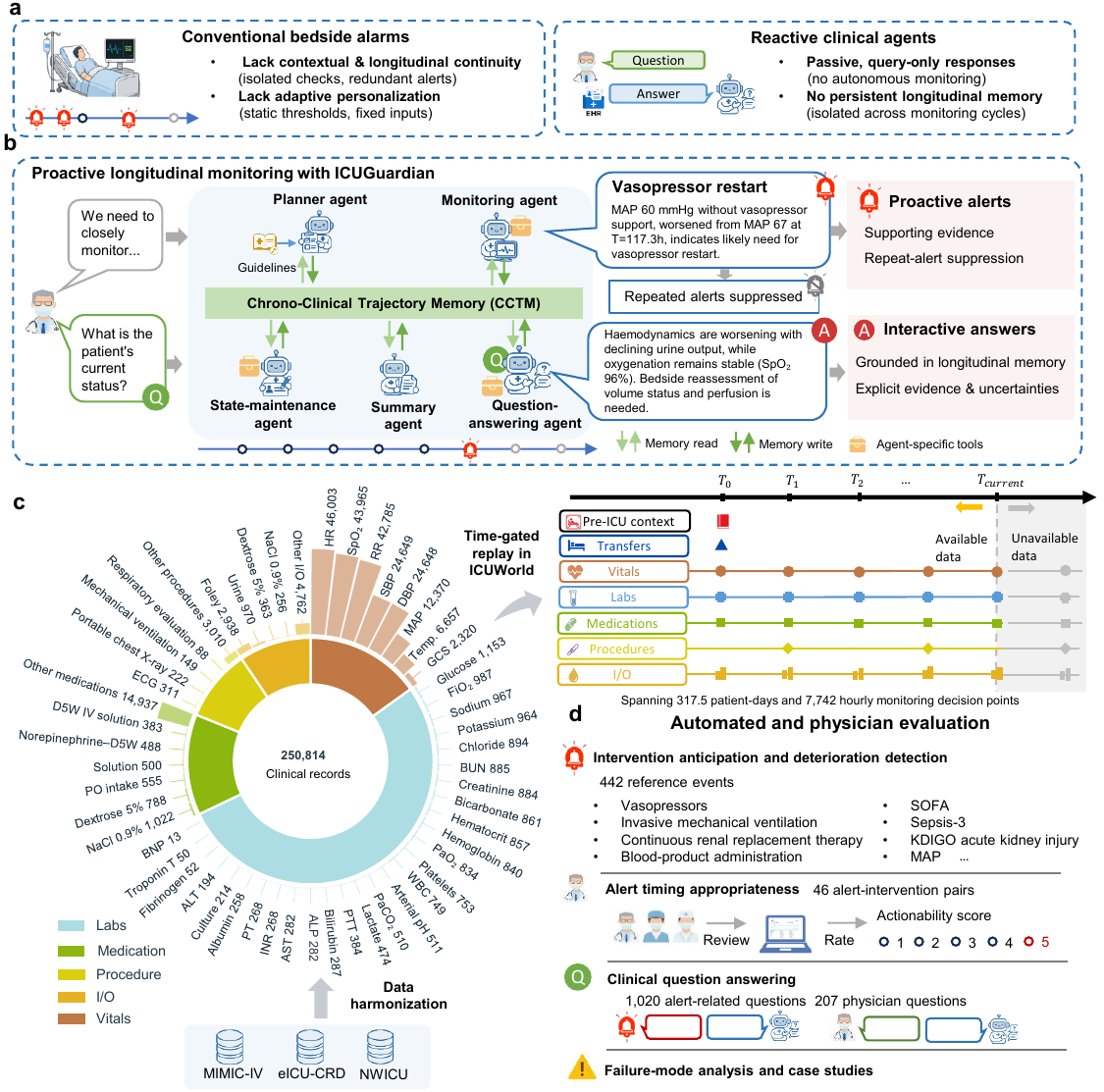}
\caption{\textbf{Overview of \agentname{} and the \benchname{}.}
\textbf{a}, Limitations of conventional bedside alarms (lacking contextual and longitudinal continuity and adaptive personalization) and reactive clinical agents (passive, query-only responses without persistent longitudinal memory).
\textbf{b}, Proactive longitudinal monitoring with \agentname{}. Clinicians assign monitoring tasks and pose clinical questions in natural language; five role-specialized sub-agents (Planner, State-maintenance, Monitoring, Summary and Question-answering agents) coordinate through a shared Chrono-Clinical Trajectory Memory (CCTM) to issue evidence-supported alerts with repeat-alert suppression and memory-grounded interactive answers across the ICU stay.
\textbf{c}, The \benchname{}. Clinical records (250,814 entries) from MIMIC-IV, eICU-CRD and NWICU are harmonized and replayed in a time-gated sandbox spanning 317.5 patient-days and 7,742 hourly monitoring decision points; at each assessment, agents receive a Clinical Snapshot summarized at the current time and can access only records available up to that time.
\textbf{d}, Automated and physician evaluation: intervention anticipation and deterioration detection across 442 reference events, source-blinded physician review of alert timing across 46 alert--intervention pairs, clinical question answering across 1,020 automated and 207 physician questions, and failure-mode analysis and case studies.
ICU, intensive care unit.}
\label{fig:overview}
\end{figure}

Recent advances in large language model~(LLM)-based clinical agents hold promise for overcoming these limitations~\cite{qiu2024agentic,freyer2026regulation}.
By combining LLM reasoning with external clinical tools, these agents can address diverse clinical needs expressed in natural language~\cite{shi2024ehragent,ferber2026mira}, iteratively explore patient records through EHR systems~\cite{ossowski2026codeclinic,shen2026realicu}, and provide transparent, traceable rationales grounded in guidelines, risk scores, and abnormal measurements to help clinicians assess the clinical appropriateness of their recommendations~\cite{zhao2026deeprare}.
\textbf{However, existing clinical agents are primarily designed to respond reactively to individual queries, whereas ICU care requires continuous, proactive monitoring}, with agents autonomously initiating repeated assessments without repeated clinician prompting, and making interdependent decisions throughout prolonged patient stays~\cite{fackler2009critical,lin2026longitudinal}.
%
Persistent agentic memory is therefore essential to track evolving patient states, preserve monitoring plans and retain prior decisions to maintain coherence across monitoring cycles, enabling sustained surveillance and collaboration with ICU clinicians~\cite{lin2026longitudinal,shen2026realicu}.
Achieving this reliably over long, information-dense stays requires coupling a deterministic monitoring workflow---which enforces consistent assessment steps and validated, structured memory updates---with language-model reasoning to interpret each patient's evolving clinical course. 
%
%
%

To address these gaps, we present \agentname{}, a proactive longitudinal clinical agent assistant for ICU monitoring~(Fig.~\ref{fig:overview}a).
%
Clinicians can assign ongoing monitoring tasks and priorities in natural language, and the system autonomously performs repeated assessments, issues proactive alerts and responds to clinical questions throughout the long-horizon stay.
%
\agentname{} draws on 56 ICU clinical tools and 12 clinical guidelines for retrieving patient data and analyzing clinical changes, organ function, medication exposure, treatment support and clinical episodes.
%
To process densely observed trajectories without context overload, \agentname{} coordinates five role-specialized sub-agents---a Planner, a State-maintenance agent, a Monitoring agent, a Question-answering agent and a Summary agent---operating within isolated, task-specific contexts~(Fig.~\ref{fig:overview}b).
%
These sub-agents maintain cross-cycle coherence through a shared \textit{Chrono-Clinical Trajectory Memory}~(CCTM).
CCTM links new clinical findings to each patient's prior disease course, active monitoring plans, and therapeutic support statuses, working in concert with episode-aware alert management to detect meaningful changes while suppressing repeated notifications for ongoing issues.
%

We evaluated \agentname{} in the \benchname{}~(Fig.~\ref{fig:overview}c) simulated sandbox, which replays over 250,000 time-stamped clinical measurements and treatment records spanning 317.5 patient-days and 7,742 hourly monitoring decision points across MIMIC-IV, eICU-CRD and NWICU~\cite{johnson2023mimic,pollard2018eicu,moukheiber2024nwicu}, restricting access to records available at each simulated timestamp.
We benchmarked \agentname{} against rule-based, machine-learning and five single-agent LLM baselines on anticipating clinical interventions and detecting deterioration across 442 reference events, alongside an automated clinical question-answering~(QA) benchmark of 1,020 queries and two clinician evaluations.
%
In continuous monitoring, compared with a standard single-agent baseline sharing the same Qwen backbone, \agentname{} substantially improved the detection of critical changes while controlling notification burden, increasing intervention recall from 41.0\% to 78.7\% while reducing total alerts by 69.1\% (and consistently outperforming conventional rule and machine-learning baselines).
%
Against this same Qwen baseline in paired QA evaluations, \agentname{} also lowered the hard-failure rate---defined as responses containing critical factual errors, major clinical omissions or unsafe reasoning---from 19.7\% to 7.0\%.
%
In a source-blinded timing review by three ICU physicians, \agentname{} alerts preceded documented clinical interventions by a median of 4.84\,h, with 82.6\% of alert timings judged clinically appropriate, suggesting potential to support earlier intervention~(Fig.~\ref{fig:clinician_timing}).
%
%
In a forward-only physician--AI interaction study, the same physicians freely queried \agentname{} at self-selected time points during trajectory replay, rating 93.7\% of 207 responses clinically usable and flagging no potential safety concerns in 97.6\%, supporting its potential for collaborative clinical decision support.

\section{Results}

\subsection{System and Study Overview}

We developed \agentname{}, an LLM-based multi-agent system that supports clinicians in monitoring ICU patients throughout their stay.
Clinicians assign monitoring tasks in natural language; \agentname{} then reassesses the patient at hourly intervals and proactively issues alerts when it identifies changes that may require clinical attention, including an emerging need for clinical intervention and clinical deterioration.
At any time, clinicians can also ask \agentname{} about the patient, for example to review the current status, examine the evidence behind an alert or clarify clinical priorities and next steps.
%
%

\agentname{} coordinates five role-specialized sub-agents in isolated contexts---a Planner that derives guideline-grounded monitoring plans, a State-maintenance agent that updates patient trajectories, a Monitoring agent that proposes alerts, a Question-answering agent that responds to clinical queries and a Summary agent that compiles periodic 12-h course summaries---equipped with role-specific subsets of 56 ICU tools and 12 clinical guidelines.
These sub-agents share a \textit{Chrono-Clinical Trajectory Memory}~(CCTM), which relates each new finding to the patient's prior clinical course, monitoring plans, treatment support and previous alerts; a deterministic, episode-aware alert manager uses this memory to withhold repeated alerts for an ongoing problem while allowing a new alert when the problem worsens or recurs (\hyperref[methods:icuagent]{Methods, Section~\ref*{methods:icuagent}}).

We evaluated \agentname{} in the \benchname{}, which replays more than 250,000 time-stamped clinical records spanning 317.5 patient-days of longitudinal ICU care across MIMIC-IV, eICU-CRD and NWICU, exposing only records available up to each simulated hour and yielding 7,742 hourly alert-decision points.
%
%
We first evaluated \agentname{} on anticipating clinical interventions and detecting clinical deterioration across 442 reference events, comparing it with five single-agent LLM baselines and with rule-based and machine-learning methods.
The primary comparator, the single-agent baseline using Qwen3.5 (hereafter the Qwen baseline), shared the backbone, general-purpose tools and record access of \agentname{}.
Three ICU physicians (9, 7 and 4 years of clinical experience) then assessed the timing of selected alerts in a source-blinded review, and we examined the physiological states in which alerts were issued.
%
%
We next evaluated clinical question answering on an automated benchmark of 1,020 questions scored against a prespecified rubric.
In a subsequent physician--AI interaction study, the same physicians posed 207 questions of their own while reviewing evolving records in the \benchname{} and rated the answers for clinical usability and safety concerns.
Finally, we examined individual cases qualitatively and characterized the errors that remained with \agentname{}.
Cohort and task composition are summarized in Supplementary Note~\ref{app:cohort_details} and Supplementary Table~\ref{tab:cohort}.

\subsection{Automated Evaluation of Alerts against Reference Events}
\label{sec:retrospective_replay_results}

We evaluated alerts generated by different systems during replay against 442 reference events across two task families: anticipating the need for clinical intervention~(intervention anticipation) and detecting multivariable or single-variable deterioration~(deterioration detection).
At the start of each replay, agents were instructed in natural language to monitor every stay and raise alerts across all target interventions and deterioration criteria at hourly steps, without prior knowledge of subsequent events.
For \textbf{intervention anticipation}, the 61 reference events were new starts of vasopressors, invasive mechanical ventilation, continuous renal replacement therapy or blood-product administration, matched by target alerts issued between the first evidence of need and treatment initiation.
%
%
For \textbf{deterioration detection}, the 381 reference events comprised 175 multivariable organ-dysfunction events in MIMIC-IV (SOFA~\cite{vincent1996sofa}, Sepsis-3~\cite{singer2016sepsis3} and KDIGO acute kidney injury~\cite{kdigo2012aki}) and 206 single-variable vital-sign or laboratory threshold crossings across all three cohorts, matched within 2\,h of onset.
Event definitions, episode- and alert-level matching rules, and evaluation metrics (recall, precision and F1) are detailed in \hyperref[methods:continuous_monitoring_benchmark]{Methods, Section~\ref*{methods:continuous_monitoring_benchmark}}.

To isolate the contribution of our agentic design, we compared \agentname{} (using Qwen3.5-397B-A17B by default) against five single-agent LLM baselines---GPT-5.5, Claude Sonnet~4.6, Gemini~3.1-Pro-Preview, DeepSeek-V4-Pro and Qwen3.5-397B-A17B.
At each hourly step, each single-agent baseline received the same Clinical Snapshot, guideline references and 51 general-purpose ICU tools (including basic cross-step note storage; Supplementary Tables~\ref{tab:app:toolset_retrieval}--\ref{tab:app:toolset_state})
, but operated without CCTM, the multi-agent workflow or episode-aware alert management; the Qwen baseline thus provided a direct, backbone-matched comparison with \agentname{}.
Alongside these general-purpose agents, we included task-specific conventional comparators: for intervention anticipation, deterministic guideline-informed rules~\cite{prescott2026surviving,helms2024oxygen,kdigo2012aki,carson2023red} and four supervised machine-learning models (LightGBM~\cite{hyland2020early}, LSTM and CNN~\cite{suresh2017clinical}, and random forest~\cite{wang2020mimicextract}) trained on an independent cohort of 14,785 MIMIC-IV stays to predict 6-h intervention need; and for deterioration detection, mechanical threshold rules reflecting standard bedside alarms, evaluated with and without a 12-h repeat-alert cooldown (details in \hyperref[methods:experimental_settings]{Methods, Section~\ref*{methods:experimental_settings}}).

\begin{figure}[!htb]
\centering
\includegraphics[width=\linewidth,trim=0 20pt 0 0,clip]{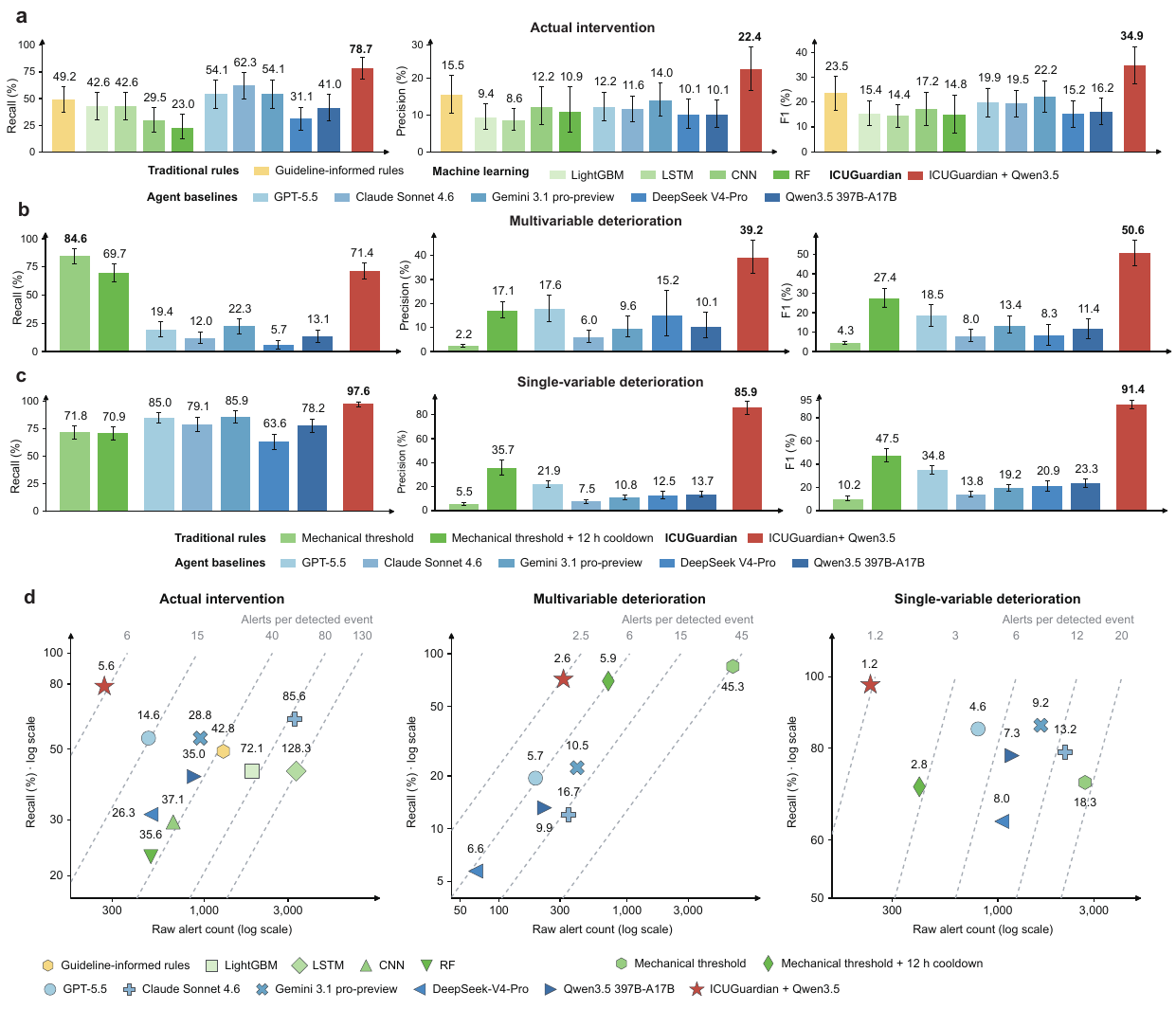}
\caption{\textbf{Event detection and alert burden across monitoring tasks.}
\textbf{a--c}, Recall, precision and F1 for intervention anticipation (\textbf{a}; pooled across the three cohorts), multivariable deterioration (\textbf{b}; MIMIC-IV) and single-variable deterioration (\textbf{c}; pooled). Precision and F1 are computed on prediction episodes in \textbf{a} and on individual alerts in \textbf{b,c}. Bold marks the highest value for each metric; error bars show 95\% confidence intervals from 20,000 ICU-stay bootstrap resamples (stratified by dataset when pooled).
\textbf{d}, Recall versus raw alert count (log scales); the upper left indicates more events captured with fewer alerts. Point labels and dashed guide lines give raw alerts per detected event, which equals 1/precision for deterioration but not for intervention anticipation, where precision is episode-based.
}
\label{fig:main_performance}
\end{figure}
\paragraph{Comparator methods struggled to combine high event capture with low alert burden across tasks.}
\label{sec:decomp}

\begin{figure}[!t]
\centering
\includegraphics[width=\linewidth]{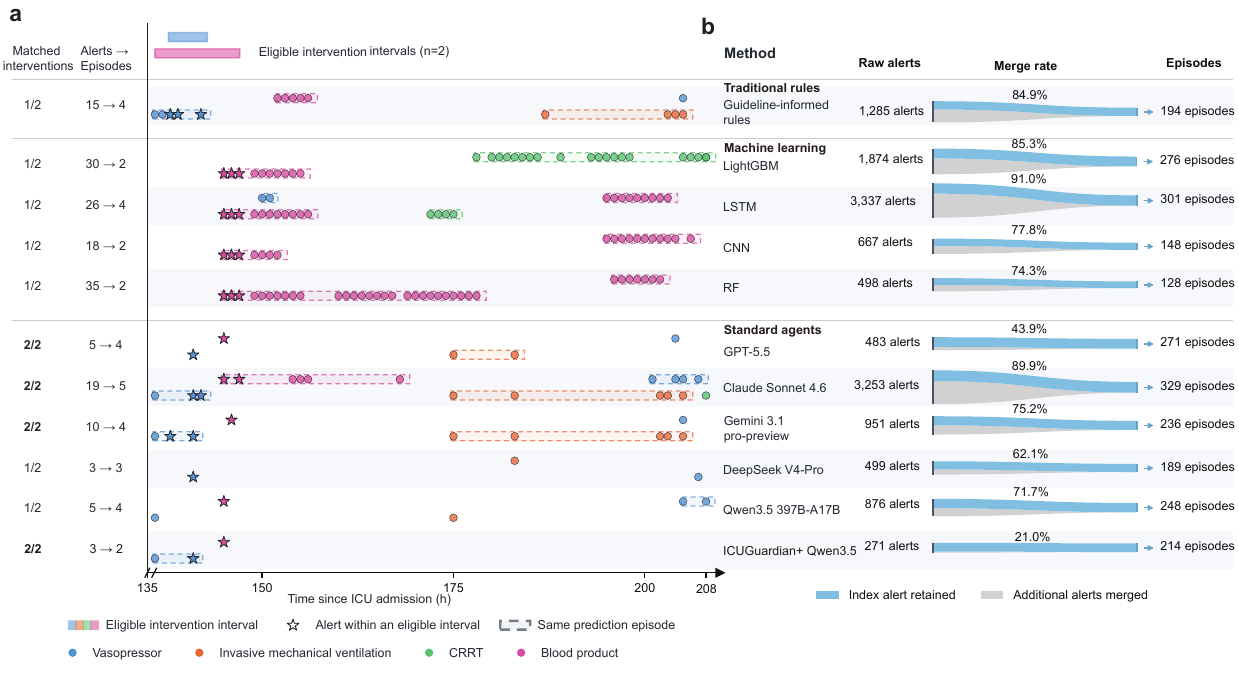}
\caption{\textbf{Repeated intervention alerting and episode-level aggregation.}
\textbf{a}, Intervention alerts in an illustrative MIMIC-IV stay with two eligible intervention intervals. Stars, alerts within eligible intervals; dashed boxes, prediction episodes; left columns, matched interventions and alert-to-episode counts. Rows align with \textbf{b}.
\textbf{b}, Aggregation of raw intervention alerts into prediction episodes across all 100 test stays. Merge rate is the proportion of alerts merged into an existing episode; band thickness uses a square-root count scale shared across methods.}
\label{fig:alert_burden}
\end{figure}

Conventional methods could not increase event capture without a steep rise in alerts.
For multivariable deterioration, the mechanical threshold rule detected 84.6\% of events but issued 45.3 alerts per detected event; a 12-h cooldown reduced this to 5.9, but recall fell to 69.7\% (Fig.~\ref{fig:main_performance}b,d).
For intervention anticipation, no rule-based or machine-learning method detected half of the events (23.0--49.2\%), and the two machine-learning models with the highest recall, LightGBM and LSTM, issued 72.1 and 128.3 alerts per detected event (Fig.~\ref{fig:main_performance}a,d).
Single-agent LLM baselines improved on this balance in two tasks.
For intervention anticipation, three of the five detected more events than any conventional method (54.1--62.3\% versus at most 49.2\%; Fig.~\ref{fig:main_performance}a), and four issued fewer alerts per detected event (14.6--35.0 versus 35.6--128.3; Fig.~\ref{fig:main_performance}d), suggesting that weighing the clinical context helps to anticipate treatment decisions.
For single-variable deterioration, four of the five detected more events than the threshold rule (78.2--85.9\% versus 71.8\%; Fig.~\ref{fig:main_performance}c).
At the hourly assessments shared by all methods, the rule evaluated only the most recent readings, whereas the agents also took account of abnormal values recorded earlier in the preceding hours.
In contrast, single-agent LLM baselines detected only 5.7--22.3\% of multivariable deterioration events (Fig.~\ref{fig:main_performance}b), although calculators for SOFA, Sepsis-3 and KDIGO stages were among their tools.
These criteria combine several organ systems and compare current with earlier values, whereas single-agent LLM baselines were not given a maintained record of the patient's course and, unless they chose to keep notes, had to re-establish it at every step.
They also repeated alerts for ongoing problems: 43.9--89.9\% of their intervention alerts were merged into an existing prediction episode (Fig.~\ref{fig:alert_burden}b), and for single-variable deterioration they issued 4.6--13.2 alerts per detected event, compared with 2.8 for the threshold rule with a cooldown (Fig.~\ref{fig:main_performance}d).
Together, these results indicate that equipping LLM agents with clinical tools improves context-dependent detection over conventional methods, whereas tracking cumulative multi-organ deterioration and suppressing repeated alerts across prolonged stays require maintained longitudinal memory and episode-level alert control.
%

\FloatBarrier

\paragraph{By following patients over time, \agentname{} improved detection and reduced alert burden.}

%
%
%
By maintaining the patient state in CCTM and applying episode-aware alert management, \agentname{} substantially improved multivariable deterioration detection while curbing repeated alerts.
It detected 71.4\% of these events, compared with 13.1\% for the Qwen baseline, a recall similar to that of the threshold rule with a cooldown (Fig.~\ref{fig:main_performance}b); although it issued more alerts than that baseline (319 versus 227; Supplementary Table~\ref{tab:app:pooled_performance}), it needed 2.6 alerts per detected event, fewer than any other method (Fig.~\ref{fig:main_performance}d).
Repeated alerts were also less frequent: in an illustrative MIMIC-IV stay, \agentname{} matched both eligible interventions with three alerts, whereas the single-agent LLM baselines that also matched both issued 5--19 alerts (Fig.~\ref{fig:alert_burden}a), and across the cohort only 21.0\% of its intervention alerts were merged into an existing prediction episode, the lowest of any method (Fig.~\ref{fig:alert_burden}b).
In the two tasks in which single-agent LLM baselines had improved on conventional methods, \agentname{} achieved the highest recall while issuing the fewest alerts: it detected 78.7\% of intervention events with 271 alerts, compared with 41.0\% with 876 alerts for the Qwen baseline (69.1\% fewer alerts; Figs.~\ref{fig:main_performance}a and~\ref{fig:alert_burden}b), and 97.6\% of single-variable deterioration events with 1.2 alerts per detected event, compared with 78.2\% and 7.3 (Fig.~\ref{fig:main_performance}c,d).
As a result, \agentname{} had the highest precision, the highest F1 and the fewest alerts per detected event in every task (Fig.~\ref{fig:main_performance}a--d); its F1 exceeded that of the Qwen baseline by 18.7--68.1 percentage points, and F1 differences favoured \agentname{} over every comparator (all Holm-adjusted $P<0.02$; Supplementary Table~\ref{tab:app:paired_differences}).
%
%
\agentname{} likewise achieved the highest F1 in each of the three datasets and for nearly every monitoring target, and these gains were reproduced in MIMIC-IV when GPT-5.5 replaced Qwen3.5 (Supplementary Fig.~\ref{fig:app:database_detection}).
Component ablations on MIMIC-IV further confirmed that removing planning, state maintenance or deterministic episode-aware alert governance substantially degraded F1 across all three tasks (Supplementary Table~\ref{tab:e4_component_ablation_qwen}).

\begin{figure}[!ht]
\centering
\includegraphics[width=1\linewidth]{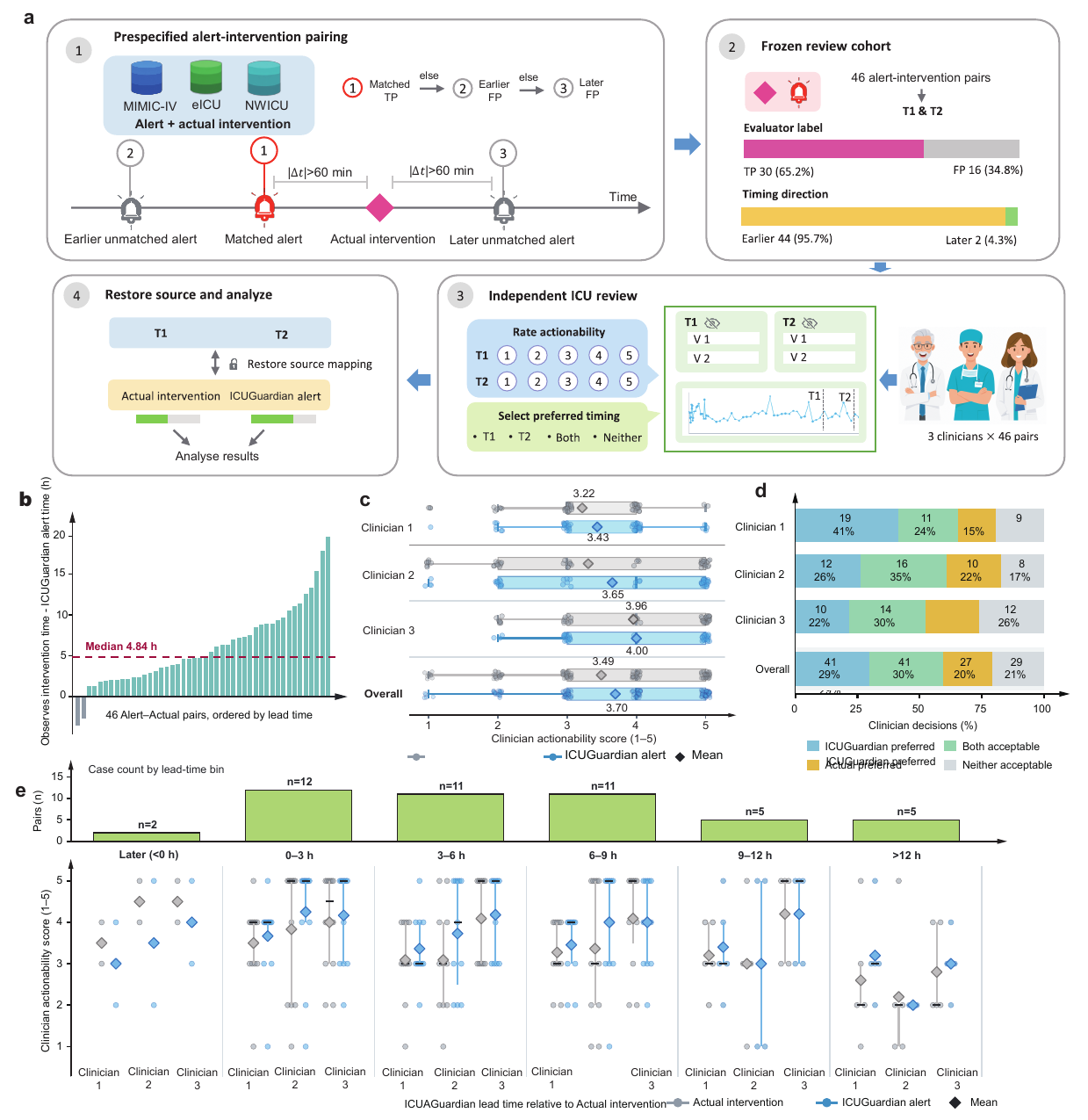}
\caption{\textbf{Source-blinded clinician evaluation of \agentname{} alert
timing.}
\textbf{a}, Selection and blinded review of candidate warning times. Three ICU physicians independently reviewed 46 alert--intervention pairs using time-gated records; the two candidate times in each pair (${>}60$\,min apart) were shown chronologically as T1 and T2, with sources concealed. Pair selection is described in \hyperref[methods:clinician_timing]{Methods, Section~\ref*{methods:clinician_timing}}.
\textbf{b}, Alert timing relative to observed interventions. Each bar
represents one selected pair. Signed lead time is observed intervention time
minus alert time; positive values indicate earlier alerts, and the dashed
line marks the median of 4.84\,h.
\textbf{c}, Clinician-rated actionability of alert and observed-intervention times, from 1 (insufficient support) to 5 (highly appropriate), with 3 indicating acceptable; $n=138$ paired assessments (46 per clinician). Dots, individual ratings; boxes, interquartile range (IQR); centre lines, medians; whiskers, most extreme values within $1.5\times\text{IQR}$; diamonds, means.
%
%
\textbf{d}, Clinician preferences between candidate warning times; bars show counts and rounded percentages (which may not sum to 100\%).
\textbf{e}, Actionability across exploratory lead-time strata ($n=2$, 12, 11, 11, 5 and 5 pairs). Vertical segments, IQRs; black horizontal ticks, medians; dots and diamonds as in \textbf{c} (IQRs and medians omitted for $n=2$).
}
\label{fig:clinician_timing}
\end{figure}

\subsection{Clinical Assessment of Detailed Alert Timing}

\label{sec:hil_results}
\label{sec:clinician_timing_results}
Automated matching of alerts to corresponding events documented later in the medical record offers a scalable way to estimate how often binary event alerts capture documented events. 
However, this reference-based approach provides only a coarse assessment and cannot fully evaluate whether alerts were issued at clinically appropriate times. 
Alerts that fall outside the reference intervals may still be clinically reasonable, while alerts that match the intervals may vary in the clinical appropriateness of their timing.
%
%
We therefore asked three ICU physicians to independently review 46 \agentname{} alerts, including alerts matched and unmatched in the automated evaluation, paired with the recorded start of the corresponding treatment. Each alert received three reviews, for 138 assessments in total (\hyperref[methods:clinician_timing]{Methods, Section~\ref*{methods:clinician_timing}}).
Viewing the records available up to each time, with treatment initiation masked and the source of each time concealed, they rated the appropriateness of a warning at both times on a five-point scale, with 3 or higher considered acceptable, and chose the better time, both or neither (Fig.~\ref{fig:clinician_timing}a).

\begin{figure}[!t]
\centering
\includegraphics[width=1\linewidth]{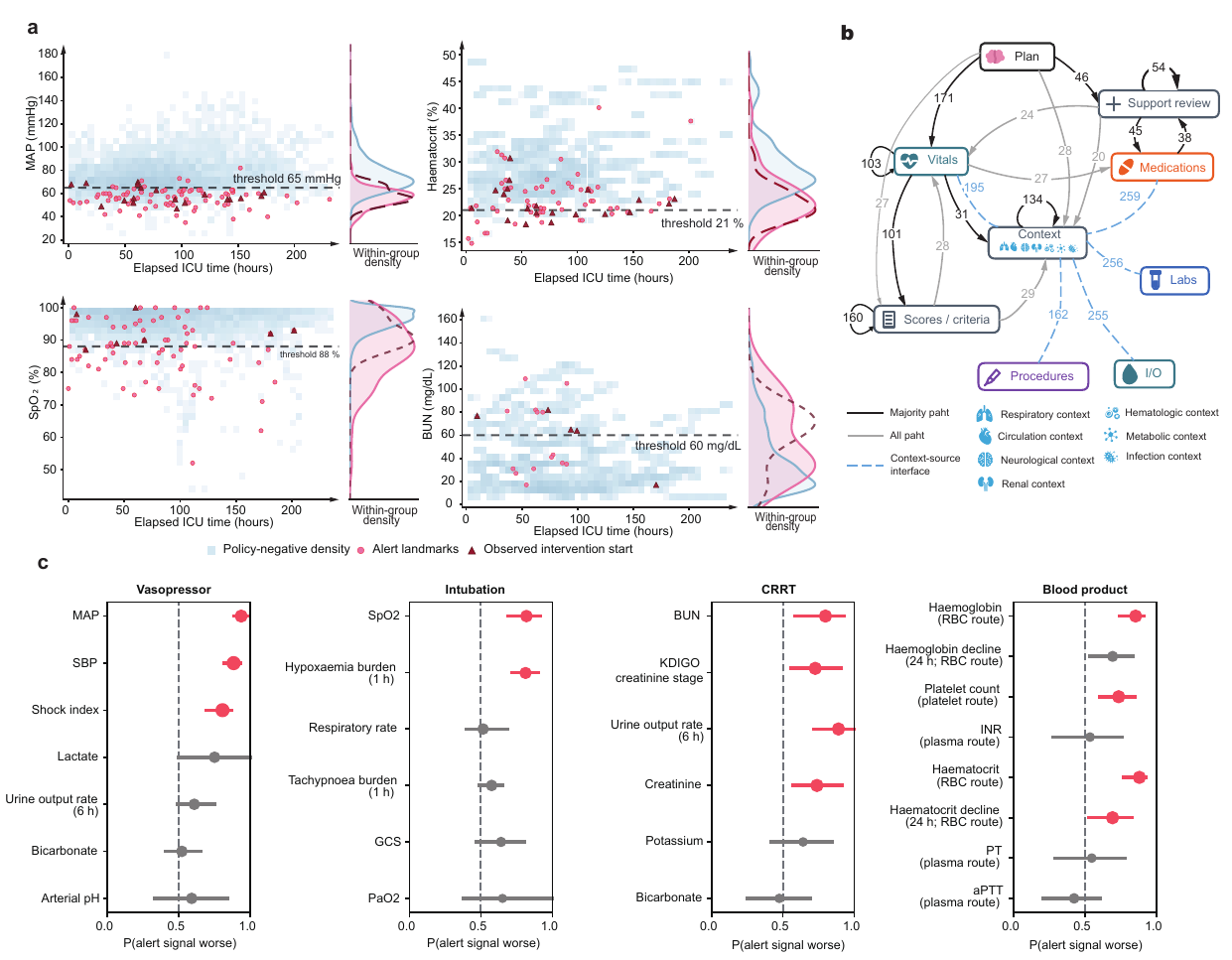}
\caption{\textbf{Physiological context and recorded information retrieval
surrounding \agentname{} alerts.}
\textbf{a}, MAP, haematocrit, SpO$_2$ and BUN across elapsed ICU time at non-alert landmarks (blue shading), target-specific alerts (pink circles) and observed intervention starts (triangles); right-hand densities are normalized within each group, and dashed lines mark prespecified thresholds. $n$ (alert/non-alert/intervention) = 81/3,336/19 (MAP), 57/3,083/20
(haematocrit), 72/4,442/7 (SpO$_2$) and 14/3,236/5 (BUN).
\textbf{b}, Transitions between information categories in tool calls at 295 intervention-candidate times from 39 MIMIC-IV stays (1,270 calls; 975 transitions). Edge labels are counts; self-loops are consecutive calls within a
category, links with fewer than 20 transitions are omitted, and dashed links
map Context calls to data sources.
\textbf{c}, Probability that the alert-time value was physiologically worse than
at matched non-alert time points (ties counted as 0.5) for the 27 of 34
prespecified signals that were estimable. Lines show pointwise 95\% confidence
intervals from 500 stay-level bootstrap resamples; pink indicates
Benjamini--Hochberg $q<0.05$ across all 34 signals
(Supplementary Note~\ref{app:alert_temporal_validity}).
MAP, mean arterial pressure; SpO$_2$, peripheral oxygen saturation; BUN, blood
urea nitrogen.}
\label{fig:e6_temporal_validity}
\end{figure}


\paragraph{Clinicians judged most reviewed alerts appropriate, especially those issued up to 6\,h before treatment.}

%
Most reviewed alerts preceded treatment (44 of 46; Fig.~\ref{fig:clinician_timing}a), with a median interval between the alert and the recorded start of treatment, referred to here as the lead time, of 4.84\,h (interquartile range, 2.48--8.86\,h; Fig.~\ref{fig:clinician_timing}b).
Despite this earlier timing, clinicians judged 82.6\% of alert times appropriate (a score of 3 or higher; 114 of 138 assessments), and the mean rating of alert times was slightly higher than that of the recorded treatment starts for each of the three physicians (overall, 3.70 versus 3.49; Fig.~\ref{fig:clinician_timing}c).
%
%
In 82 of 138 assessments (59\%), clinicians preferred the alert time (41, 30\%) or judged both times acceptable (41, 30\%), whereas they preferred the recorded start in 27 (20\%; Fig.~\ref{fig:clinician_timing}d).
Because these alerts came hours before treatment was started, these findings suggest that introducing \agentname{} into the clinical workflow could prompt clinicians to initiate appropriate interventions earlier.
In an exploratory analysis, we then examined how far ahead of treatment alerts remained acceptable (Fig.~\ref{fig:clinician_timing}e and Supplementary Table~\ref{tab:app:hil_posthoc}).
Alerts issued up to 6\,h before treatment received the highest ratings, with all three physicians rating the alert time above the recorded start (mean, 3.90 versus 3.61); between 6 and 12\,h, mean ratings of alert times remained above the acceptability threshold (3.82 for 6--9\,h and 3.53 for 9--12\,h), whereas beyond 12\,h they fell below it (2.73).
Taken together, alerts issued up to 12\,h before treatment were generally judged appropriate, and those within 6\,h most consistently so, suggesting the window in which \agentname{} could support earlier intervention.

\paragraph{Alerts were issued in physiological states resembling those at the recorded start of the corresponding treatment.}

\label{sec:e6_temporal_validity}
%
%
Beyond direct clinician review, we also examined whether patients’ physiology at alert times resembled that at the recorded start of the corresponding treatment.
If an alert reflects a genuine clinical need, the patient's physiology when the alert is issued should resemble the physiology at the recorded start of the corresponding treatment, for example, a low mean arterial pressure before vasopressor treatment begins.
Extending the assessment beyond the 46 alerts reviewed by clinicians, we therefore compared physiological states at all 271 intervention alerts with those at real treatment start times.
Four signals were selected to illustrate the clinical states associated with the corresponding intervention targets:
mean arterial pressure for vasopressors, haematocrit for blood-product administration, oxygen saturation for invasive mechanical ventilation and blood urea nitrogen for continuous renal replacement therapy.
Their distributions were compared across alert times, non-alert time points and observed intervention starts.
Visual comparison showed overlap between the physiological distributions at alert times and observed intervention starts. 
Relative to non-alert time points, alert-time distributions generally shifted towards abnormal values relevant to the corresponding intervention targets (Fig.~\ref{fig:e6_temporal_validity}a).
This descriptive similarity placed alerts within physiological states resembling those in which the corresponding interventions were initiated.

To quantify these associations, we compared alert and matched non-alert time points across the 27 of 34 prespecified physiological signals with sufficient matched observations (Supplementary Note~\ref{app:alert_temporal_validity}).
Thirteen comparisons showed significant differences after
Benjamini--Hochberg correction ($q<0.05$), including lower mean arterial
pressure for vasopressor alerts, lower oxygen saturation for invasive
mechanical ventilation alerts, higher blood urea nitrogen and creatinine for continuous renal
replacement therapy alerts, and lower haemoglobin and haematocrit
for blood-product alerts (Fig.~\ref{fig:e6_temporal_validity}c).
These findings link alerts to physiological abnormalities relevant
to their respective intervention targets.
Alerts that were never followed by treatment are examined individually in the failure-mode analysis below.


\begin{figure}[!ht]
\centering
\includegraphics[width=1\linewidth]{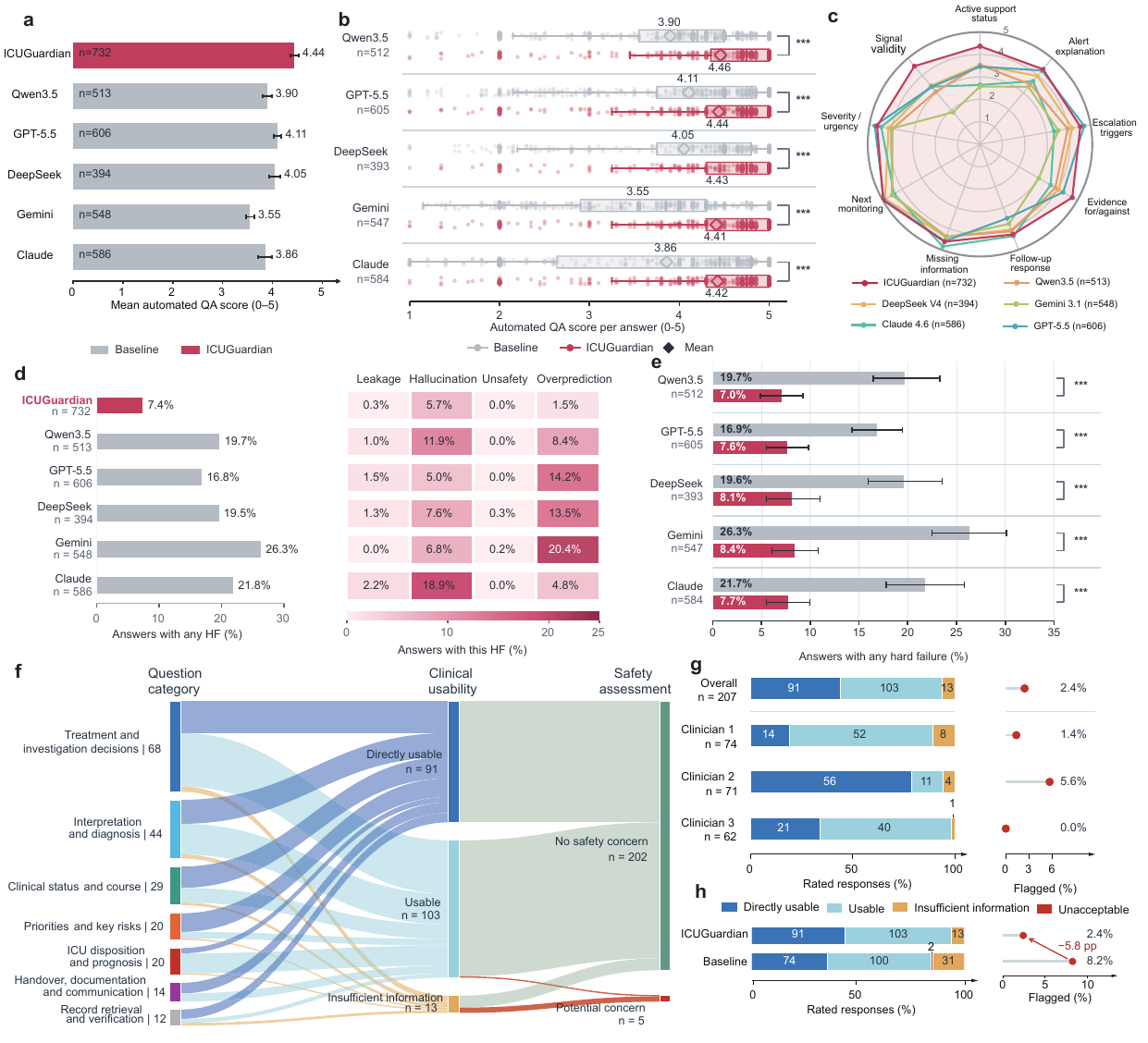}
\caption{\textbf{Automated question answering and clinician interaction with
\agentname{}.}
\textbf{a}, Mean automated rubric scores (1--5; higher is better) for
\agentname{} and five single-agent LLM baselines on all eligible answers from
the 1,020-question benchmark ($n$, scored answers). Error bars show 95\%
percentile confidence intervals from ICU-stay bootstrap resamples.
\textbf{b}, Paired scores on jointly eligible questions ($n$, paired questions;
matched sets differ between comparators). Boxes show the interquartile range
(IQR), centre lines the median, whiskers the most extreme values within
1.5 times the IQR, diamonds the mean and points individual scores.
\textbf{c}, Mean scores by question type, using the answers in \textbf{a}.
\textbf{d}, Rates of any hard failure and of four grouped failure categories,
which may overlap within an answer, using the answers in \textbf{a}.
\textbf{e}, Paired hard-failure rates on the matched sets in \textbf{b};
confidence intervals are from paired ICU-stay bootstrap resamples stratified by
dataset.
In \textbf{b} and \textbf{e}, two-sided paired label-swap tests are clustered
by ICU stay and Holm-corrected across the five comparators for each outcome;
$^{***}P_{\mathrm{adj}}<0.001$.
\textbf{f}, Question categories, clinical usability and safety concerns for 207
\agentname{} responses to questions posed by three ICU physicians at
self-selected time points; ribbon widths represent response counts, and safety
flags denote potentially important errors or omissions.
\textbf{g}, Usability distributions (stacked bars; percentages with counts
labelled) and safety-flagging rates (points), overall and by physician ($n$,
rated responses).
\textbf{h}, Paired comparison of \agentname{} with a single-agent LLM baseline
using the same backbone on the 207 questions, with anonymized answers rated
separately; conventions as in \textbf{g}. The arrow shows the difference in
safety-flagging rates (\agentname{} minus baseline); pp, percentage points.}

\label{fig:qa_evaluation}
\end{figure}

\subsection{Evaluation of Clinical Question Answering}

We next evaluated whether \agentname{} could answer clinicians' questions about its alerts and the patient's condition with appropriate rationales, first on an automated benchmark that compared it with all five single-agent LLM baselines under standardized conditions (\hyperref[methods:clinical_qa]{Methods, Section~\ref*{methods:clinical_qa}}) and then in a real physician--AI interaction study in which physicians asked their own questions (\hyperref[methods:qa_interaction]{Methods, Section~\ref*{methods:qa_interaction}}).
Only agent-based methods were evaluated, because rule-based and
machine-learning methods do not produce text.

\paragraph{On the automated benchmark, \agentname{} answers scored higher and contained fewer hard failures.}
The benchmark comprised 1,020 questions, each linked to a reference
intervention or single-variable deterioration event and asked at a fixed time relative to it, for example 1\,h before treatment began or several hours later to assess the response (Supplementary Note~\ref{app:qa_questions} and Supplementary Table~\ref{tab:app:qa_question_packs}).
The nine question types ranged from explaining an alert and judging its
urgency to weighing evidence for and against the concern, planning the next monitoring steps and assessing the treatment response .
Claude Opus 4.8 scored each answer from 1 to 5 using a prespecified rubric, a set of question-type criteria covering directness, organization of supporting evidence, clinical framing and safety of scope; the score was capped if the answer contained any of nine prespecified hard failures, such as hallucinated facts, unsafe recommendations or use of future information (Supplementary Table~\ref{tab:app:qa_hardfail}).
Because a question was posed only after a system had alerted on its event (\hyperref[methods:clinical_qa]{Methods, Section~\ref*{methods:clinical_qa}}), \agentname{}, which detected more events, answered more questions than any single-agent LLM baseline (732 versus 394--606).
Its mean rubric score was 4.44, compared with 3.55--4.11 for the single-agent LLM baselines (Fig.~\ref{fig:qa_evaluation}a).
On questions answered by both systems, it scored higher than each single-agent LLM baseline (all Holm-adjusted $P<0.001$; Fig.~\ref{fig:qa_evaluation}b), including the Qwen baseline (4.46 versus 3.90 on 512 questions), and this difference was observed in each of the three datasets and all nine question types (Supplementary Tables~\ref{tab:app:qa_matched_qwen} and~\ref{tab:app:qa_matched_types}; Fig.~\ref{fig:qa_evaluation}c).
Hard failures occurred in 7.4\% of \agentname{} answers and in 16.8--26.3\% of those of the single-agent LLM baselines (Fig.~\ref{fig:qa_evaluation}d) and were less frequent with \agentname{} in each paired comparison, for example 7.0\% versus 19.7\% with the Qwen baseline (all Holm-adjusted $P<0.001$; Fig.~\ref{fig:qa_evaluation}e).
\agentname{} had the lowest rate of overprediction, inappropriate escalation or major omission (1.5\% versus 4.8--20.4\%), whereas its rate of hallucinated facts or unsupported diagnoses (5.7\%) was below that of the Qwen baseline (11.9\%) but within the range of the other single-agent LLM baselines (5.0--18.9\%; Fig.~\ref{fig:qa_evaluation}d).

\paragraph{Physicians rated most \agentname{} answers usable, with fewer safety
concerns than for a single-agent LLM baseline.}
In the interaction study, three ICU physicians followed patients' records
through a forward-only replay and paused it to ask their own questions, which
were not restricted to predefined types and did not require a preceding alert
(Supplementary Fig.~\ref{fig:app:qa_workflow}b).
We finally collected 207 questions, most concerned treatment and
investigation decisions (32.9\%) or interpretation and diagnosis (21.3\%), with
the remainder spread across clinical status, priorities, prognosis, handover
and record retrieval (Fig.~\ref{fig:qa_evaluation}f).
Each question was answered by \agentname{} and by a single-agent LLM baseline using the same language model. Blinded to the source, physicians reviewed both answers side by side and assigned each a quantitative clinical usability score on a four-point scale. They also flagged potentially important errors or omissions as safety concerns (\hyperref[methods:qa_interaction]{Methods, Section~\ref*{methods:qa_interaction}}).
Physicians rated 93.7\% of \agentname{} answers basically or directly usable,
including 44.0\% directly usable, and none unacceptable, and flagged no safety
concerns in 97.6\% (Fig.~\ref{fig:qa_evaluation}f,g).
Ratings differed between physicians, with the proportion rated directly usable
ranging from 18.9\% to 78.9\% (Fig.~\ref{fig:qa_evaluation}g).
For the same questions, answers from the single-agent LLM baseline were less often rated
basically or directly usable (84.1\% versus 93.7\%) or directly usable (35.7\%
versus 44.0\%) and more often flagged for safety concerns (8.2\% versus 2.4\%;
Fig.~\ref{fig:qa_evaluation}h).
%




\begin{figure}[!t]
\centering
\includegraphics[width=\linewidth]{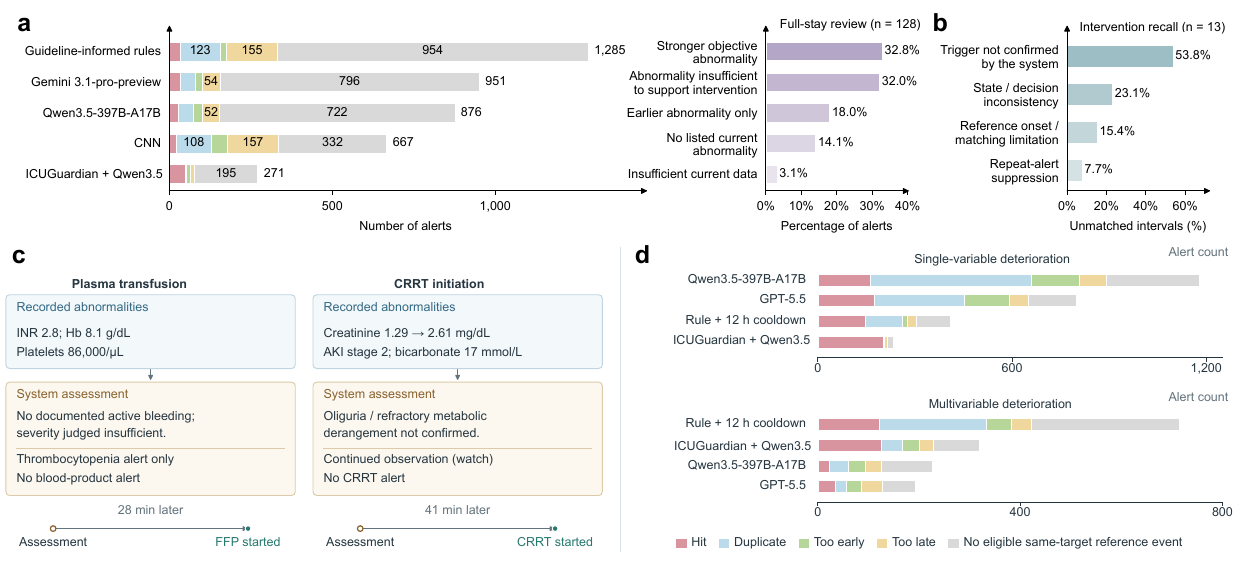}
\caption{\textbf{Failure-mode analysis across monitoring tasks.}
\textbf{a}, Individual-alert classification of raw intervention alerts (without episode merging) against 61 eligible reference intervals pooled across three datasets, for \agentname{} and comparators selected by F1; numbers show alert counts. Right, objective abnormality at the time of the 128 \agentname{} alerts without a same-target intervention recorded during the stay, as percentages of these alerts.
\textbf{b}, Causes of the 13 intervention reference intervals missed by \agentname{}, as percentages of these intervals.
\textbf{c}, Illustrative missed intervention intervals for plasma transfusion in MIMIC-IV (left) and continuous renal replacement therapy (CRRT) initiation in NWICU (right), in which \agentname{} recorded the underlying abnormalities but judged specific intervention criteria unconfirmed.
\textbf{d}, Classification of raw alerts for single-variable deterioration pooled across three datasets (206 primary reference events; top) and multivariable deterioration in MIMIC-IV (175 primary reference events; bottom). In \textbf{a} and \textbf{d}, grey denotes alerts without an eligible same-target reference event in the stay.}
\label{fig:failure_modes}
\end{figure}

\subsection{Failure-Mode Analysis and Case Studies}

To understand how \agentname{} reached its decisions and where it still failed, we used the patient state, tool calls and rationale recorded at each assessment to classify missed events and unmatched alerts across the cohort and to reconstruct individual cases (Supplementary Notes~\ref{app:failure_decomposition} and~\ref{app:case_reconstruction}).


\paragraph{Most missed interventions reflected conservative application of plan
criteria.}
Of the 61 intervention reference intervals, \agentname{} missed 13
(Fig.~\ref{fig:failure_modes}b).
In 7 of these, it had recorded the relevant abnormality but judged that the
criteria for the specific intervention were not yet met.
For example, it missed plasma transfusion despite recognizing coagulopathy,
citing no documented active bleeding (Fig.~\ref{fig:failure_modes}c, left).
It also recognized acute kidney injury but issued no kidney replacement
therapy alert because oliguria or refractory metabolic derangement
remained unconfirmed (Fig.~\ref{fig:failure_modes}c, right).
The remaining six unmatched intervals involved state integration or
alert-decision inconsistencies ($n=3$), reference-onset ambiguity or
one-to-one matching constraints ($n=2$), and repeat-alert suppression
($n=1$; Fig.~\ref{fig:failure_modes}b).
%
%

\paragraph{Alerts without eligible reference events limited precision
across tasks.}

%
To examine what limited precision across the replayed ICU stays, we classified all emitted raw alerts individually across tasks.
Alerts without eligible same-target reference events were a prominent contributor, particularly in intervention anticipation and multivariable deterioration, accounting for 72.0\% (195 of 271) and 28.5\% (91 of 319) of all \agentname{} alerts in the two tasks, respectively, compared with 82.4\% (722 of 876) and 44.9\% (102 of 227) for the Qwen baseline (Fig.~\ref{fig:failure_modes}a,d).
%
%
However, consistent with the physiological shift at alert times across all intervention alerts, most of these coincided with an abnormality even when no treatment followed: of the 128 with no recorded same-target intervention at any time in the stay, 64.8\% (83 of 128) coincided with a target-relevant abnormality (Fig.~\ref{fig:failure_modes}a, right), although this alone does not
establish that intervention was indicated.
Duplicate alerts were another substantial source of alert burden:
they accounted for 42.3\% (498 of 1,177) of alerts for the Qwen baseline in single-variable deterioration and 29.6\% (211 of 714) for the threshold rule with a cooldown in multivariable deterioration.
The corresponding proportions for \agentname{} were 0\% (0 of 234) and 12.9\% (41 of 319), with markedly fewer duplicates in absolute
terms (Fig.~\ref{fig:failure_modes}d).

\paragraph{Case studies illustrated the use of longitudinal patient context.}
In an illustrative MIMIC-IV stay, the State-maintenance agent
followed the Planner's monitoring plan to update MAP trends, organ
function and treatment history (Supplementary Fig.~\ref{fig:case_analysis}a).
A recorded MAP of 64\,mmHg prompted a hypotension alert.
The maintained state subsequently documented a transient recovery to
67\,mmHg followed by a decline to 60\,mmHg.
The Monitoring agent combined this deterioration with prior
norepinephrine use and the absence of active vasopressor support to
issue a restart warning, 78.6\,min before the recorded restart.
In response to a physician's question about a sudden rise in heart rate,
\agentname{} similarly reconstructed the recent trajectory and integrated
vital signs, laboratory trends, procedures and medications
(Supplementary Fig.~\ref{fig:case_analysis}b).
The answer prioritized possible causes while specifying evidence requiring
verification, including ECG results, the infection source, volume status
and tolerance of a spontaneous breathing trial.

\section{Discussion}

\paragraph{Overview.} 
LLM-based clinical agents are primarily designed to respond reactively to individual queries, whereas ICU monitoring requires continuously integrating physiological trends, laboratory results, treatments and clinical guidelines without exacerbating alarm fatigue.
\agentname{} addresses this gap by combining ICU clinical tools, longitudinal memory and a multi-agent monitoring workflow in the \benchname{}.
Using ICU monitoring as a proof of concept, we demonstrate the potential for LLM-based agents with persistent memory to become long-horizon clinical assistants that autonomously carry out ongoing patient monitoring under clinician supervision, moving beyond passive responses to isolated queries.

Our main contributions can be summarized as follows:

\paragraph{Contribution 1: a comprehensive ICU agent harness of clinical tools and guidelines.}
\agentname{} provides an agent harness ($\mathcal{T}$) comprising 56 ICU-related clinical tools with a Model Context Protocol (MCP) interface and 12 parsed free-text clinical guidelines.
%
%
Alongside 5 \agentname{}-specific read-only memory tools, 51 general-purpose clinical tools cover EHR exploration, record reorganization by organ system, guideline-based calculators (such as oxygenation indices, SOFA scores and KDIGO acute kidney injury stages), event tracking and 4 shared SQL state-table memory tools.
Together, these tools and guidelines address gaps in general-purpose agents' access to structured clinical records and ICU-specific knowledge, enabling \agentname{} to support more grounded ICU decision-making.

\paragraph{Contribution 2: A longitudinal trajectory memory and multi-agent workflow transform clinical agents into long-horizon assistants.}
\agentname{} couples \textit{Chrono-Clinical Trajectory Memory}~(CCTM, $\mathcal{M}$) with a role-specialized multi-agent workflow ($\bm{\mathcal{A}}$).
CCTM maintains a continuous record of patient states and agent actions in five interlinked, SQL-backed files covering monitoring plans, patient states, alerts, question--answer records and 12-h summaries.
It serves as the working memory of the five sub-agents, which use it to coordinate their specialized roles, from establishing monitoring plans and updating patient states to assessing alerts and answering questions.
This shared-memory multi-agent workflow enables clinical agents to manage complex, longitudinal ICU monitoring tasks and provide proactive assistance over extended periods.

\paragraph{Contribution 3: An interactive, time-gated \benchname{} for evaluating longitudinal monitoring tasks.}
\benchname{} is a simulated environment for evaluating methods on longitudinal monitoring tasks.
It spans three critical-care databases (MIMIC-IV, eICU-CRD and NWICU) and harmonizes more than 250,000 clinical records from 100 test ICU stays (317.5 patient-days; 7,742 monitoring decision points).
During retrospective replay, the environment advances in 1-h steps and exposes only records available at each timestamp, excluding free-text notes, discharge diagnosis codes and future treatment end times to limit hindsight leakage.
This time-gated design supports evaluation of intervention anticipation, deterioration detection, clinical question answering and physician--AI interaction under real-time monitoring conditions.

In evaluation, our experiments support three main findings on how \agentname{} performs in the ICU scenario.
\begin{itemize}[itemsep=6pt,parsep=3pt]
    \item \textbf{\agentname{} improves event capture while reducing alert burden.} Compared with rule-based, machine-learning and single-agent LLM baselines, \agentname{} substantially improved intervention and deterioration detection while curbing repeated alerts across cohorts and LLM backbones.
    \item \textbf{\agentname{} improves clinical question answering quality and supports more reliable physician--AI interaction.} By grounding responses with detailed rationales, \agentname{} outperformed all baselines with fewer hard failures and higher physician-rated usability and safety.
    \item  \textbf{\agentname{} issues clinically appropriate alerts with useful lead times, based on conservative, physiologically grounded reasoning.} Compared with recorded intervention start times, source-blinded physicians judged a substantial proportion of the alert times proposed by \agentname{} to be more appropriate. Although translating alerts into actual interventions still depends on many other real-world factors, including resource availability~\cite{stanworth2020covid} and variation in practice~\cite{raasveld2023input,carson2023red}, our retrospective evaluation suggests that \agentname{} has the potential to alert clinicians at clinically appropriate times and support more timely intervention.
\end{itemize}
Together, these findings underscore the value of \agentname{} in prompting earlier bedside review and assisting clinicians with longitudinal ICU monitoring and collaborative decision-making over extended time horizons.

\paragraph{Limitations and future work.}
Despite this potential, our work still has two main limitations.
First, retrospective replay follows fixed EHR trajectories rather than live clinical care; although time-gating limits hindsight leakage, it cannot capture unrecorded bedside findings or reveal how acting on earlier alerts would alter subsequent treatment and patient courses.
Second, our evaluation was bounded in scale and scope, covering 100 event-enriched stays over 72--96\,h windows with prespecified targets and three participating physicians, rather than unselected stays or fixed-horizon risk forecasting.
Future work should evaluate \agentname{} in larger, multicentre cohorts spanning full ICU stays, followed by prospective, clinician-supervised testing in routine ICU workflows to assess its impact on monitoring and intervention.

\section{Methods}

We first introduce \agentname{}, describing its tool set, longitudinal memory, and multi-agent workflow. We then present the \benchname{}, an ICU monitoring benchmark built from three critical-care databases, detailing its simulated ICU environment and experimental and evaluation procedures.

%

\subsection{\agentname{} Framework}
\label{methods:icuagent}

\begin{figure}[!t]
\centering
\includegraphics[width=\linewidth]{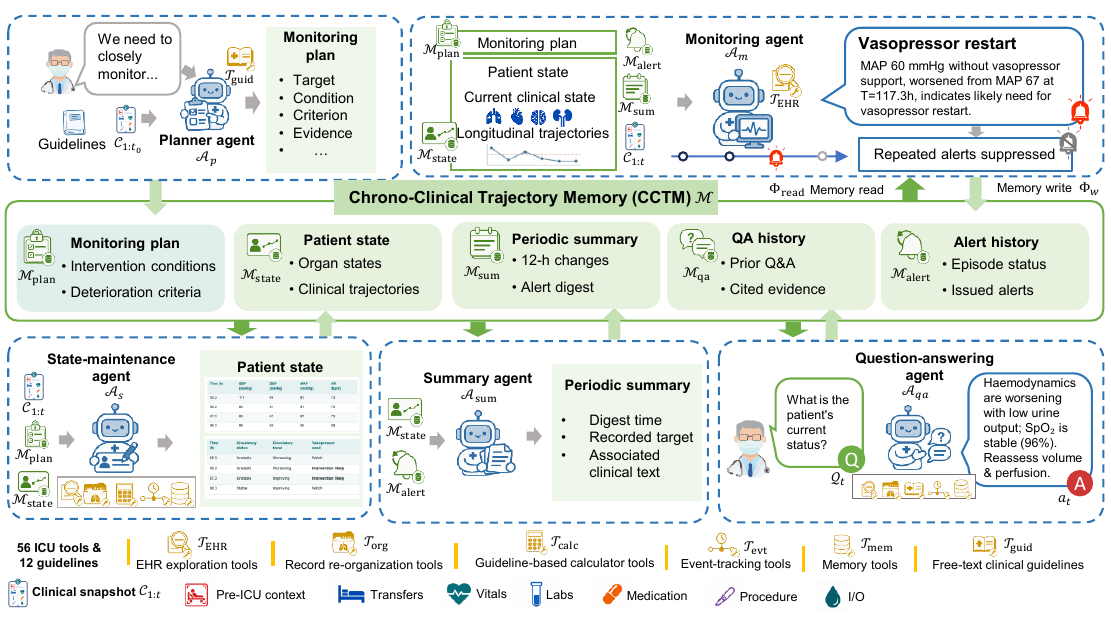}
\caption{\textbf{Multi-agent architecture and memory-mediated workflow of
\agentname{}.}
Five role-specialized sub-agents---the Planner ($\mathcal{A}_p$),
State-maintenance agent ($\mathcal{A}_s$), Monitoring agent ($\mathcal{A}_m$),
Summary agent ($\mathcal{A}_{\mathrm{sum}}$) and Question-answering agent
($\mathcal{A}_{qa}$)---coordinate through the Chrono-Clinical Trajectory Memory
(CCTM, $\mathcal{M}$), which comprises five linked components: the monitoring
plan ($\mathcal{M}_{\mathrm{plan}}$), patient state
($\mathcal{M}_{\mathrm{state}}$), periodic summaries
($\mathcal{M}_{\mathrm{sum}}$), question--answer history
($\mathcal{M}_{\mathrm{qa}}$) and alert history ($\mathcal{M}_{\mathrm{alert}}$).
Dark-green arrows denote memory reads ($\Phi_{\mathrm{read}}$) and light-green
arrows validated memory writes ($\Phi_w$).
At initialization ($t_0$), $\mathcal{A}_p$ combines the clinicians' monitoring
instructions, clinical guidelines ($\mathcal{T}_{\mathrm{guid}}$) and the initial
Clinical Snapshot ($\mathcal{C}_{1:t_0}$) into a monitoring plan, which is written
to $\mathcal{M}_{\mathrm{plan}}$ once accepted.
At each hourly assessment, $\mathcal{A}_s$ updates the patient state and
trajectories in $\mathcal{M}_{\mathrm{state}}$, and $\mathcal{A}_m$ assesses
intervention needs and deterioration from $\mathcal{C}_{1:t}$ and the CCTM, with
deterministic episode-aware rules suppressing repeated alerts.
Every 12\,h, $\mathcal{A}_{\mathrm{sum}}$ writes a summary of the preceding
interval to $\mathcal{M}_{\mathrm{sum}}$. $\mathcal{A}_{qa}$ answers clinicians'
free-text questions ($\mathcal{Q}_t$) with evidence-grounded responses ($a_t$)
whenever they are asked.
Bottom bars show the 56 ICU tools and 12 clinical guidelines, partitioned into
six subsets ($\mathcal{T}_{\mathrm{EHR}}$, $\mathcal{T}_{\mathrm{org}}$,
$\mathcal{T}_{\mathrm{calc}}$, $\mathcal{T}_{\mathrm{evt}}$,
$\mathcal{T}_{\mathrm{mem}}$ and $\mathcal{T}_{\mathrm{guid}}$; icons beside each
sub-agent indicate its subset), and the seven clinical modalities summarized in
the Clinical Snapshot ($\mathcal{C}_{1:t}$).}
\label{fig:icuagent_overview}
\end{figure}

\agentname{} is an LLM-based multi-agent system for long-horizon ICU monitoring that decouples upfront, guideline-grounded plan formulation from iterative state maintenance, alert assessment and on-demand clinical dialogue~\cite{kim2024llmcompiler}, connected by shared longitudinal memory (Fig.~\ref{fig:icuagent_overview}).

%

\subsubsection{Problem Formulation}

Formally, \agentname{} can be represented as $A = \{\bm{\mathcal{A}}, \mathcal{M}, \mathcal{T}\}$, where $\bm{\mathcal{A}}$, $\mathcal{M}$, and $\mathcal{T}$ denote the sub-agent ensemble, its memory, and its available external tools and guideline resources, respectively. The system comprises five sub-agents, \emph{i.e.}, a Planner, a State-maintenance agent, a Monitoring agent, a Question-answering agent, and a Summary agent collaborating according to a fixed high-level workflow, denoted as $\bm{\mathcal{A}} = \{\mathcal{A}_{p}, \mathcal{A}_{s}, \mathcal{A}_{m}, \mathcal{A}_{qa}, \mathcal{A}_{\mathrm{sum}}\}$.

%
For a given ICU patient, the monitoring task is formulated as $\mathcal{X} = \{\mathcal{I}, \mathcal{R}_{1:T}\}$, where $\mathcal{I}$ denotes the clinicians' free-text monitoring instructions for the patient, and $\mathcal{R}_{1:T}$ denotes the patient's EHR records as they evolve over $T$ time steps. 
At assessment $t$, the workflow constructs a Clinical Snapshot $\mathcal{C}_{1:t}$ from the available patient records $\mathcal{R}_{1:t}$, summarising current measurements and treatment status.
%
Let $\mathcal{G}$ denote the set of monitoring targets specified by $\mathcal{I}$. For each target $g \in \mathcal{G}$ at time step $t$, our agent outputs an alert decision with reasoning:
\begin{equation}
    (y_t,r_t) = A(\mathcal{R}_{1:t} \mid \mathcal{I})
\end{equation}
where $y_t = 1$ indicates that an alert is raised for target $g$ at time step $t$, and $y_t = 0$ otherwise. The agent also generates free-text reasoning $r_t$ to justify its decision, grounded in the relevant clinical guidelines or abnormal measurement values.
Additionally, clinicians may pose free-text questions $\mathcal{Q}_t$ at any point during monitoring, to which the agent returns a free-text response $a_t$:
\begin{equation}
    a_t = A(\mathcal{R}_{1:t} \mid \mathcal{Q}_t)
\end{equation}

Next we will detail the three main parts, $\{\bm{\mathcal{A}}, \mathcal{M}, \mathcal{T}\}$, in our system. For simplicity, we denote all prompts used below as $P_i$ and provide their full text in Supplementary Note~\ref{app:icuagent_prompts}.
We next describe the tool collection $\mathcal{T}$, longitudinal memory $\mathcal{M}$ and multi-agent workflow $\bm{\mathcal{A}}$.

\subsubsection{Collection of Basic Clinical Tools and Resources}
\label{methods:icuagent_tools}

Continuous ICU monitoring relies on several basic tools to perform underlying operations, such as accessing EHR systems, calculating quantitative assessments, reading related clinical guidelines, and, also, managing its own memory files.
%
%
We therefore developed a collection of 56 clinical tools, specified them in a unified set of structured description files, and provided a Model Context Protocol (MCP) interface to support agent use.
Among these, 5 tools are dedicated to memory organization for our \agentname{}, while the remaining 51 address general clinical needs. Generally, these tools can be organized into 5 high-level categories:
\begin{itemize}[itemsep=6pt,parsep=3pt]
    \item \textbf{EHR exploration tools} enable agents to retrieve patient data from the EHR file systems on demand. For example, an agent can request vital signs or medication records, and the tools return the relevant entries with timestamps and source information.
    
    \item \textbf{Record re-organization tools} organize retrieved patient data by organ system, such as cardiovascular, respiratory, renal, and neurological status, and report missing or potentially unreliable data to clarify the limitations of the available evidence.
    
    \item \textbf{Guideline-based calculator tools} translate clinical guidelines into executable procedures for calculating established clinical indices and scores. These tools are adapted from MDCalc~\cite{mdcalcwebsite}, MedCalc-Bench~\cite{khandekar2024medcalcbench}, and medcalcmcp~\cite{winninghealthmedcalcmcp}, and cover measures such as oxygenation indices, Sequential Organ Failure Assessment (SOFA) scores, and Kidney Disease: Improving Global Outcomes (KDIGO) acute kidney injury (AKI) stages. Given extracted patient data, they return calculated results, the contributing measurements, and any missing inputs.
    
    \item \textbf{Event-tracking tools} group related measurements and treatments over time to summarize the course of a clinical event. Unlike `EHR exploration' tools, which retrieve individual data points, they use saved structured elements to track events such as hypotension and determine whether each event is new, ongoing, or resolved, as well as whether related support, such as vasopressors, has been initiated.
 
\item \textbf{Memory tools} provide basic SQL operations that allow agents, including the single-agent LLM baselines, to define memory tables, retrieve and update records, and append notes across assessments. Additional read-only tools specific to \agentname{} are integrated with its memory mechanism to retrieve information such as recent alerts and patient-state summaries.
    
\end{itemize}
Beyond these tools, we also collected 12 publications relevant to ICU intervention decisions, covering clinical guidelines and assessment frameworks:
\begin{itemize}[itemsep=6pt,parsep=3pt]
    \item \textbf{Free-text clinical guidelines} provide agents with task-specific guidance from a collection of parsed free-text ICU guidelines. This collection covers the circulatory, respiratory, renal and haematological systems, together with the assessment of multisystem organ dysfunction. The sources include clinical guidelines, consensus statements and assessment studies. Although these resources are not ``operational'' tools in the narrow sense, they provide agents with external clinical knowledge in free-text form. Agents can also access~(or, in other words, call) them on demand to retrieve relevant guidance, much as they invoke other tools to obtain information. Therefore, for simplicity, we include them in this section and represent them as a subset in $\mathcal{T}$.
\end{itemize}  
The tools and guidelines are listed in Supplementary Tables~\ref{tab:app:toolset_retrieval}--\ref{tab:guideline_sources}.
%

%
Denoting this unified collection of 56 clinical tools and 12 parsed guidelines by $\mathcal{T}$, we partition $\mathcal{T}$ into six subsets, one for each category, to manage the resources available to each sub-agent:
\begin{equation}
\left\{
\mathcal{T}_{\mathrm{EHR}},
\mathcal{T}_{\mathrm{org}},
\mathcal{T}_{\mathrm{calc}},
\mathcal{T}_{\mathrm{evt}},
\mathcal{T}_{\mathrm{mem}},
\mathcal{T}_{\mathrm{guid}}
\right\}.
\end{equation}

\subsubsection{Chrono-Clinical Trajectory Memory Design}
\label{methods:cctm}

Longitudinal ICU monitoring is a long-horizon task that requires agents to track changes in a patient's clinical state over time. 
We therefore designed \textit{Chrono-Clinical Trajectory Memory} (CCTM), a structured memory module that continuously tracks and summarizes the patient's clinical state and the agent's prior actions.

\textbf{Memory structure.} CCTM comprises five interlinked structured JSON files, backed by SQL tables as:
\begin{equation}
\mathcal{M}=\left(
\mathcal{M}_{\mathrm{plan}},
\mathcal{M}_{\mathrm{state}},
\mathcal{M}_{\mathrm{alert}},
\mathcal{M}_{\mathrm{qa}},
\mathcal{M}_{\mathrm{sum}}
\right),
\label{eq:icuagent_memory}
\end{equation}
each file captures a distinct aspect of history context: 
\begin{itemize}[itemsep=6pt,parsep=3pt]
    \item The monitoring-plan memory, $\mathcal{M}_{\mathrm{plan}}$, stores the task objectives, evidence requirements, intervention conditions, and prespecified deterioration criteria identified during the beginning of the monitoring task assigned. 
    The task objectives also serve as shared monitoring-target indices, linking the monitoring plan to corresponding records in the other files.
    This memory is protected from subsequent updates, ensuring that the agent maintains consistency and retains the original monitoring objectives. 

    \item    The patient state $\mathcal{M}_{\mathrm{state}}$ stores two types of information: the current clinical state and the longitudinal trajectories of clinical variables, calculated scores and treatment support. 
    The current clinical state is organized by organ system, with a separate free-text summary indexed under each system (e.g., circulatory, respiratory, and renal function).
    It also includes a list of `evidence' items, each describing an abnormal finding in free text and linked to a monitoring-target index and its recorded timestamp. These items indicate, for each monitoring target, whether the evidence supports continued alerting or suggests a likely near-term need for intervention.
    The longitudinal trajectory part tracks measurement values across assessments in structured fields. For each clinical variable or calculated score, it records the current value, a categorical trend~(e.g., worsening or improving), and whether the corresponding deterioration criterion is met~(a binary indicator). For each treatment, it records whether it is active, along with timestamps for each start, stop, or restart event.

    \item $\mathcal{M}_{\mathrm{alert}}$ and $\mathcal{M}_{\mathrm{qa}}$ capture the agent's interaction history. $\mathcal{M}_{\mathrm{alert}}$ records when each alert episode began, resolved, or recurred, along with alerts previously issued by the agent. $\mathcal{M}_{\mathrm{qa}}$ stores earlier questions and answers together with the patient data that each answer cited, such as vital signs and medications.
    %

    \item The $\mathcal{M}_{\mathrm{sum}}$ stores periodic summaries of the earlier course of the stay. Each summary covers the period since the previous one and records how many alerts were issued and their monitoring targets, such as vasopressor initiation, together with changes in the patient state and newly abnormal signals.

    \item \textbf{Cross-file linkage.} Records across the memory files can be linked heuristically using shared monitoring-target indices and timestamps. Each plan in $\mathcal{M}_{\mathrm{plan}}$ is assigned an index that identifies its monitoring target; for example, an ``acute kidney injury (AKI)'' index denotes a plan to monitor for AKI. Related patient-state records in $\mathcal{M}_{\mathrm{state}}$ carry the same index, linking them to the plan. Records in $\mathcal{M}_{\mathrm{state}}$, $\mathcal{M}_{\mathrm{alert}}$, and $\mathcal{M}_{\mathrm{qa}}$ also include timestamps, which allow the agent to identify the patient state associated with a given alert or answer.

\end{itemize}
Such file-level separation enables us to distinguish patient facts from agent actions, preventing issued alerts from being mistaken for confirmed clinical events and thereby reducing the risk that unsupported alerts propagate through the agent’s memory; an illustrative example of linked records across all five memory components is provided in Supplementary Fig.~\ref{fig:app:memory_example}. Next, we will introduce its writing and reading operation.

\paragraph{Memory writing.} The writing operation $\Phi_{w}$ can be formulated as:
\begin{equation}
    \mathcal{M} \leftarrow \Phi_{w} (\mathcal{M},R  \mid \mathbb{I}),
\label{eq:icuagent_memory_write}
\end{equation}
where $R$ denotes a certain record generated during agent execution, for example a raised alert or a finished question--answer interaction, and $\mathbb{I} \in \{\mathrm{plan}, \mathrm{state},\mathrm{alert}, \mathrm{qa}, \mathrm{sum}\}$ indicates its target memory file. 

Specifically, $\Phi_{w}$ first validates each proposed update. It checks that required fields are present and have the expected data types (e.g., score values must be numeric), that records include the necessary monitoring-target indices and timestamps, and that protected content, such as the initialized monitoring plan, is not modified.
Validated updates are then merged into the corresponding memory components. Records for $\mathcal{M}_{\mathrm{alert}}$ and $\mathcal{M}_{\mathrm{qa}}$ are appended. In $\mathcal{M}_{\mathrm{state}}$, the overall assessment of the patient's clinical state is updated, while quantitative measurements are appended with timestamps. At each 12-h shift boundary, the workflow summarises the new state changes and monitoring events and stores the summary in $\mathcal{M}_{\mathrm{sum}}$. $\mathcal{M}_{\mathrm{plan}}$ is not updated after initialization because its contents are protected.

\paragraph{Memory reading.} The reading operation $\Phi_{\mathrm{read}}$ can be formulated as:
\begin{equation}
    C=\Phi_{\mathrm{read}}(\mathcal{M} \mid \mathbb{I}),
\end{equation}
where $\mathbb{I}$ specifies the memory type to retrieve, and $C$ denotes the returned free-text content, which is loaded into the LLM's context to support its task. For $\mathcal{M}_{\mathrm{plan}}$, the whole recorded plan returns. For $\mathcal{M}_{\mathrm{state}}$, the system returns the current clinical state and 12-hour trajectories of tracked variables, including all records from the most recent 6 hours and up to eight uniformly sampled records per variable from the preceding 6 hours. For $\mathcal{M}_{\mathrm{alert}}$, the system returns up to 12 of the most recent alert records from the past 12 hours. For $\mathcal{M}_{\mathrm{qa}}$ the recent 12 question-answer records are returned. For  $\mathcal{M}_{\mathrm{sum}}$, the system returns up to four of the most recently generated historical summaries.

\textbf{Notably,} unlike in the write operation, $\mathbb{I}$ here can specify multiple memory components (e.g., $\{\mathrm{qa}, \mathrm{state}\}$), whose contents are read in order and concatenated into a single text string. Additionally, we need to emphasize, $\Phi_{\mathrm{read}}$ and former $\mathcal{T}_{\mathrm{mem}}$ serve distinct roles. $\Phi_{\mathrm{read}}$ provides a fixed predefined memory view for the current task, while $\mathcal{T}_{\mathrm{mem}}$ exposes basic SQL read operation, allowing LLM to retrieve further information as needed, such as older records, after reviewing what $\Phi_{\mathrm{read}}$ returns.

\subsubsection{Multi-Agent Workflow}
 
\label{methods:icuagent_workflow}

In this section, we introduce the main multi-agent workflow of \agentname{}, which comprises five sub-agents: a Planner, a State-Maintenance Agent, a Monitoring Agent, a Question-Answering Agent, and a Summary Agent: $\bm{\mathcal{A}} = \{\mathcal{A}_{p}, \mathcal{A}_{s}, \mathcal{A}_{m}, \mathcal{A}_{qa}, \mathcal{A}_{\mathrm{sum}}\}$, and is built on the collected tools $\mathcal{T}$ and the memory module $\mathcal{M}$.

\paragraph{Initialization and planning.} Assuming the monitoring task is assigned at time step $t_0$, the case can be represented as $\mathcal{X} = \{\mathcal{I}, \mathcal{R}_{1:t_0}\}$. The planner agent is first called and delivers a detailed monitoring plan $\Pi$ based on collected guidelines and the task instruction:
\begin{equation}
\Pi=\mathcal{A}_p\!\left(\mathcal{I},\mathcal{C}_{1:t_0} \mid P_1, \mathcal{T}_\mathrm{guid}\right),
\label{eq:icuagent_planning}
\end{equation}
where $\mathcal{A}_{p}$ is LLM-powered, and $\mathcal{C}_{1:t_0} = \Phi_{\mathrm{sum}}(\mathcal{R}_{1:t_0})$ provides a clinical snapshot by heuristically extracting predefined ICU items from the patient's EHR (Section~\ref{methods:data_replay}).
Guided by prompt $P_{1}$, it analyzes the monitoring task from two supplemental perspectives, \emph{i.e.}, intervention and deterioration, and generates a holistic monitoring plan $\Pi$, specifying the deterioration criteria to monitor, the conditions suggesting a need for intervention, and the clinical variables and treatment status to track (an example plan is shown in Supplementary Fig.~\ref{fig:app:memory_example}a).
The Planner then self-reflects on its generated plan, prompted by $P_2$:
\begin{equation}
\mathcal{A}_p\!\left(\Pi \mid P_2, \mathcal{T}_\mathrm{guid}\right) \in \{\mathrm{pass},\mathrm{reject}\},
\end{equation}
Rejected plans are returned to the first step with the specific defects and revised for a limited number of rounds, while accepted plans are retained and stored in memory:
\begin{equation}
\mathcal{M}\leftarrow
\Phi_w\!\left(
    \mathcal{M},
    \Pi    \mid \mathbb{I} = \mathrm{plan}
\right).
\label{eq:icuagent_plan_write}
\end{equation}

\paragraph{Patient-state maintenance.} After the initial time step $t_0$, at each reassessment step $t$, the State-Maintenance Agent $\mathcal{A}_{s}$ is called to integrate new evidence with the patient's prior course, update the clinical state and provide a preliminary assessment of each monitoring target:
\begin{equation}
    R_t^{\mathrm{state}}= \mathcal{A}_s\!\left(\mathcal{C}_{1:t},  \mathcal{M}_{\{\mathrm{plan},\mathrm{state}\}} \mid P_3,\mathcal{T}_{s} \right).
\label{eq:icuagent_state_maintenance}
\end{equation}
For simplicity, we write $\mathcal{M}_{\{\mathrm{plan},\mathrm{state}\}}$ to denote $\Phi_{\mathrm{read}}(\mathcal{M} \mid \{\mathrm{plan},\mathrm{state}\})$. Hereafter, by default, $\mathcal{M}_i$ refers to the content retrieved from memory via the read operation, rather than the underlying memory file. The State-Maintenance Agent has access to the tool set
$\mathcal{T}_s = \mathcal{T}_{\mathrm{EHR}} \cup \mathcal{T}_{\mathrm{org}} \cup \mathcal{T}_{\mathrm{calc}} \cup \mathcal{T}_{\mathrm{evt}} \cup \mathcal{T}_{\mathrm{mem}}$,
which enables it to retrieve patient records beyond the default clinical snapshot $\mathcal{C}_{1:t}$, access older information stored in memory, and use auxiliary analysis tools. 
The prompt $P_3$ defines two tasks. 
%
%
First, the agent updates the longitudinal trajectories of clinical variables and scores that have new evidence or are due for reassessment, computing derived indices and recording whether each prespecified deterioration criterion is met in $\mathcal{M}_{\mathrm{state}}$.
%
%
Second, it assesses each monitoring target against its plan, judging whether the evidence supports continued observation or a likely near-term need for intervention, and records this preliminary assessment as part of the current clinical state, together with any unresolved conditions.

Finally, the agent writes the updated patient-state record to memory to keep the latest measurements and their trends current:
\begin{equation}
\mathcal{M}\leftarrow
\Phi_w\!\left(
    \mathcal{M},R_t^{\mathrm{state}}
    \mid \mathbb{I} = \mathrm{state}
\right).
\label{eq:icuagent_state_write}
\end{equation}

\paragraph{Alert assessment.} At the same time step $t$, after the patient state is updated, the Monitoring agent $\mathcal{A}_m$ is called to assess intervention needs and clinical deterioration, proposing alert decisions with reasoning:
\begin{equation}
(y_t,r_t, g_t)= \mathcal{A}_m\!\left(
    \mathcal{C}_{1:t}, \mathcal{M}_{\{\mathrm{plan},\mathrm{state},\mathrm{alert},\mathrm{sum}\}}
    \mid P_4,\mathcal{T}_m\right).
\label{eq:icuagent_monitoring}
\end{equation}
Here, $\mathcal{T}_{m} = \mathcal{T}_{\mathrm{EHR}}$ enables the agent to retrieve additional patient information, such as treatment records.
%
%
Guided by $P_4$, it integrates current clinical state, recent trends, alert history, periodic course summaries and ongoing treatment to assess intervention conditions against the monitoring plans and to evaluate in context the prespecified deterioration indicators recorded during state maintenance.
The agent sets $y_t=1$ when an intervention warning is warranted or a prespecified deterioration criterion is met, and each proposed alert specifies its target, priority and supporting evidence.
%
The free-text reasoning $r_t$ explains the proposed decision with reference to the monitoring plans and patient evidence, and $g_t \in \mathcal{G}$ marks the monitoring-target index this alert belongs to (\emph{e.g.}, AKI) according to the record in $\mathcal{M}_{\mathrm{plan}}$.

The proposed alert with $y_t=1$ is then evaluated using deterministic episode-aware rules through checking the patient state in $\mathcal{M}_{\mathrm{state}}$ and the alert history in $\mathcal{M}_{\mathrm{alert}}$. 
For candidate intervention warning, we read the Boolean indicator of  treatment activity indexed by $g_t$ with latest timestamp in $\mathcal{M}_{\mathrm{state}}$. If the corresponding treatment is active, we set $y_t = 0$ and suppress the warning.
%
%
For remaining candidates, we check the episode record indexed by $g_t$ in $\mathcal{M}_{\mathrm{alert}}$. If an unresolved episode is already active, we set $y_t = 0$ to suppress repeated alerts unless $\mathcal{M}_{\mathrm{state}}$ records newly timestamped evidence after the previous alert together with clinical worsening; otherwise (new onset or recurrence after resolution), $y_t$ remains 1.
%
%
%
%
%
When $y_t$ remains 1, we send the alert $(y_t, r_t)$ to the user and record the emitted alert $(y_t, r_t, g_t)$ and any episode-status updates, together with its timestamp, in memory as $R_t^{\mathrm{alert}}$:
\begin{equation}
\mathcal{M} \leftarrow
\Phi_w\!\left(
    \mathcal{M}, R_t^{\mathrm{alert}}
    \mid \mathbb{I} = \mathrm{alert}
\right).
\label{eq:icuagent_event_write}
\end{equation}

\paragraph{Question answering.} During monitoring, clinicians may ask the agent questions in free text. Assuming the system receives a question at time step $t$, the Question-answering agent $\mathcal{A}_{qa}$ is invoked to answer it:
\begin{equation}
a_t = \mathcal{A}_{qa}\!\left(
    \mathcal{Q}_t,\mathcal{C}_{1:t},  \mathcal{M}_{\{\mathrm{state},\mathrm{alert},\mathrm{qa}\}}
    \mid P_5,\mathcal{T}_{qa} \right).
\label{eq:icuagent_qa}
\end{equation}
When additional evidence is needed, the tool set $\mathcal{T}_{qa} = \mathcal{T}_{\mathrm{EHR}} \cup \mathcal{T}_{\mathrm{org}} \cup \mathcal{T}_{\mathrm{evt}} \cup \mathcal{T}_{\mathrm{mem}} \cup \mathcal{T}_{\mathrm{guid}}$ allows the agent to retrieve longer trajectories, organ-system context, guideline references and current treatment support.
Prompt $P_{5}$ instructs the agent to identify the clinical focus and time frame of the question, integrate relevant evidence, and explain its answer. The agent distinguishes current findings from historical events and notes any uncertainties that may affect its conclusion.

Once the response is generated, the workflow constructs a structured record $R_t^{\mathrm{qa}}$ containing the question $\mathcal{Q}_t$, the answer $a_t$, and the cited patient data (such as vital signs and medications), together with its timestamp:
\begin{equation}
\mathcal{M}\leftarrow
\Phi_w\!\left(
    \mathcal{M}, R_t^{\mathrm{qa}}
    \mid \mathbb{I} = \mathrm{qa}
\right).
\label{eq:icuagent_qa_write}
\end{equation}

\paragraph{Periodic summary update.}
\agentname{} periodically invokes an LLM to summarize the patient's course since the previous summary. Specifically, every 12 hours, the summary agent $\mathcal{A}_{\mathrm{sum}}$  retrieves the patient-state and alert records for the interval from $\mathcal{M}_{\mathrm{state}}$ and $\mathcal{M}_{\mathrm{alert}}$, respectively, and provides them to the LLM with prompt $P_6$:
\begin{equation}
R_t^{\mathrm{sum}}
=
\mathcal{A}_{\mathrm{sum}}
\left(
\mathcal{M}_{\{\mathrm{state},\mathrm{alert}\}}
\mid P_6
\right).
\end{equation}
The summary describes changes in the patient's clinical state and longitudinal trajectories, newly abnormal signals, and the number of alerts issued during the interval and their monitoring targets.

The resulting structured record $R_t^{\mathrm{sum}}$ is then submitted to the memory-writing operation:
\begin{equation}
\mathcal{M} \leftarrow
\Phi_w\!\left(
    \mathcal{M}, R_t^{\mathrm{sum}}
    \mid \mathbb{I} = \mathrm{sum}
\right).
\label{eq:icuagent_summary_write}
\end{equation}

\subsection{\benchname{} Curation}
\label{methods:sandbox}

\benchname{} provides a simulated evaluation environment for longitudinal ICU
monitoring, combining retrospective patient-record replay, a unified
clinical data interface and the shared tool collection.
In the sandbox, the patient's state is updated according to the timestamps recorded in the EHR, and the assessed agents can access only information available at the current assessment time.
The following sections describe the source cohorts, harmonization and
record-visibility rules.

\subsubsection{Data Sources and Cohort}
\begin{figure}[!t]
\centering
\includegraphics[width=\linewidth]{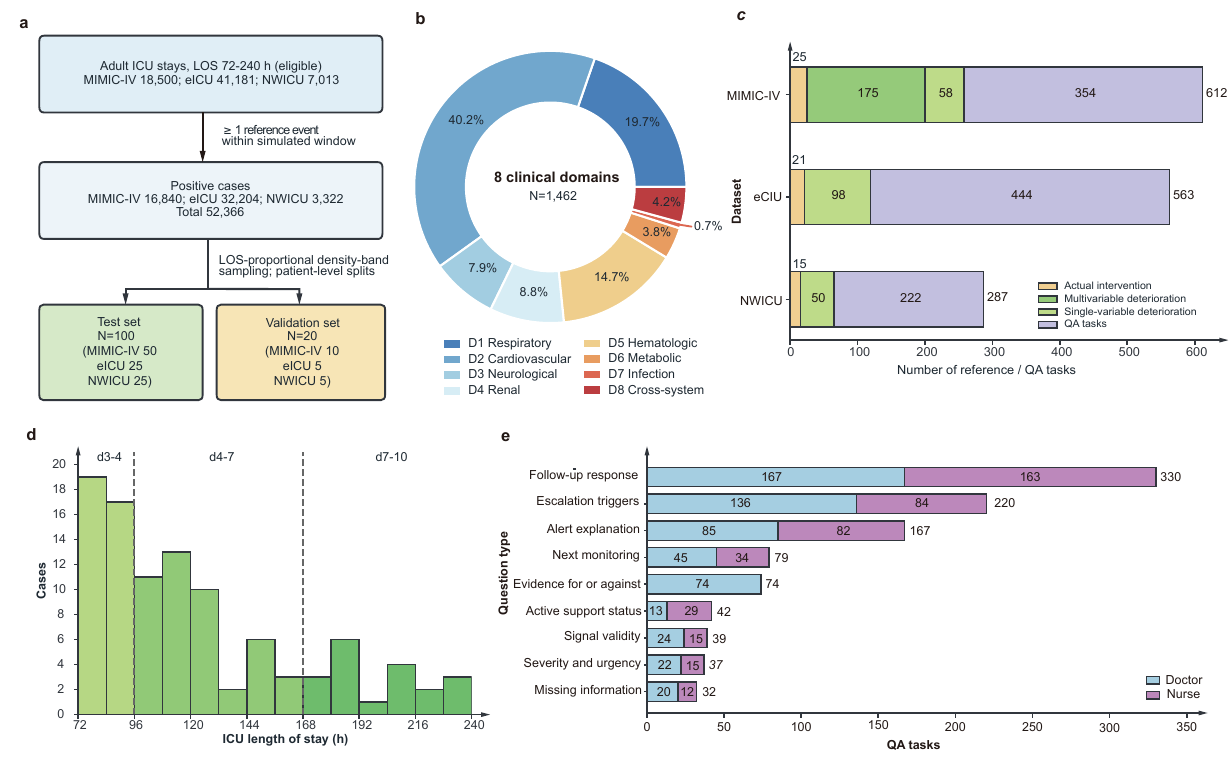}
\caption{\textbf{Cohort construction and clinical scope of \benchname.}
\textbf{a}, Selection of adult ICU stays with a length of stay (LOS) of
72--240\,h and at least one eligible reference event within the simulation
window. Patient-level sampling yields 100 test stays (50 MIMIC-IV, 25 eICU-CRD
and 25 NWICU) and 20 validation stays (10, 5 and 5, respectively).
\textbf{b}, Count-weighted distribution of 1,462 benchmark items across seven
clinical domains and an additional cross-system category. Items comprise
reference events and questions, rather than independent patients.
\textbf{c}, Composition by dataset: 61 actual-intervention episodes, 175
multivariable deterioration events, 206 single-variable deterioration events
and 1,020 clinical questions. Multivariable events are evaluated in MIMIC-IV
only. Paired Q\&A comparisons use jointly eligible scored answers from the
1,020-question benchmark.
\textbf{d}, LOS distribution of the 100 test stays in 12-hour bins. Dashed
lines mark the 96- and 168-hour boundaries used for stratified sampling.
\textbf{e}, The question pool by nine question types and intended
clinician role (physician or nurse). Counts describe benchmark composition,
not the number of answers scored for each system.}
\label{fig:data_form}
\end{figure}

We used three de-identified critical-care databases accessed under
PhysioNet credentialed data use agreements: MIMIC-IV v2.2~\cite{johnson2023mimic},
a single-centre database from the Beth Israel Deaconess Medical Center;
eICU-CRD v2.0~\cite{pollard2018eicu}, a multi-centre database from hospitals
across the United States; and NWICU v0.1.0~\cite{moukheiber2024nwicu}, from the
Northwestern Medicine health system.
Adult ICU stays (age $\geq$18 years) lasting 72--240\,h were eligible.
Stays shorter than 96\,h were replayed in full and longer stays over
their final 72\,h, balancing computational cost against coverage of the later
course of prolonged stays; reference events outside this replay window were
excluded.
Validation and test stays were sampled from each database with fixed
random seeds; test sampling excluded validation patients, and each patient
contributed at most one stay.
Sampling was stratified by length of stay (72--96, 96--168 and 168--240\,h),
with quotas proportional to each stratum's share of eligible stays in that
database.
The test set contained 100 stays (50 MIMIC-IV, 25 eICU-CRD and 25 NWICU) and
the validation set 20 (10, 5 and 5, respectively); the validation set was used
to develop \agentname{}, including refinement of its prompts.
Within each stratum, sampling favored stays with reference events in the replay window, so the test cohort was enriched for reference events relative to an unselected ICU population (Fig.~\ref{fig:data_form}, Supplementary Note~\ref{app:cohort_details} and Supplementary Table~\ref{tab:cohort}).
%

\subsubsection{Data Harmonization and Time-Gated Replay}
\label{methods:data_replay}

%
For each database, we implemented a rule-based adapter that used predefined mappings to link local codes and variable names to the corresponding clinical variables and event types in a common schema. The schema covered vital signs, laboratory results, medications, procedures, transfers and fluid intake and output.
The adapters also reconciled differences in how the databases organize their
records, for example by merging the separate vital-sign streams of eICU-CRD,
reconstructing medication administrations in NWICU from its electronic
medication administration record and converting NWICU temperatures from
Fahrenheit to Celsius.
Free-text notes and diagnosis codes assigned at discharge from the current
hospitalization were excluded to limit hindsight leakage into agent inputs.

We defined each ICU admission as time zero and converted database
timestamps, or time offsets in eICU-CRD, to minutes relative to that
admission.
A simulated clock advanced in 1-h steps.
At simulated time $T$, a record became visible once its recorded time had been
reached, including time-stamped diagnoses from the current or earlier
admissions, and ongoing treatments such as infusions were shown as active,
with their end times withheld until reached.
In MIMIC-IV and NWICU, laboratory results were timed by specimen collection,
which can precede their availability to clinicians.
When replay began partway through a stay, earlier ICU records were available
as history, but agents were assessed only within the replay window.
At each step, newly visible records were added to the patient record that
agents could query, and a Clinical Snapshot summarized what was known at that
time: demographics and admission diagnosis, the latest vital signs, laboratory
results from the preceding 24\,h, active medications and procedures, recent
transfers, 6-h and 24-h fluid balances and, during the first 24\,h after ICU
admission, the pre-ICU context.
The snapshot contained no reference-event labels or risk scores.
All record access, including tool calls, was restricted to the current stay
and simulated time, and notes stored by an agent across steps were visible
only to that agent.

\subsection{Baseline Methods}
\label{methods:experimental_settings}

All baselines were evaluated in \benchname{} once for each patient trajectory, with hourly assessments starting at the beginning of the replay window.
%
%
By default, \agentname{} used Qwen3.5-397B-A17B across all sub-agents (with GPT-5.5 additionally evaluated on the 50 MIMIC-IV stays).
%
%
We considered four main baseline categories: single-agent LLM baselines, guideline-informed rules, machine-learning models, and threshold-based rules. 
All single-agent LLM baselines were evaluated using the same system and prompts. For all LLMs, we used medium reasoning effort, a sampling temperature of 0, and a maximum output length of 2,048 tokens per call.

\paragraph{Single-agent LLM baselines.}

%
We evaluated five language models as single function-calling agents for ICU monitoring: GPT-5.5, Claude Sonnet~4.6, Gemini~3.1-Pro-Preview, DeepSeek-V4-Pro, and Qwen3.5-397B-A17B.
%
%
At each hourly assessment, each agent received the Clinical Snapshot $\mathcal{C}_{1:t}$, guideline references, and could select tools from the shared collection of 51 general-purpose clinical tools.
Four of these tools allowed the agent to maintain its own memory across assessments by defining a table for the current ICU stay, retrieving or updating entries, and appending notes.
%
%
%
Full implementation details, prompts and shared memory tools for the single-agent LLM baselines are provided in Supplementary Notes~\ref{app:single_agent_implementation} and~\ref{app:baseline_prompt_text} and Supplementary Table~\ref{tab:app:toolset_state}.


\paragraph{Guideline-informed rules.}
For intervention anticipation monitoring, existing clinical ICU alert systems primarily generate bedside alerts using predefined rules and scoring systems.
%
%
We constructed a deterministic rule-based comparator for intervention anticipation using published clinical guidelines~\cite{prescott2026surviving,helms2024oxygen,kdigo2012aki,carson2023red}.
The rules expressed intervention indications as combinations of physiological thresholds and clinical conditions.
Because rule-based eligibility criteria typically rely on a fixed set of inputs that may be incomplete in real-world EHRs, diagnosis-based criteria were not used to exclude a patient when the relevant diagnostic information was missing. Physiological thresholds, treatment requirements, and explicitly documented exclusion criteria remained binding.
An alert was issued when at least one rule for the intervention was
satisfied, the corresponding support was known to be inactive and no
documented treatment limitation applied.
The rule definitions, evidence windows and diagnostic waivers are detailed in Supplementary Note~\ref{app:baseline_settings} and Supplementary Table~\ref{tab:app:guideline_rules}.

\paragraph{Machine-learning models.}
Beyond rule-based scoring, several deep learning methods have also been proposed for intervention anticipation in clinical monitoring.
We implemented LightGBM~\cite{hyland2020early}, LSTM and CNN~\cite{suresh2017clinical}, and random-forest~\cite{wang2020mimicextract} baselines, adapting architectures and input representations from previous studies to the four intervention targets and available data.
Each model was trained to predict intervention initiation within the next 6\,h.
All models were trained and developed on separate MIMIC-IV cohorts
(14,785 training stays and 3,623 development stays), excluding patients
in the benchmark test and validation cohorts, and were applied to all
three databases without retraining.

\paragraph{Threshold rules.}
In current clinical practice, deterioration monitoring often relies on fixed, general-purpose threshold rules. 
%
%
At each hourly assessment, we applied prespecified single-variable thresholds to the latest available measurements or to changes over specified time windows (Section~\ref{sec:gt}).
In MIMIC-IV, we also evaluated multivariable criteria based on the SOFA score~\cite{vincent1996sofa}, Sepsis-3 definitions~\cite{singer2016sepsis3} and KDIGO acute kidney injury staging~\cite{kdigo2012aki}.
We evaluated two alerting strategies. The first issued an alert at every assessment where a criterion was met. The second suppressed repeat alerts for the same criterion during the same hospital stay for 12\,h after an alert was issued. We selected the suppression interval using the validation cohort (Supplementary Note~\ref{app:baseline_settings}).

\textbf{Notably,} the latter three method types (guideline-informed rules, machine-learning models, and threshold-based rules) do not fully match the scope of our approach. They are typically designed for specific, predefined monitoring tasks, whereas our agent system can support a broader range of tasks, provide more detailed explanations for its outputs, and respond to clinicians' free-text queries. We therefore compare these methods with our agent system on the specific monitoring tasks they are designed to address. These comparisons provide task-specific performance references and should not be interpreted as comparisons between fully equivalent systems.


\subsection{Automated Evaluation}
\label{methods:alert_evaluation}

We automatically evaluated methods for longitudinal monitoring and clinical question answering.

\subsubsection{Evaluation on Longitudinal Monitoring Tasks}
\label{methods:continuous_monitoring_benchmark}

\paragraph{Tasks and reference events.}
\label{sec:gt}

Reference events for the three monitoring tasks were defined from the
source records or from community-maintained derived concepts, independently of any evaluated system.
Intervention reference events were new starts of vasopressors, invasive
mechanical ventilation, continuous renal replacement therapy or blood-product
administration that could be evaluated independently.
Continuing, planned, perioperative and left-censored treatments were excluded,
and records belonging to the same clinical need were merged into a single
episode.
Working back from each treatment start, prespecified rules generated candidate
onset times from the preceding vital signs, laboratory results, treatments and
clinical context.
Where this evidence was insufficient or ambiguous, an LLM (Claude Opus 4.8)
reviewed the case to estimate when the need for the intervention first became
recognizable.
Physicians then confirmed the inclusion, merging or exclusion of each episode and accepted or revised its candidate onset (Supplementary Table~\ref{tab:intervention_onset_rules}).
Multivariable deterioration events, computed in MIMIC-IV with mimic-code
concepts~\cite{johnson2018mimic}, were SOFA~$\geq$~6~\cite{vincent1996sofa}, an
increase in SOFA of at least 2 points within 48\,h, Sepsis-3
onset~\cite{singer2016sepsis3}, KDIGO acute kidney injury
stage~$\geq$~2~\cite{kdigo2012aki} and worsening KDIGO stage.
Single-variable deterioration events, computed in all three databases, were
SpO$_2$ below 88\% and mean arterial pressure below 65\,mmHg (each on two
consecutive readings), lactate of at least 4\,mmol/L, arterial pH below 7.25,
potassium above 6\,mEq/L, platelets below $100\times10^9$/L and a decrease
within 24\,h of at least 2\,g/dL in haemoglobin.
%
%
%
Two additional subgroups---severe organ dysfunction (SOFA~$\geq$~8, $n=8$, nested within SOFA~$\geq$~6) and Glasgow Coma Scale decline of at least 2 points within 24\,h ($n=40$, which may reflect sedation)---were excluded from the 442 primary reference events (Supplementary Table~\ref{tab:cohort}).

\paragraph{Matching and metrics.}
For intervention anticipation, alerts for the same stay and target were grouped
into prediction episodes before matching.
A new episode began when the gap since the preceding alert exceeded 6\,h for
vasopressors, 12\,h for blood products or 24\,h for invasive mechanical
ventilation and continuous renal replacement therapy.
Prediction episodes that overlapped a warning interval were matched to
reference events one to one by maximum-overlap assignment.
For deterioration, an alert for the same criterion issued between 120\,min
before and 120\,min after an event could be matched to it, again one to one,
with the closest alert matched first.
Recall was the proportion of reference events that were matched, precision the
proportion of prediction units (prediction episodes for intervention
anticipation and individual alerts for deterioration) that matched a reference
event and F1 their harmonic mean:
\[
\text{recall}=\frac{N_{\text{matched}}}{N_{\text{ref}}},\qquad
\text{precision}=\frac{N_{\text{matched}}}{N_{\text{pred}}},\qquad
\text{F1}=\frac{2\times\text{recall}\times\text{precision}}
{\text{recall}+\text{precision}},
\]
where $N_{\text{ref}}$, $N_{\text{pred}}$ and $N_{\text{matched}}$ are the
numbers of reference events, prediction units and one-to-one matches.
Lead time was the interval from the first matched alert to the start of
treatment.
Alert burden was summarized as raw alert counts, raw alerts per detected event
and, for intervention alerts, the proportion merged into an existing episode.

\subsubsection{Evaluation on Question-Answering}
\label{methods:clinical_qa}
\label{methods:qa_task_construction}
\paragraph{Question construction.}
Following earlier EHR question-answering benchmarks that used predefined templates~\cite{pampari-etal-2018-emrqa,bardhan-etal-2022-drugehrqa}, we generated questions by filling the templates with details from extracted intervention and single-variable deterioration event anchors.
For each of nine question types, we defined the evaluation target, clinical
role (physician or nurse) and applicable question stages, and wrote a base
template with event placeholders.
Intervention-linked questions could be asked before the intervention, once
support became visible or at follow-up, with wording that did not reveal the
later intervention; deterioration-linked questions covered initial assessment
and follow-up.
%
Instantiated questions were scheduled at prespecified offsets aligned to the hourly replay, yielding 1,020 questions (303 intervention-linked and 717 single-variable deterioration-linked) that were fixed before evaluation (Supplementary Fig.~\ref{fig:app:qa_workflow}a, Supplementary Note~\ref{app:qa_protocol} and Supplementary Table~\ref{tab:app:qa_question_packs}).
A system received a question only if it had already issued an alert matched to
the same anchor, and it could access only information visible at that time.

\paragraph{Scoring rubric.}
Each answer was scored against a case-specific reference packet, containing
evidence anchors visible at the question time, key points, information gaps
and common errors, and a type-specific rubric with four axes: directness,
organization of supporting evidence, clinical framing and safety of scope.
Later records were used only to detect use of future information.
An LLM judge (Claude Opus 4.8), blinded to system identity, rated each axis from 1 to 5 and flagged
prespecified hard failures, such as fabricated facts, unsafe recommendations or
major omissions.
The final score was computed deterministically from the type-weighted axis scores and capped at the lowest ceiling among any hard failures (Supplementary Note~\ref{app:qa_scoring} and Supplementary Table~\ref{tab:app:qa_hardfail}).

\subsection{Physician Evaluation}
\label{methods:Physician_evaluation}

In addition to automated evaluation, we also involved physicians to assess the outputs of the different methods.

\subsubsection{Alert Timing Review}
\label{methods:alert_timing}
\label{methods:clinician_timing}

\paragraph{Alert pair construction.}
For each eligible recorded intervention, we paired its start time with one
\agentname{} alert for the same target, selected in order of preference: the
matched alert, otherwise the nearest preceding unmatched alert, otherwise the
nearest subsequent one. For this review, matching was performed on individual
alerts.
The two times had to be more than 60\,min apart, each intervention contributed
one pair and all resulting pairs were reviewed.
For these paired reviews, signed lead time was defined as the recorded treatment start minus the alert time and grouped as $<0$, 0--3, $>$3--6, $>$6--9, $>$9--12 and $>$12\,h.

\paragraph{Physician assessment.}
The paired times were presented in chronological order as T1 and T2, with
their sources concealed.
For each time, physicians saw a cumulative view of the record up to that time (except current-admission diagnoses; Supplementary Fig.~\ref{fig:app:timing_interface}),
covering vital signs, laboratory results, fluid intake and output, treatments and target-relevant clinical context; records revealing the target intervention were masked, and missing values were not treated as normal.
Three ICU physicians rated all pairs independently.
%
Each time was rated for actionability, that is, how appropriate a warning would
be at that time, from 1 (insufficient clinical support) to 5 (clear support for
a highly appropriate time), with 3 denoting an acceptable warning time despite
incomplete support; physicians then chose the better time, both or neither.
%
The primary outcome was the within-physician difference in ratings between the alert time and the recorded treatment start. Review procedures and statistical analyses are described in Supplementary Note~\ref{app:clinician_timing}.

\subsubsection{Physician--AI Interaction Study}
\label{methods:qa_interaction}

\paragraph{Interaction platform.}
The same three ICU physicians chose the stays they reviewed and followed each
patient's record through a forward-only replay, pausing at any point to ask
free-text questions based on the information available at that time
(Supplementary Fig.~\ref{fig:app:qa_workflow}b).
%
%
For each question, \agentname{} and a single-agent LLM baseline using the same language model received the same question time, initial evidence from the
time-gated record and access to the same set of time-gated clinical tools; the single-agent LLM baseline answered from this evidence and tool queries alone, whereas \agentname{} also drew on CCTM and its memory-query tools through the question-answering agent (physician instructions and the interaction interface are detailed in Supplementary Fig.~\ref{fig:app:qa_clinician_interface}).
%
%

\paragraph{Physician ratings.}
System identities were concealed and answer order was randomized.
Each physician rated the answers to their own questions on a four-point scale
(unacceptable, insufficiently informative, basically usable or directly
usable), with an unable-to-rate option, and separately flagged potentially
important errors or omissions as safety concerns.
The first answer was rated and locked before the second was shown, and after
rating both answers, physicians chose the better one.
%



\section*{Data availability}

All datasets and clinical knowledge resources used in this study are publicly released or publicly accessible resources subject to their original access terms. The longitudinal ICU evaluation cohorts comprise MIMIC-IV v2.2~\cite{johnson2023mimic}, eICU-CRD v2.0~\cite{pollard2018eicu} and NWICU v0.1.0~\cite{moukheiber2024nwicu}, together with the open-access MIMIC-IV Clinical Database Demo used for the online interactive demonstration. External clinical knowledge and tool resources include the 12 published clinical guidelines listed in Supplementary Table~\ref{tab:guideline_sources}, derived concept definitions from mimic-code~\cite{johnson2018mimic}, and clinical calculator specifications adapted from MDCalc~\cite{mdcalcwebsite}, MedCalc-Bench~\cite{khandekar2024medcalcbench} and medcalcmcp~\cite{winninghealthmedcalcmcp}. MIMIC-IV, eICU-CRD and NWICU are available through PhysioNet (\url{https://physionet.org/}) under credentialed access and the applicable data-use agreements. Case-level records, harmonized timelines and derived evaluation files that cannot be redistributed under the original dataset licences or data-use agreements will be accompanied by scripts and instructions allowing credentialed users to regenerate them from the original public sources.

\section*{Code availability}
The source code for \agentname{}, the \benchname{} replay harness, baseline implementations and evaluation scripts is publicly available at \url{https://github.com/MAGIC-AI4Med/ICUGuardian} under the Apache 2.0 license. An interactive web demonstration built on the publicly accessible MIMIC-IV Clinical Database Demo, enabling readers and reviewers to explore \benchname{} and interact with \agentname{} on longitudinal ICU stays, is available at \url{https://fridge-mosaic-shabby.ngrok-free.dev/demo-qa/login?return_to=%2Fdemo-qa%2F}; login credentials (username and password) for testing can be obtained by contacting the corresponding authors.




\section*{Competing interests}
The authors declare no competing interests.

\clearpage
\bibliographystyle{unsrt}
\bibliography{references}

\section*{Acknowledgements}
This work is supported by the Shanghai Municipal Special Program for Basic Research on General AI Foundation Models (Grant No. 2025SHZDZX026D12).

\section*{Author information}
\paragraph{Author contributions.}
C.W. and Y.Z. conceived and supervised the study. C.W., Q.Y. and P.Q. designed the methods and experiments. Q.Y., P.Q. and Q.Z. developed \agentname{} and \benchname{}, performed the benchmark experiments and conducted the statistical analyses. K.W. and S.T. performed the clinical evaluation, reviewed the reference events and provided clinical input. Q.Y. and C.W. wrote the manuscript. Y.Q. prepared the figures and contributed to writing. All authors reviewed and approved the final manuscript.

\section*{Ethics declarations}
This study was a retrospective analysis of de-identified critical-care records from MIMIC-IV v2.2, eICU-CRD v2.0 and NWICU v0.1.0, accessed through PhysioNet under credentialed data use agreements; original data collection and de-identification were approved by the respective institutional review boards with waivers of individual patient informed consent. All evaluations were conducted retrospectively in a simulated replay environment without affecting clinical care.

\clearpage
\appendix
\setcounter{section}{0}
\renewcommand{\thesection}{\arabic{section}}
\renewcommand{\theHsection}{supp.\arabic{section}}
\titleformat{\section}{\bfseries\fontsize{13}{15}\selectfont}{Supplementary Note \thesection:}{0.5em}{#1}[]
\setcounter{figure}{0}
\renewcommand{\theHfigure}{supp.\arabic{figure}}
\setcounter{table}{0}
\renewcommand{\theHtable}{supp.\arabic{table}}
\renewcommand{\figurename}{Supplementary Fig.}
\renewcommand{\tablename}{Supplementary Table}
\captionsetup[figure]{name={Supplementary Fig.}}
\captionsetup[table]{name={Supplementary Table}}


\makeatletter
\setlength{\@fptop}{0pt}
\setlength{\@fpsep}{14pt}
\setlength{\@fpbot}{0pt plus 1fil}
\makeatother
\setcounter{topnumber}{4}
\setcounter{bottomnumber}{4}
\setcounter{totalnumber}{6}
\renewcommand{\topfraction}{0.92}
\renewcommand{\bottomfraction}{0.85}
\renewcommand{\textfraction}{0.05}
\renewcommand{\floatpagefraction}{0.65}


\section*{Supplementary Figures}

\begin{figure}[!htbp]
\centering
\includegraphics[width=\linewidth,height=0.74\textheight,keepaspectratio]{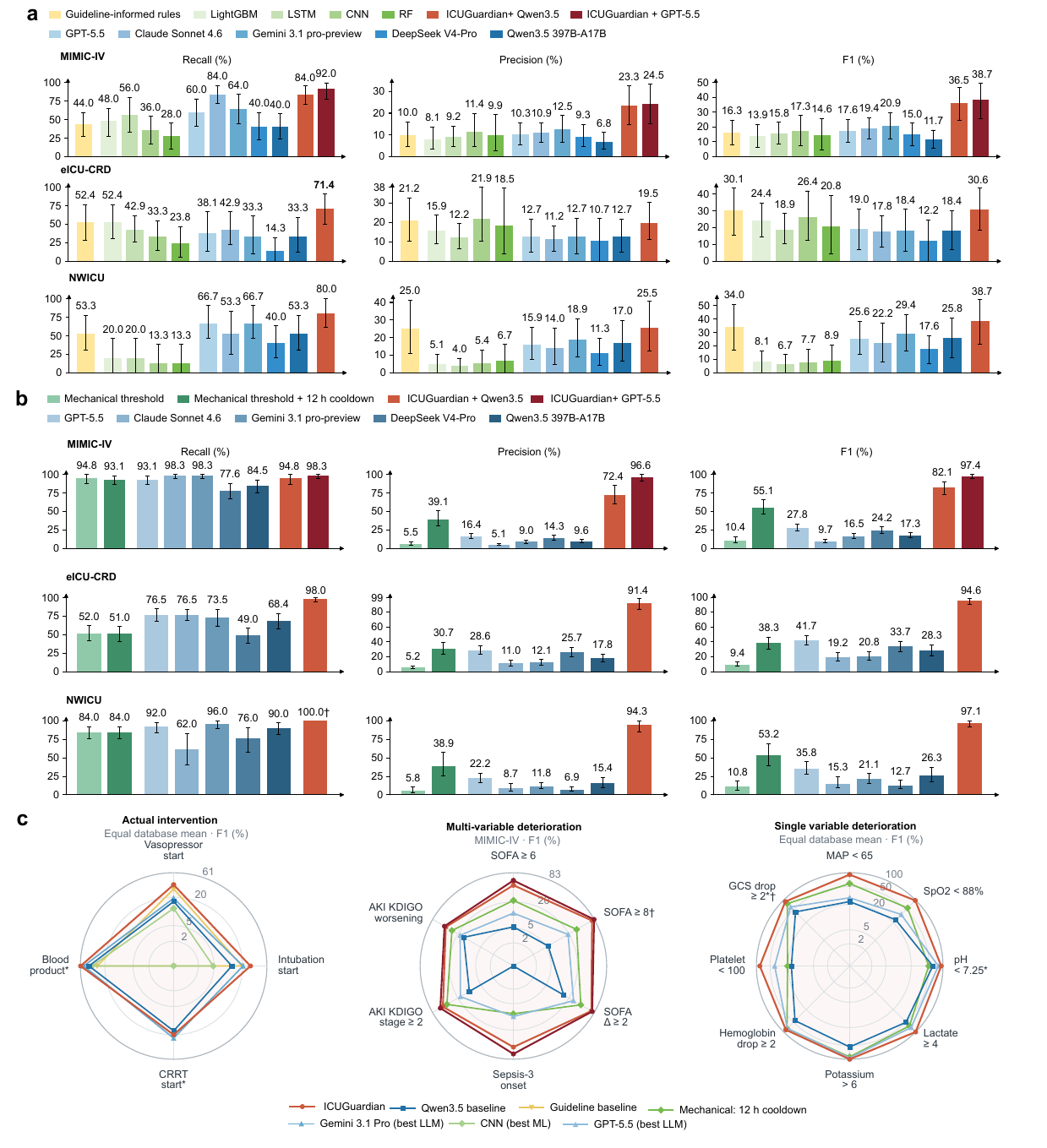}
\caption{\textbf{Database-specific detection performance and task-level F1 profiles.}
\textbf{a,b}, Recall, precision and F1 for intervention anticipation (\textbf{a}) and single-variable deterioration (\textbf{b}) across MIMIC-IV, eICU-CRD and NWICU. Dark red denotes \agentname{} + GPT-5.5 in MIMIC-IV (\textbf{a}, \textbf{b} and \textbf{c}, middle). Error bars show 95\% stay-bootstrap confidence intervals (omitted for 100\% recall in NWICU, $\dagger$).
\textbf{c}, Task-level F1 for intervention anticipation (left), multivariable deterioration (middle; MIMIC-IV only) and single-variable deterioration (right) on $\log(1+\mathrm{F1})$ radial scales. Left and right radars show equal-weighted means across eligible datasets ($^*$, two datasets); daggers ($\dagger$) mark subgroup tasks excluded from primary overall F1. Best family baselines were selected by overall F1 within each task family.}
\label{fig:app:database_detection}
\end{figure}

\begin{figure}[p]
\vspace{-16mm}
\centering
\includegraphics[width=\linewidth,trim=0 0 0 16bp,clip]{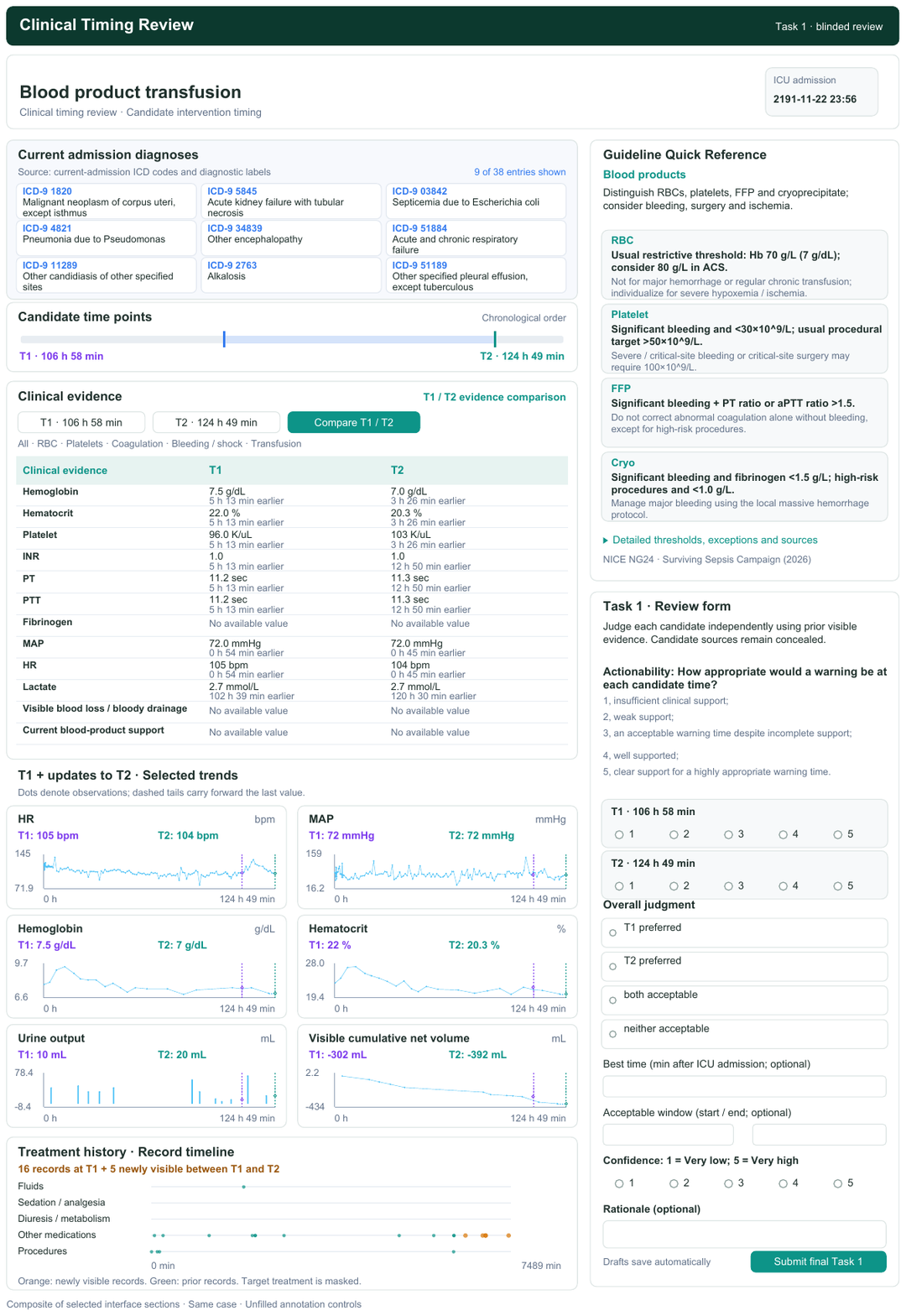}
\caption{\textbf{Interface for the clinician timing review.} Composite of
interface sections for one representative case, shown in the comparison view.
Left, current-admission diagnoses; the two candidate times (T1, T2) in
chronological order; the most recent target-relevant measurements at each time,
with the time elapsed since each measurement; selected trends; and a treatment
timeline in which records that became visible between T1 and T2 are highlighted
and the target treatment is masked. Separate T1 and T2 views show the same panels
truncated at each time. Right, a guideline quick reference for the target
intervention and the review form, comprising an actionability score for each
time, an overall judgement and optional fields. Interface text was translated
from Chinese.}
\label{fig:app:timing_interface}
\end{figure}

\begin{figure}[!htbp]
\centering
\includegraphics[width=\linewidth,pagebox=cropbox]{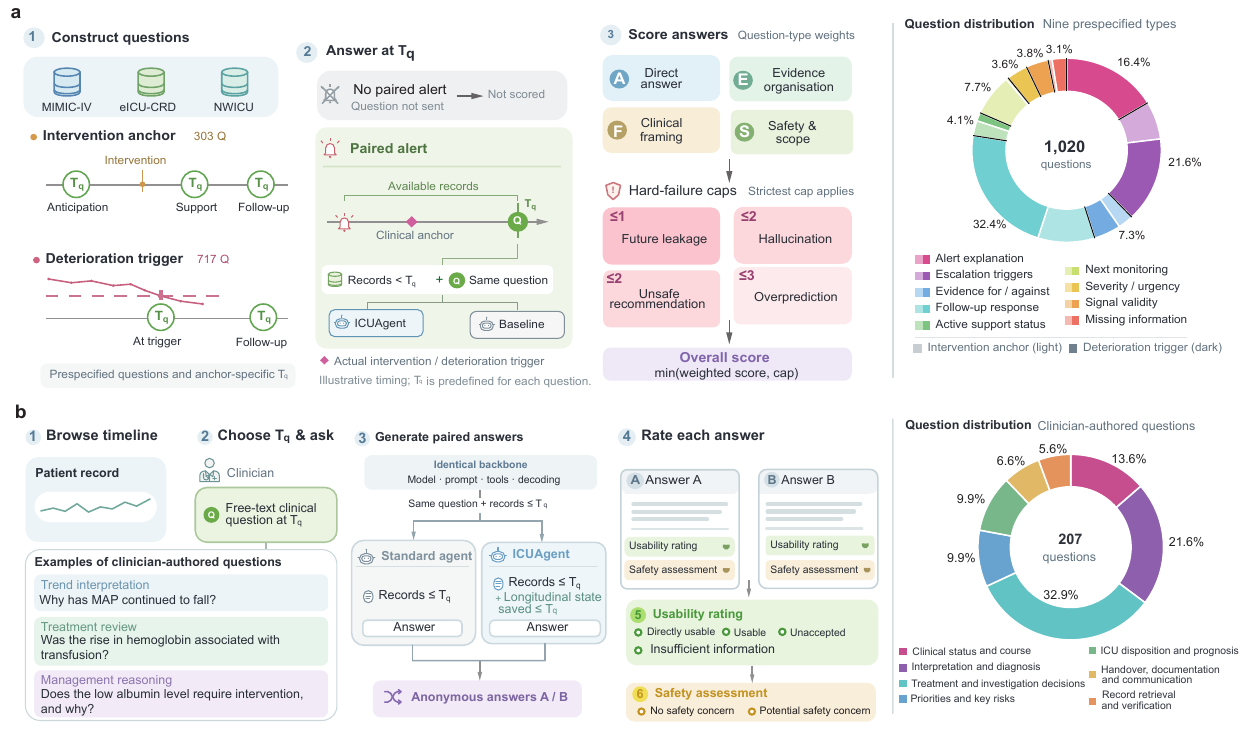}
\caption{\textbf{Workflows for automated clinical question answering and
the physician--AI interaction study in the \benchname{}.}
\textbf{a}, Automated clinical question answering; the doughnut chart shows the
nine question types among the 1,020 questions.
\textbf{b}, Physician--AI interaction study; the doughnut chart shows the 207
physician-authored questions across seven clinical topic categories.
Procedures are described in Methods.}
\label{fig:app:qa_workflow}
\end{figure}

\begin{figure}[p]
\vspace{-16mm}
\centering
\includegraphics[width=\linewidth]{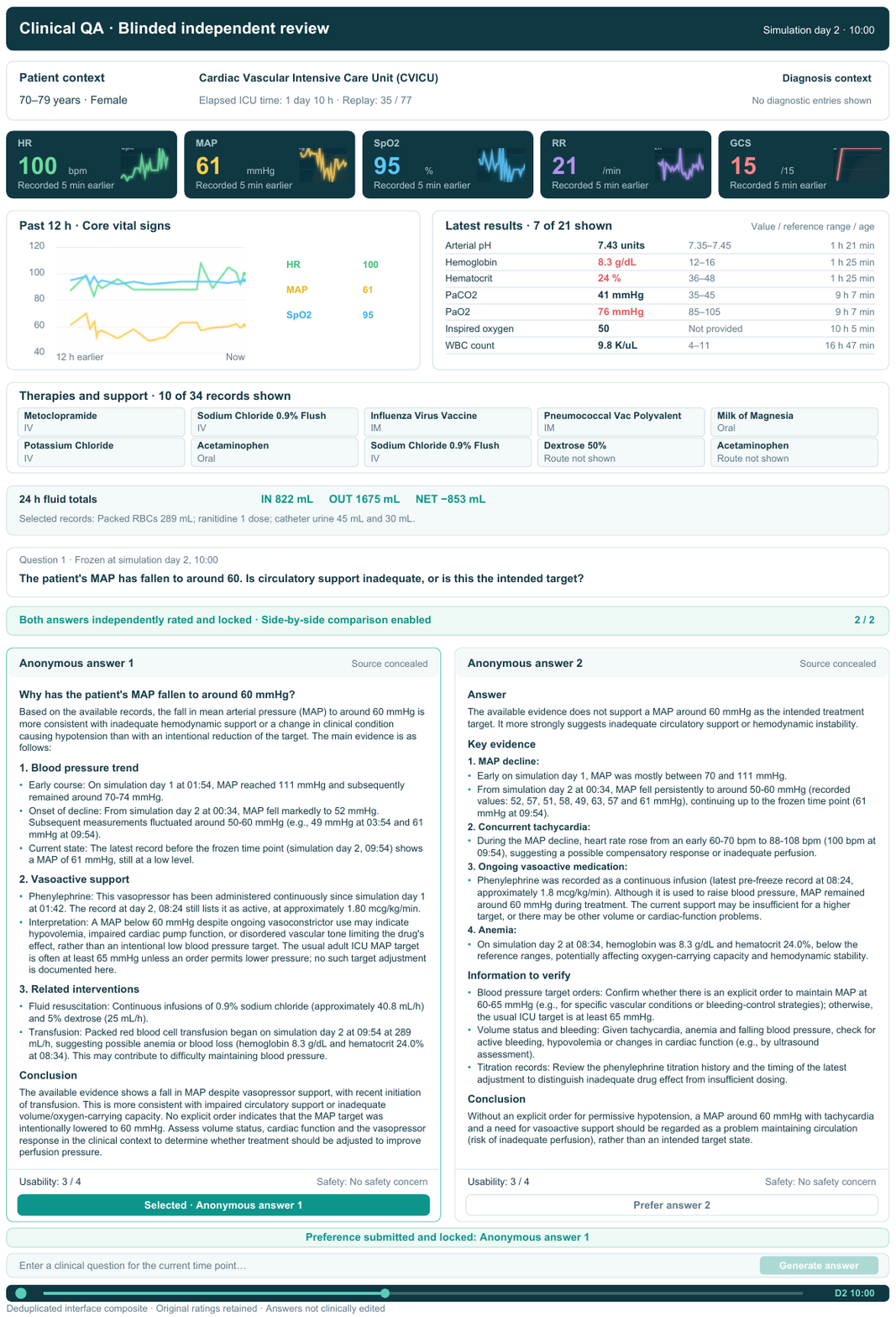}
\caption{\textbf{Interface for the physician--AI interaction study.}
Composite of interface sections for one question after both answers had been
rated. Physicians received written instructions covering case review, question
submission and answer assessment; questions were entered without prescribed
wording or question-type quotas, and replay remained frozen at the question time
until both answers had been assessed. Top, patient context and the time-gated
record: vital signs with the time since measurement, trends of core vital signs,
laboratory results with reference ranges, therapies and fluid totals. Middle,
the physician's question and the two anonymised answers in randomized order,
each with its usability rating and safety assessment, followed by the recorded
preference. Bottom, question entry and replay controls. Ratings are shown as
recorded. Interface text and answers were translated from Chinese without
clinical editing.}
\label{fig:app:qa_clinician_interface}
\end{figure}

\begin{figure}[!htbp]
\centering
\includegraphics[width=\linewidth]{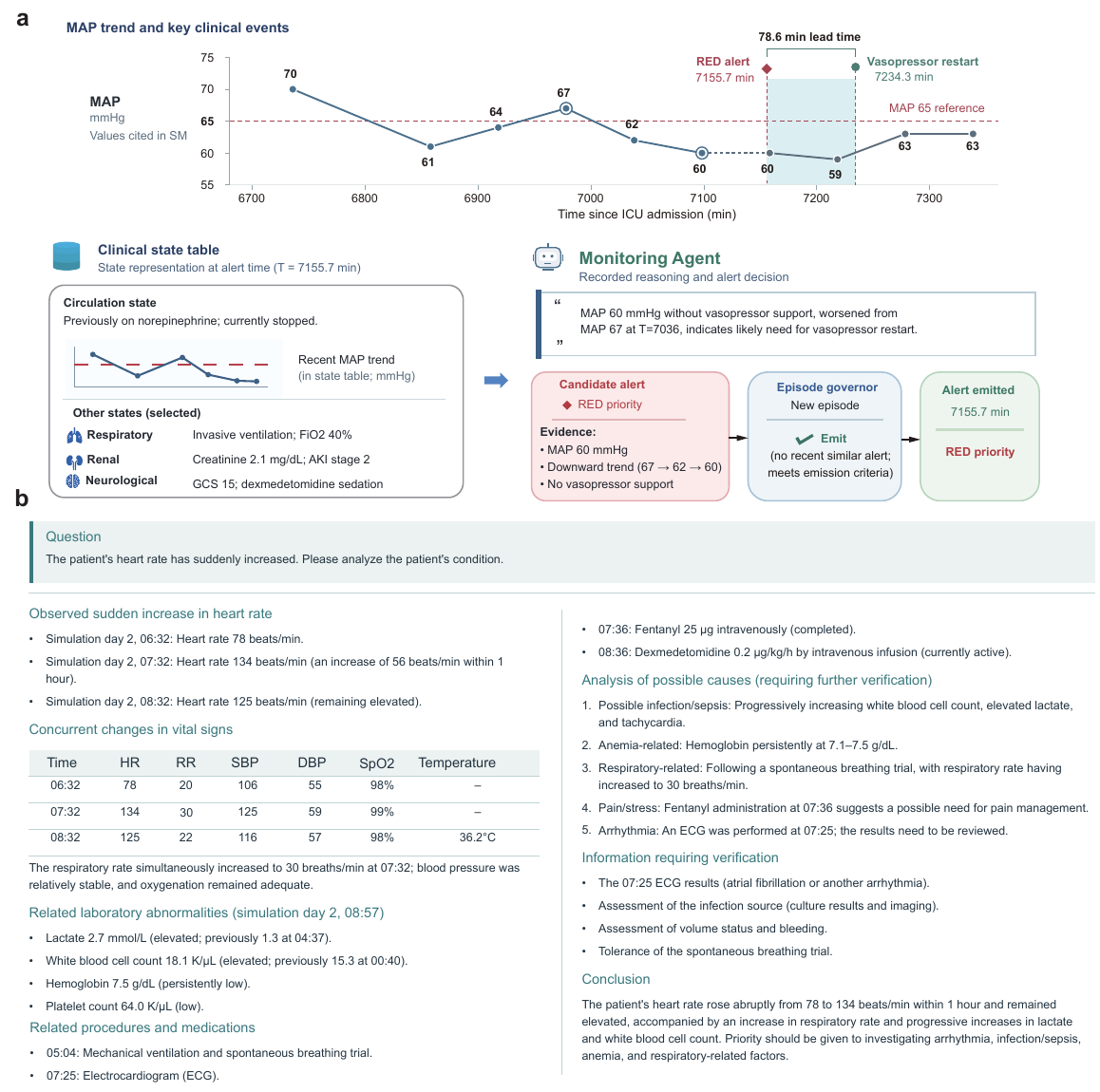}
\caption{\textbf{Illustrative longitudinal monitoring and clinical question-answering cases of \agentname{}.}
\textbf{a}, Mean arterial pressure (MAP) trajectory and recorded alert workflow for an illustrative vasopressor-restart alert in MIMIC-IV, showing the maintained clinical state at the alert time ($T = 7{,}155.7$\,min) and the recorded rationale and episode-governance decision of the Monitoring agent. Points show MAP values cited in state-maintenance (SM) records; connecting lines are for display only.
\textbf{b}, \agentname{} response to a physician's question about a sudden rise in heart rate during replay.}
\label{fig:case_analysis}
\end{figure}

\begin{figure}[!htbp]
\centering
\includegraphics[width=\linewidth]{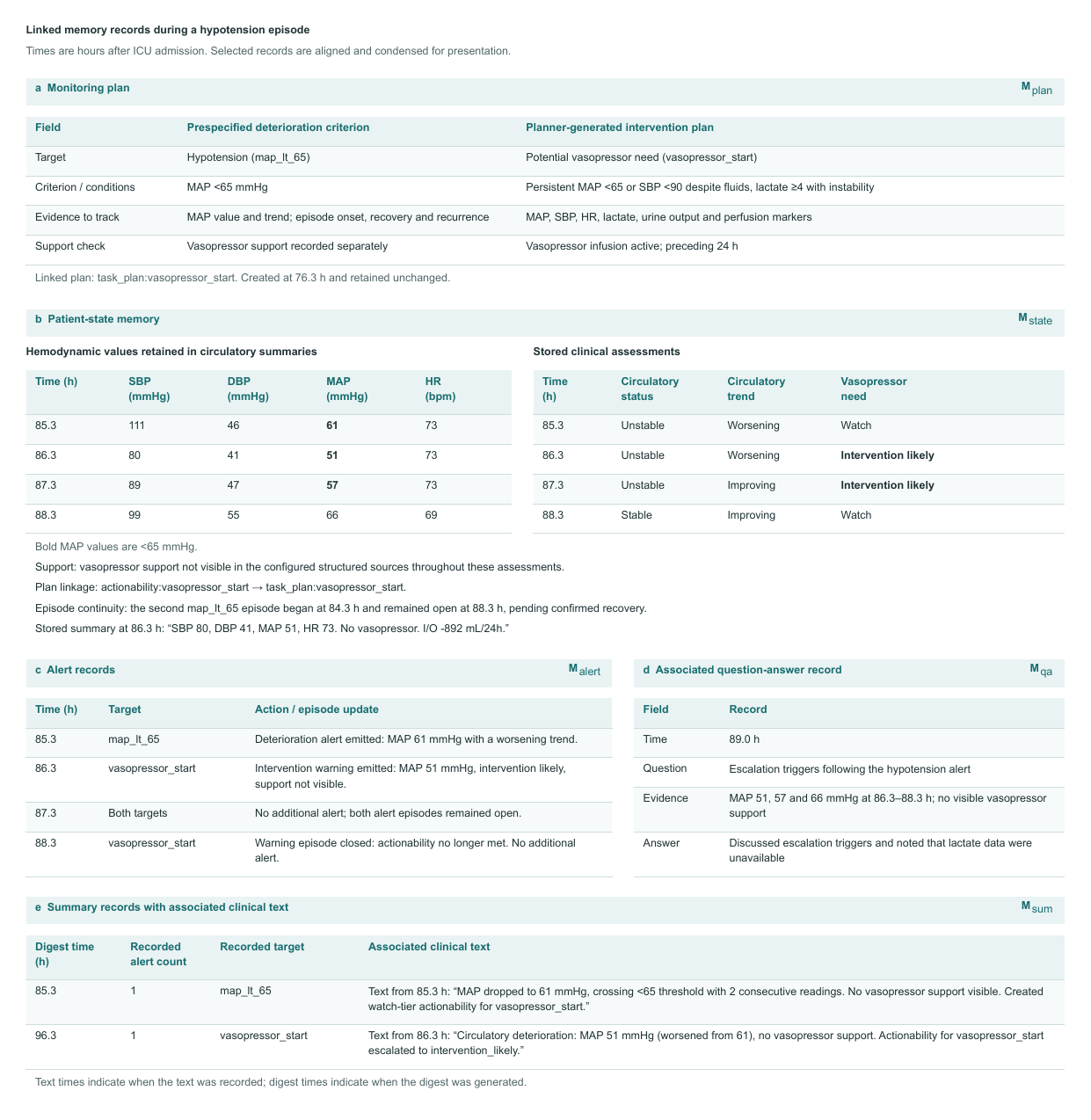}
\caption{\textbf{Linked memory records for one MIMIC-IV test stay during a
hypotension episode (85--89\,h after ICU admission).}
Records are linked by the monitoring-target index \texttt{vasopressor\_start}
and by timestamps.
\textbf{a}, Monitoring plan ($\mathcal{M}_{\mathrm{plan}}$): one prespecified
deterioration criterion and one Planner-generated intervention plan
(Supplementary Note~\ref{app:icuagent_planner_prompt}), approved at
self-reflection.
\textbf{b}, Patient-state memory ($\mathcal{M}_{\mathrm{state}}$): retained
haemodynamic values and stored clinical assessments.
\textbf{c}, Alert records ($\mathcal{M}_{\mathrm{alert}}$).
\textbf{d}, Associated question--answer record ($\mathcal{M}_{\mathrm{qa}}$),
obtained by replaying the saved state at the question time.
\textbf{e}, Summary records ($\mathcal{M}_{\mathrm{sum}}$), shown alongside
clinical text recorded at the indicated times. Times are hours after ICU
admission; records are reproduced from the run log, aligned and condensed for
presentation.}
\label{fig:app:memory_example}
\end{figure}

\FloatBarrier
\clearpage


\section*{Supplementary Tables}

\begingroup
\small
\setlength{\tabcolsep}{4pt}
\begin{longtable}{@{}>{\raggedright\arraybackslash}p{\dimexpr0.40\linewidth-3.2\tabcolsep\relax}*{4}{>{\centering\arraybackslash}p{\dimexpr0.15\linewidth-1.2\tabcolsep\relax}}@{}}
\caption{Characteristics and reference-event counts of the frozen test cohort ($n=100$). Values are median [IQR] for continuous variables and $n$ (\%) for categorical variables unless otherwise indicated. Reference-event counts include the 48 subgroup events excluded from the 442 primary reference events (Methods).}\label{tab:cohort}\\
\toprule
Characteristic & \makecell{MIMIC-IV\\($n=50$)} & \makecell{eICU-CRD\\($n=25$)} & \makecell{NWICU\\($n=25$)} & \makecell{Overall\\($n=100$)} \\
\midrule
\endfirsthead
\multicolumn{5}{l}{\small\itshape Continued from previous page}\\
\toprule
Characteristic & MIMIC-IV & eICU-CRD & NWICU & Overall \\
\midrule
\endhead
\midrule
\multicolumn{5}{r}{\small\itshape Continued on next page}\\
\endfoot
\bottomrule
\endlastfoot
\multicolumn{5}{l}{\textit{Demographics}} \\
ICU stays, $n$ & 50 & 25 & 25 & 100 \\
Unique patients, $n$ & 50 & 25 & 25 & 100 \\
Age, years & 68\newline [55--82] & 70\newline [61--78] & 69\newline [65--74] & 70\newline [57--78] \\
Female & 28 (56) & 10 (40) & 11 (44) & 49 (49) \\
\midrule
\multicolumn{5}{l}{\textit{Admission and ICU context}} \\
Admission source available & 50 (100) & 25 (100) & 22 (88) & 97 (97) \\
Admission from emergency department & 25 (50) & 14 (56) & 17 (68) & 56 (56) \\
Medical or mixed ICU & 15 (30) & 20 (80) & 17 (68) & 52 (52) \\
Cardiovascular or cardiothoracic ICU & 17 (34) & 3 (12) & 6 (24) & 26 (26) \\
Surgical or trauma ICU & 14 (28) & 2 (8) & 0 (0) & 16 (16) \\
Neurological or neurosurgical ICU & 4 (8) & 0 (0) & 2 (8) & 6 (6) \\
\midrule
\multicolumn{5}{l}{\textit{ICU stay and simulation}} \\
ICU length of stay, h & 110.8\newline [89.6--143.8] & 109.9\newline [92.5--146.6] & 116.3\newline [89.3--156.5] & 113.4\newline [89.8--147.0] \\
\quad $d_{3\text{--}4}$ (3--4 d) & 18 (36) & 9 (36) & 9 (36) & 36 (36) \\
\quad $d_{4\text{--}7}$ (4--7 d) & 23 (46) & 11 (44) & 11 (44) & 45 (45) \\
\quad $d_{7\text{--}10}$ (7--10 d) & 9 (18) & 5 (20) & 5 (20) & 19 (19) \\
Replay window, h & 72.0\newline [72.0--77.4] & 72.0\newline [72.0--78.3] & 72.0\newline [72.0--77.7] & 72.0\newline [72.0--77.7] \\
Replayed over final 72\,h & 32 (64) & 16 (64) & 16 (64) & 64 (64) \\
\midrule
\multicolumn{5}{l}{\textit{Reference events}} \\
Reference events, $n$ & 292 & 133 & 65 & 490 \\
\end{longtable}
\endgroup

\begingroup
\footnotesize
\setlength{\tabcolsep}{3pt}
\renewcommand{\arraystretch}{1.15}
\begin{longtable}{@{}
>{\raggedright\arraybackslash}p{0.15\linewidth}
>{\raggedright\arraybackslash}p{0.29\linewidth}
>{\raggedright\arraybackslash}p{0.25\linewidth}
>{\raggedright\arraybackslash}p{\dimexpr0.31\linewidth-6\tabcolsep\relax}@{}}
\caption{Prespecified rules for candidate onset times and clinical
review of intervention episodes. In addition to the exclusions in Methods,
uncertain intervention records, episodes without an identifiable warning
opportunity and borderline vasopressor episodes were excluded. All search
windows precede the observed treatment start; thresholds identify candidate
evidence for review and do not by themselves establish a clinical indication.
Confirmed onsets were recorded as a time point or an interval, and each
reference warning interval $[T_{\mathrm{onset,lower}}, T_{\mathrm{int}})$ ran
from the earliest confirmed onset to treatment initiation.}
\label{tab:intervention_onset_rules}\\
\toprule
\textbf{Intervention / branch} &
\textbf{Candidate evidence and search window} &
\textbf{Candidate onset time} &
\textbf{Clinical review focus} \\
\midrule
\endfirsthead
\multicolumn{4}{l}{\small\itshape Continued from previous page}\\
\toprule
\textbf{Intervention / branch} &
\textbf{Candidate evidence and search window} &
\textbf{Candidate onset time} &
\textbf{Clinical review focus} \\
\midrule
\endhead
\midrule
\multicolumn{4}{r}{\small\itshape Continued on next page}\\
\endfoot
\bottomrule
\endlastfoot
\textbf{Vasopressor initiation} &
Within 6\,h: MAP $\leq65$\,mmHg; if no MAP value qualifies,
use SBP $\leq90$\,mmHg. Lactate $\geq2$\,mmol/L within 12\,h provides
supporting evidence. &
Group hypotensive observations separated by gaps of $\leq3$\,h.
Select the most recent group and use its first abnormal observation. &
Determine whether this is a new hypotensive or hypoperfusion episode
related to the treatment, considering fluid response, existing vasopressor
support and anaesthetic or procedural context. \\
\addlinespace
\textbf{Intubation} &
Within 6\,h: SpO$_2$ $\leq92$\%, RR $\geq25$/min or GCS $\leq12$.
Strong evidence: SpO$_2$ $\leq90$\%, RR $\geq30$/min, GCS $\leq8$,
PaO$_2$ $\leq60$\,mmHg, pH $\leq7.25$ or PaCO$_2$ $\geq50$\,mmHg. &
Group abnormal observations separated by gaps of $\leq3$\,h and
select the most recent group. Use its first abnormal observation if strong
evidence is present within 6\,h or the group contains at least two abnormal
observations. &
Identify a new oxygenation, ventilation or airway-protection problem;
distinguish acute new intubation from an existing airway, documentation of
ventilation status and a procedural airway. \\
\addlinespace
\textbf{Red-cell transfusion} &
Within 24\,h: Hb $\leq8.5$\,g/dL. MAP $\leq65$\,mmHg or
HR $\geq110$/min within 6\,h provides haemodynamic supporting evidence. &
Use the most recent laboratory time meeting the Hb criterion. &
Assess whether anaemia, bleeding or related symptoms explain this
transfusion in its clinical context. Hypotension alone does not establish
a red-cell transfusion indication. \\
\addlinespace
\textbf{Plasma transfusion} &
Within 24\,h: INR $\geq1.5$, PT $\geq18$\,s or fibrinogen
$\leq150$\,mg/dL. &
Use the earliest laboratory time meeting any criterion and flag
as biochemical evidence only. &
Look for active bleeding, a procedure or anticoagulation reversal.
Laboratory abnormalities alone do not confirm an evaluable clinical need. \\
\addlinespace
\textbf{Platelet transfusion} &
Within 24\,h: platelet count $\leq100\times10^9$/L. &
Use the most recent qualifying laboratory time and flag as
biochemical evidence only. &
Consider the degree of thrombocytopenia, bleeding site and procedural
context. The screening threshold is not itself a transfusion indication. \\
\addlinespace
\textbf{Cryoprecipitate and other blood products} &
No separate automated candidate rule is defined. Review
product-specific laboratory, bleeding and procedural evidence before
transfusion. &
Determine the first relevant clinical evidence through individual
case review; assign no candidate time if evidence is insufficient. &
Verify the product and its therapeutic purpose, and whether
coagulopathy, bleeding or procedural context explains the transfusion. \\
\addlinespace
\textbf{CRRT: metabolic abnormalities} &
Within 24\,h: potassium $\geq6$\,mmol/L, pH $\leq7.20$ or
bicarbonate $\leq15$\,mmol/L. &
Use the earliest laboratory time meeting any metabolic criterion. &
Determine whether the abnormality relates to this renal-support
need, considering persistence, existing treatment, renal function and other
clinical context. \\
\addlinespace
\textbf{CRRT: renal dysfunction} &
No metabolic criterion above is met, but within 24\,h creatinine
is $\geq4$\,mg/dL or BUN is $\geq80$\,mg/dL. &
Use the earliest laboratory time meeting either renal criterion
and flag possible left censoring for review. &
Review prior renal function, uraemia, volume overload and existing
dialysis. Creatinine or BUN elevation alone does not establish a new
CRRT need. \\
\addlinespace
\textbf{No qualifying evidence} &
No candidate evidence is available within the relevant window, or
the available evidence does not explain the treatment. &
Assign no candidate time, flag the indication as unclear and refer
for individual case review. &
Distinguish absent clinical evidence from missing data and record
the additional information needed. \\
\end{longtable}
\noindent\footnotesize BUN, blood urea nitrogen; CRRT, continuous
renal replacement therapy; GCS, Glasgow Coma Scale; Hb, haemoglobin; HR, heart
rate; INR, international normalised ratio; MAP, mean arterial pressure;
PaCO$_2$, arterial partial pressure of carbon dioxide; PaO$_2$, arterial
partial pressure of oxygen; PT, prothrombin time; RR, respiratory rate;
SBP, systolic blood pressure; SpO$_2$, peripheral oxygen saturation.
\par
\endgroup

\begin{table}[!hbp]
\centering
\footnotesize
\caption{\textbf{Guideline-informed rules.} Branch identifiers follow the rule
configuration. $^\dagger$Approximated from structured records: maximal oxygenation, NIV
or HFNC with FiO$_2$ 1.0 within 60\,min before each of two SpO$_2$ values; medical
therapy for hyperkalaemia, intravenous insulin labelled for hyperkalaemia between two
potassium values $>$6.5\,mmol/L; haemodynamic instability, two most recent MAP values
$<$65\,mmHg or ongoing vasopressor support; clinically significant bleeding,
time-stamped admission-diagnosis labels (eICU-CRD only). $^\ddagger$No structured
mapping; treated as unknown.}
\label{tab:app:guideline_rules}
\begin{tabular}{>{\RaggedRight\arraybackslash}p{2.6cm}>{\RaggedRight\arraybackslash}p{10.2cm}>{\RaggedRight\arraybackslash}p{1.6cm}}
\toprule
Target (branches) & Conditions & Source \\
\midrule
Vasopressor (V01--V02) & Sepsis-related hypotension after initial crystalloid
  resuscitation$^\ddagger$; or shock-related branch with the two most recent MAP
  values $<$65\,mmHg (diagnostic waiver applied)
  & \cite{prescott2026surviving} \\
IMV (I01--I02) & Hypoxaemic cardiac or respiratory arrest$^\ddagger$; or severe
  hypoxaemia despite maximal oxygenation$^\dagger$ with PaO$_2$/FiO$_2$ $<$60 or
  SpO$_2$ $<$88\% & \cite{helms2024oxygen} \\
IMV (I03--I13) & Shock with vasopressor support and two most recent MAP
  $<$65\,mmHg; respiratory or mixed acidosis with pH $<$7.3; repeated RR
  $>$30/min; repeated SpO$_2$ $<$86\%; worsening or persistent hypoxaemia despite
  oxygenation$^\dagger$; respiratory distress, new or worsening altered
  consciousness, retained secretions, agitation or intolerance of the
  oxygenation modality$^\ddagger$ & \cite{helms2024oxygen} \\
IMV (I14--I18) & Acute exacerbation of COPD (no diagnostic waiver) with impending
  respiratory arrest or severe distress$^\ddagger$, NIV failure or
  contraindication, persistent severe acidosis or falling pH after NIV, or
  explained GCS $<$8 & \cite{davidson2016bts} \\
CRRT (K01--K04) & RRT need (AKI with life-threatening fluid, electrolyte or
  acid--base disturbance; potassium $>$6.5\,mmol/L despite medical
  therapy$^\dagger$; or pulmonary oedema with persistent hypoxaemia) plus
  haemodynamic instability$^\dagger$
  & \cite{kdigo2012aki,jourdain2025rrt} \\
Red cells (B01--B05) & Hb $<$7 (stable adults), $<$7.5 (cardiac surgery), $<$8
  (orthopaedic surgery or pre-existing cardiovascular disease) or $<$10\,g/dL
  (acute myocardial infarction) & \cite{carson2023red,pagano2025ami} \\
Platelets (B06--B12) & Platelets $<$10, $<$20 or $<$50$\times10^9$/L by population,
  bleeding or planned procedure$^\ddagger$ & \cite{metcalf2025platelet} \\
Plasma (B13--B14) & Clinically significant bleeding$^\dagger$ with abnormal
  coagulation ratio; or high-bleeding-risk procedure$^\ddagger$ with coagulopathy
  & \cite{nice2015blood} \\
Cryoprecipitate (B15--B16) & Clinically significant bleeding$^\dagger$ with
  fibrinogen $<$1.5\,g/L; or high-bleeding-risk procedure$^\ddagger$ with
  fibrinogen $<$1.0\,g/L & \cite{nice2015blood} \\
\bottomrule
\end{tabular}
\end{table}

\begin{table}[!htbp]
\centering
\caption{Pooled detection performance and alert counts for all methods. Values are percentages with 95\% bootstrap confidence intervals from 20,000 ICU-stay resamples, stratified by database for pooled estimates. Matched/reference gives detected reference episodes or events over eligible ones. Prediction episodes apply to intervention anticipation only (Methods).}
\label{tab:app:pooled_performance}
\resizebox{\textwidth}{!}{%
\begin{tabular}{l r r r c c c}
\toprule
Method & Raw alerts & Episodes & Matched/reference & Recall (\%) & Precision (\%) & F1 (\%) \\
\midrule
\multicolumn{7}{l}{\emph{Intervention anticipation, three databases pooled (100 stays; 61 reference episodes)}} \\[2pt]
Guideline-informed rules & 1,285 & 194 & 30/61 & 49.2 (37.3--61.3) & 15.5 (10.5--20.8) & 23.5 (16.7--30.3) \\
LightGBM & 1,874 & 276 & 26/61 & 42.6 (30.2--55.6) & 9.4 (6.2--13.0) & 15.4 (10.4--20.6) \\
LSTM & 3,337 & 301 & 26/61 & 42.6 (30.2--56.0) & 8.6 (5.7--11.7) & 14.4 (9.8--19.0) \\
CNN & 667 & 148 & 18/61 & 29.5 (18.5--41.7) & 12.2 (7.3--17.6) & 17.2 (10.6--24.0) \\
Random forest & 498 & 128 & 14/61 & 23.0 (12.1--35.1) & 10.9 (5.4--17.5) & 14.8 (7.5--22.9) \\
GPT-5.5 & 483 & 271 & 33/61 & 54.1 (41.2--67.4) & 12.2 (8.4--16.2) & 19.9 (14.1--25.6) \\
Claude Sonnet 4.6 & 3,253 & 329 & 38/61 & 62.3 (50.0--74.6) & 11.6 (8.1--15.2) & 19.5 (14.1--24.8) \\
Gemini 3.1-Pro-Preview & 951 & 236 & 33/61 & 54.1 (41.0--67.7) & 14.0 (9.6--18.7) & 22.2 (15.8--28.7) \\
DeepSeek-V4-Pro & 499 & 189 & 19/61 & 31.1 (20.8--42.0) & 10.1 (6.3--14.2) & 15.2 (9.8--20.7) \\
Qwen3.5-397B-A17B & 876 & 248 & 25/61 & 41.0 (29.3--53.7) & 10.1 (6.5--14.0) & 16.2 (10.8--21.7) \\
\agentname{} + Qwen3.5 & 271 & 214 & 48/61 & 78.7 (68.6--88.3) & 22.4 (16.7--28.6) & 34.9 (27.4--42.3) \\
\midrule
\multicolumn{7}{l}{\emph{Multivariable deterioration, MIMIC-IV (50 stays; 175 reference events)}} \\[2pt]
Mechanical threshold & 6,710 & --- & 148/175 & 84.6 (77.6--90.9) & 2.2 (1.8--2.8) & 4.3 (3.4--5.4) \\
Mechanical threshold + 12\,h cooldown & 714 & --- & 122/175 & 69.7 (61.7--77.5) & 17.1 (13.9--20.9) & 27.4 (22.9--32.4) \\
GPT-5.5 & 193 & --- & 34/175 & 19.4 (13.4--26.1) & 17.6 (12.4--23.5) & 18.5 (13.1--24.1) \\
Claude Sonnet 4.6 & 350 & --- & 21/175 & 12.0 (7.2--17.3) & 6.0 (3.9--8.8) & 8.0 (5.2--11.2) \\
Gemini 3.1-Pro-Preview & 408 & --- & 39/175 & 22.3 (15.9--28.9) & 9.6 (6.2--14.6) & 13.4 (9.3--18.4) \\
DeepSeek-V4-Pro & 66 & --- & 10/175 & 5.7 (2.2--10.0) & 15.2 (6.3--25.4) & 8.3 (3.3--13.9) \\
Qwen3.5-397B-A17B & 227 & --- & 23/175 & 13.1 (7.7--19.3) & 10.1 (5.8--16.4) & 11.4 (6.8--16.8) \\
\agentname{} + Qwen3.5 & 319 & --- & 125/175 & 71.4 (64.4--78.2) & 39.2 (32.7--46.6) & 50.6 (44.0--57.5) \\
\midrule
\multicolumn{7}{l}{\emph{Single-variable deterioration, three databases pooled (100 stays; 206 reference events)}} \\[2pt]
Mechanical threshold & 2,707 & --- & 148/206 & 71.8 (66.0--77.4) & 5.5 (4.3--7.0) & 10.2 (8.2--12.8) \\
Mechanical threshold + 12\,h cooldown & 409 & --- & 146/206 & 70.9 (65.2--76.4) & 35.7 (30.0--42.3) & 47.5 (41.8--53.5) \\
GPT-5.5 & 799 & --- & 175/206 & 85.0 (80.2--89.5) & 21.9 (19.3--24.8) & 34.8 (31.3--38.6) \\
Claude Sonnet 4.6 & 2,159 & --- & 163/206 & 79.1 (72.2--85.6) & 7.5 (6.2--9.2) & 13.8 (11.5--16.6) \\
Gemini 3.1-Pro-Preview & 1,634 & --- & 177/206 & 85.9 (80.0--91.3) & 10.8 (9.1--12.8) & 19.2 (16.4--22.4) \\
DeepSeek-V4-Pro & 1,048 & --- & 131/206 & 63.6 (56.4--70.3) & 12.5 (9.9--15.9) & 20.9 (17.1--25.5) \\
Qwen3.5-397B-A17B & 1,177 & --- & 161/206 & 78.2 (71.9--84.0) & 13.7 (11.4--16.3) & 23.3 (19.8--27.1) \\
\agentname{} + Qwen3.5 & 234 & --- & 201/206 & 97.6 (94.9--99.5) & 85.9 (79.8--91.6) & 91.4 (87.6--94.7) \\
\bottomrule
\end{tabular}%
}
\end{table}

\begin{table}[!htbp]
\centering
\caption{Paired differences between \agentname{} and each comparator. Differences are \agentname{} minus comparator, in percentage points, with 95\% confidence intervals from 20,000 paired ICU-stay bootstrap resamples (stratified by database for pooled analyses). $P$ values are from two-sided paired label-swap tests clustered by ICU stay (exact tests for machine-learning models; 20,000 permutations otherwise), Holm-adjusted across the 51 pooled comparisons for intervention anticipation and single-variable deterioration and across the 174 database-level comparisons for multivariable deterioration. Comparisons with guideline-informed rules applied a common support-status gate to both methods, under which \agentname{} recall, precision and F1 were 77.0\%, 22.3\% and 34.6\%. }
\label{tab:app:paired_differences}
\resizebox{\textwidth}{!}{%
\begin{tabular}{l c c c c c c}
\toprule
Comparator & $\Delta$ Recall & $P$ & $\Delta$ Precision & $P$ & $\Delta$ F1 & $P$ \\
\midrule
\multicolumn{7}{l}{\emph{Intervention anticipation, three databases pooled (100 stays; 61 reference episodes)}} \\[2pt]
Guideline-informed rules & 27.9 (14.7 to 41.9) & 0.004 & 6.8 (2.3 to 11.6) & 0.018 & 11.0 (4.6 to 17.7) & 0.015 \\
LightGBM & 36.1 (20.0 to 51.7) & 0.004 & 13.0 (8.0 to 18.5) & $<$0.001 & 19.5 (12.3 to 26.7) & $<$0.001 \\
LSTM & 36.1 (19.7 to 51.4) & 0.004 & 13.8 (8.6 to 19.3) & $<$0.001 & 20.5 (13.3 to 27.6) & $<$0.001 \\
CNN & 49.2 (34.5 to 62.5) & $<$0.001 & 10.3 (3.9 to 16.9) & 0.018 & 17.7 (8.9 to 26.5) & 0.005 \\
Random forest & 55.7 (42.5 to 68.5) & $<$0.001 & 11.5 (4.8 to 18.1) & 0.015 & 20.1 (11.1 to 28.7) & 0.002 \\
GPT-5.5 & 24.6 (14.5 to 35.1) & 0.004 & 10.3 (6.9 to 14.2) & 0.002 & 15.0 (10.5 to 20.1) & 0.002 \\
Claude Sonnet 4.6 & 16.4 (3.3 to 29.7) & 0.039 & 10.9 (7.1 to 15.3) & 0.002 & 15.4 (10.1 to 21.0) & 0.002 \\
Gemini 3.1-Pro-Preview & 24.6 (13.0 to 36.2) & 0.006 & 8.4 (4.8 to 12.6) & 0.002 & 12.7 (7.4 to 18.2) & 0.002 \\
DeepSeek-V4-Pro & 47.5 (33.3 to 61.4) & 0.002 & 12.4 (7.7 to 17.6) & 0.002 & 19.7 (12.8 to 26.8) & 0.002 \\
Qwen3.5-397B-A17B & 37.7 (25.5 to 50.0) & 0.002 & 12.3 (8.2 to 17.0) & 0.002 & 18.7 (12.9 to 24.9) & 0.002 \\
\midrule
\multicolumn{7}{l}{\emph{Multivariable deterioration, MIMIC-IV (50 stays; 175 reference events)}} \\[2pt]
Mechanical threshold & $-$13.1 ($-$18.5 to $-$8.3) & 0.009 & 37.0 (30.8 to 44.0) & 0.009 & 46.3 (40.2 to 52.5) & 0.009 \\
Mechanical threshold + 12\,h cooldown & 1.7 ($-$2.9 to 6.2) & 1.000 & 22.1 (17.9 to 26.7) & 0.009 & 23.2 (19.3 to 26.9) & 0.009 \\
GPT-5.5 & 52.0 (43.7 to 60.3) & 0.009 & 21.6 (14.9 to 28.9) & 0.009 & 32.1 (24.5 to 40.0) & 0.009 \\
Claude Sonnet 4.6 & 59.4 (51.4 to 67.4) & 0.009 & 33.2 (26.7 to 40.4) & 0.009 & 42.6 (35.6 to 49.9) & 0.009 \\
Gemini 3.1-Pro-Preview & 49.1 (41.0 to 57.7) & 0.009 & 29.6 (24.6 to 35.0) & 0.009 & 37.2 (31.6 to 43.2) & 0.009 \\
DeepSeek-V4-Pro & 65.7 (58.1 to 73.2) & 0.009 & 24.0 (12.5 to 35.4) & 0.009 & 42.3 (33.7 to 51.1) & 0.009 \\
Qwen3.5-397B-A17B & 58.3 (49.7 to 66.9) & 0.009 & 29.1 (23.0 to 35.4) & 0.009 & 39.2 (32.3 to 46.4) & 0.009 \\
\midrule
\multicolumn{7}{l}{\emph{Single-variable deterioration, three databases pooled (100 stays; 206 reference events)}} \\[2pt]
Mechanical threshold & 25.7 (19.6 to 32.0) & 0.002 & 80.4 (74.9 to 85.6) & 0.002 & 81.2 (77.8 to 84.2) & 0.002 \\
Mechanical threshold + 12\,h cooldown & 26.7 (20.6 to 32.8) & 0.002 & 50.2 (44.3 to 55.8) & 0.002 & 43.9 (38.6 to 49.0) & 0.002 \\
GPT-5.5 & 12.6 (7.5 to 17.9) & 0.002 & 64.0 (58.4 to 69.3) & 0.002 & 56.5 (52.3 to 60.7) & 0.002 \\
Claude Sonnet 4.6 & 18.4 (11.4 to 25.8) & 0.002 & 78.3 (72.5 to 83.7) & 0.002 & 77.6 (73.6 to 81.1) & 0.002 \\
Gemini 3.1-Pro-Preview & 11.7 (5.3 to 18.4) & 0.015 & 75.1 (69.5 to 80.2) & 0.002 & 72.1 (68.2 to 75.7) & 0.002 \\
DeepSeek-V4-Pro & 34.0 (27.1 to 41.2) & 0.002 & 73.4 (67.4 to 78.8) & 0.002 & 70.5 (65.5 to 74.8) & 0.002 \\
Qwen3.5-397B-A17B & 19.4 (13.1 to 26.0) & 0.002 & 72.2 (66.7 to 77.3) & 0.002 & 68.1 (63.9 to 71.9) & 0.002 \\
\bottomrule
\end{tabular}%
}
\end{table}

\begin{table}[!htbp]
\centering
\caption{Component ablation of \agentname{} against the Qwen baseline on the 50
MIMIC-IV test stays (25 intervention reference episodes, 175 multivariable and
58 single-variable deterioration events). All variants used the
Qwen3.5-397B-A17B backbone under the same replay clock and evaluation protocol.
PromptGov replaced deterministic episode governance with prompt-guided alert
decisions; No-SM removed the state-maintenance stage, so that the Monitoring
agent queried the time-gated tools directly; and NoPlanner removed the Planner,
its monitoring plan and the prespecified deterioration criteria. Boldface
denotes the highest recall, precision and F1 within each task, including ties;
alert counts are not ranked.}
\label{tab:e4_component_ablation_qwen}
\begin{threeparttable}
\small
\begin{tabular}{lrrrr}
\toprule
System & Recall (\%) & Precision (\%) & F1 (\%) & Raw alerts \\
\midrule
\multicolumn{5}{l}{\emph{Intervention anticipation}} \\[2pt]
Qwen baseline             & 40.0 &  6.8 & 11.7 & 473 \\
\agentname{} (full)        & \textbf{84.0} & \textbf{23.3} & \textbf{36.5} & 104 \\
\agentname{}--PromptGov     & 72.0 & 21.7 & 33.3 & 139 \\
\agentname{}--No-SM         & 52.0 & 10.9 & 18.1 & 152 \\
\agentname{}--NoPlanner     & 36.0 & 10.8 & 16.7 &  88 \\
\midrule
\multicolumn{5}{l}{\emph{Multivariable deterioration}} \\[2pt]
Qwen baseline             & 13.1 & 10.1 & 11.4 & 227 \\
\agentname{} (full)         & \textbf{71.4} & \textbf{39.2} & \textbf{50.6} & 319 \\
\agentname{}--PromptGov     & 48.0 & 34.1 & 39.9 & 246 \\
\agentname{}--No-SM         & 23.4 & 19.9 & 21.5 & 206 \\
\agentname{}--NoPlanner     & 27.4 & 18.6 & 22.2 & 258 \\
\midrule
\multicolumn{5}{l}{\emph{Single-variable deterioration}} \\[2pt]
Qwen baseline             & 84.5 &  9.6 & 17.3 & 509 \\
\agentname{} (full)         & \textbf{94.8} & \textbf{72.4} & \textbf{82.1} &  76 \\
\agentname{}--PromptGov     & 91.4 & 22.4 & 35.9 & 237 \\
\agentname{}--No-SM         & 51.7 & 16.3 & 24.8 & 184 \\
\agentname{}--NoPlanner     & 41.4 &  9.9 & 16.0 & 242 \\
\bottomrule
\end{tabular}
\end{threeparttable}
\end{table}

\begin{table}[!htbp]
\centering
\small
\caption{Paired analysis of clinician-rated alert actionability. The overall
comparison is the primary outcome; lead-time groups are exploratory.
Differences are \agentname{} minus observed-intervention scores. Confidence
intervals and one-sided $P$ values are stay-clustered; Holm $P$ covers the seven
exploratory comparisons. The positive-lead-time $\leq6$\,h subgroup combines the
two corresponding strata and excludes later alerts.}
\label{tab:app:hil_posthoc}
\resizebox{\textwidth}{!}{%
\begin{tabular}{lcrrlrr}
\toprule
Lead-time group & Pairs/stays & \agentname{} & Observed
  & Difference (95\% CI) & $P$ & Holm $P$ \\
\midrule
Overall & 46/36 & 3.70 & 3.49 & $0.20\;(-0.10,\,0.51)$ & 0.115 & --- \\
$0<\mathrm{lead}\leq6$\,h & 23/22 & 3.90 & 3.61 & $0.29\;(-0.05,\,0.60)$ & 0.064 & 0.447 \\
\midrule
Later alerts ($<0$\,h) & 2/2 & 3.50 & 4.17 & $-0.67\;(-2.33,\,1.00)$ & 0.750 & 1.000 \\
$0$--$3$\,h & 12/11 & 4.03 & 3.78 & $0.25\;(-0.12,\,0.59)$ & 0.156 & 0.938 \\
$>3$--$6$\,h & 11/11 & 3.76 & 3.42 & $0.33\;(-0.24,\,0.88)$ & 0.164 & 0.938 \\
$>6$--$9$\,h & 11/10 & 3.82 & 3.58 & $0.24\;(-0.53,\,0.87)$ & 0.296 & 1.000 \\
$>9$--$12$\,h & 5/5 & 3.53 & 3.47 & $0.07\;(-1.33,\,1.47)$ & 0.500 & 1.000 \\
$>12$\,h & 5/4 & 2.73 & 2.53 & $0.20\;(-0.11,\,0.75)$ & 0.375 & 1.000 \\
\bottomrule
\end{tabular}%
}
\end{table}

\begin{table}[!htbp]
\centering
\footnotesize
\caption{\textbf{Questions and extracted evidence by event anchor.} Offsets are relative
to treatment initiation for intervention-linked events and to onset for
deterioration-linked events. Evidence was extracted at $T_q$ over fixed look-back windows
of 6--72\,h; deterioration-linked packets also included the trigger value and time.
Counts are questions in the 1,020-question pool.}
\label{tab:app:qa_question_packs}
\begin{tabular}{>{\RaggedRight\arraybackslash}p{3.3cm}>{\RaggedRight\arraybackslash}p{5.3cm}>{\RaggedRight\arraybackslash}p{5.6cm}}
\toprule
Event anchor (questions) & Question types (offset) & Extracted evidence \\
\midrule
\multicolumn{3}{l}{\textit{Intervention-linked}} \\
Vasopressor initiation (139) & Next monitoring, escalation triggers ($-1$\,h); follow-up
  ($+3$\,h) & MAP and lactate trends; urine output; vasopressors or inotropes \\
Invasive mechanical ventilation (65) & Next monitoring, escalation triggers ($-1$\,h);
  follow-up ($+3$\,h) & SpO$_2$ and respiratory-rate trends; blood gases; oxygen or
  ventilatory support \\
CRRT initiation (21) & Missing information, next monitoring ($-1$\,h); follow-up ($+12$\,h)
  & Creatinine, potassium and pH trends; urine output; renal replacement therapy \\
Blood-product administration (78) & Evidence for and against ($-1$\,h); active support
  status ($+1$\,h); follow-up ($+12$\,h) & Haemoglobin and platelet trends; MAP;
  coagulation data; blood products \\
\midrule
\multicolumn{3}{l}{\textit{Deterioration-linked}} \\
MAP $<$65\,mmHg (293) & Alert explanation, escalation triggers ($+1$\,h); follow-up
  ($+3$\,h) & MAP trend; lactate; urine output; vasopressors or inotropes \\
SpO$_2$ $<$88\% (160) & Alert explanation, escalation triggers ($+1$\,h); follow-up
  ($+3$\,h) & SpO$_2$ trend; PaO$_2$; oxygen or ventilatory support \\
GCS decline $\geq$2 points (115) & Signal validity, evidence for and against ($+1$\,h);
  follow-up ($+6$\,h) & Current and baseline GCS; sedatives, analgesics and ventilation as
  possible confounders; respiratory and acid--base data \\
Haemoglobin drop $\geq$2\,g/dL (43) & Alert explanation, active support status ($+1$\,h);
  follow-up ($+12$\,h) & Current and prior haemoglobin; platelets; MAP; coagulation data;
  blood products \\
Platelet count $<100\times10^9$/L (42) & Severity and urgency, missing information
  ($+1$\,h); follow-up ($+24$\,h) & Platelet and INR trends; infection data; blood
  products \\
pH $<$7.25 (30) & Severity and urgency, missing information ($+1$\,h); follow-up ($+6$\,h)
  & pH trend and specimen type; lactate; acid--base and renal data \\
Lactate $\geq$4\,mmol/L (23) & Severity and urgency, evidence for and against ($+1$\,h);
  follow-up ($+6$\,h) & Lactate trend; MAP; infection and haemodynamic data \\
Potassium $>$6\,mmol/L (11) & Severity and urgency, next monitoring ($+1$\,h); follow-up
  ($+3$\,h) & Potassium and pH trends; renal and haemodynamic data; renal replacement
  therapy \\
\bottomrule
\end{tabular}
\end{table}

\begin{table}[!htbp]
\centering
\caption{Hard-failure categories and their deterministic score ceilings in the
automated clinical Q\&A rubric. When several categories are flagged, the final
score is capped at the minimum applicable ceiling.}
\label{tab:app:qa_hardfail}
\begin{tabular}{p{9cm}c}
\toprule
Hard-failure category & Score ceiling \\
\midrule
Future leakage                 & 1.0 \\
Hidden reference-event match claim & 1.0 \\
Unsafe recommendation          & 2.0 \\
Role-boundary violation        & 2.0 \\
Hallucinated fact              & 2.0 \\
Unsupported diagnosis          & 2.0 \\
Major omission                 & 3.0 \\
Inappropriate escalation       & 3.0 \\
Over-prediction                & 3.0 \\
\bottomrule
\end{tabular}
\end{table}

\begin{table}[!htbp]
\centering
\caption{Matched automated question-answering comparison with the Qwen baseline using jointly eligible scored questions from the 1,020-question pool (512 question pairs from 82 ICU stays). Differences are \agentname{} minus baseline on the 1--5 rubric scale. Pointwise 95\% confidence intervals use 20,000 ICU-stay bootstrap resamples. Paired label-swap tests are clustered by ICU stay and Holm-adjusted across the
15 dataset--comparator comparisons.}
\label{tab:app:qa_matched_qwen}
\small
\begin{tabular}{lrrrr}
\toprule
Dataset & Questions & Stays & Difference (95\% CI) & Adjusted $P$ \\
\midrule
MIMIC-IV & 184 & 38 & 0.784 (0.592--0.991) & $7.50\times10^{-6}$ \\
eICU-CRD & 199 & 24 & 0.303 (0.088--0.526) & 0.0350 \\
NWICU & 129 & 20 & 0.640 (0.488--0.821) & $2.10\times10^{-5}$ \\
\bottomrule
\end{tabular}
\end{table}

\begin{table}[!htbp]
\centering
\caption{Descriptive matched question-type scores for \agentname{} and the Qwen
baseline within the 1,020-question pool. Scores are on the 1--5 scale;
differences are \agentname{} minus baseline, with equal weight per answer.
Question counts sum to 512. The 12 signal-validity questions came from 12 stays
and were all anchored to GCS decline. No question-type-specific significance
tests were performed.}
\label{tab:app:qa_matched_types}
\small
\begin{tabular}{p{4.5cm}rrrr}
\toprule
Question type & Questions & \agentname{} & Qwen baseline & Difference \\
\midrule
Alert explanation & 114 & 4.432 & 3.488 & +0.944 \\
Signal validity & 12 & 4.546 & 3.067 & +1.479 \\
Severity and urgency & 31 & 4.645 & 3.963 & +0.682 \\
Evidence for and against & 24 & 4.806 & 3.871 & +0.935 \\
Missing information & 20 & 4.605 & 4.345 & +0.260 \\
Active support status & 18 & 4.350 & 3.528 & +0.822 \\
Next monitoring & 13 & 4.977 & 4.508 & +0.469 \\
Escalation triggers & 113 & 4.538 & 3.965 & +0.573 \\
Follow-up response & 167 & 4.286 & 4.123 & +0.163 \\
\bottomrule
\end{tabular}
\end{table}

\begin{table}[!htbp]
\centering
\caption{%
  EHR exploration tools. All tools are time-gated to the current
  simulation step $T$.}
\label{tab:app:toolset_retrieval}
\begin{tabular}{p{7.2cm}p{8.5cm}}
\toprule
Tool & Description \\
\midrule
\multicolumn{2}{l}{\textit{Vital signs and laboratory results}} \\[3pt]
\texttt{get\_vitals}             & Vital-sign records (HR, BP, MAP, SpO$_2$, RR, temperature) within a caller-specified time window \\
\texttt{get\_labs}               & Laboratory results within a time window, optionally filtered by analyte name \\
\texttt{query\_clinical\_data}   & General-purpose cross-category query interface for any structured clinical data type \\
\texttt{get\_vital\_trend}       & Time series and summary statistics for a single vital-sign parameter \\
\texttt{get\_lab\_trend}         & Time series and summary statistics for a single laboratory analyte \\
\midrule
\multicolumn{2}{l}{\textit{Medications, procedures, and fluid balance}} \\[3pt]
\texttt{get\_medication\_timeline}  & Medication timeline filtered by drug class or name; includes infusions spanning the step boundary \\
\texttt{get\_active\_medications}   & List of medications active at the current time step \\
\texttt{get\_procedure\_timeline}   & Procedure and treatment timeline; includes procedures spanning the step boundary \\
\texttt{get\_active\_procedures}    & List of procedures and treatments active at the current time step \\
\texttt{get\_io\_events}            & Intake/output event records, filterable by category (e.g.\ urine, crystalloid, blood products) \\
\texttt{get\_io\_balance}           & Cumulative fluid intake/output balance summary over a specified window \\
\midrule
\multicolumn{2}{l}{\textit{Clinical timeline and patient background}} \\[3pt]
\texttt{get\_icu\_timeline}         & Integrated ICU event timeline (vital alarms, nursing interventions, charted events) \\
\texttt{get\_transfer\_timeline}    & In-hospital transfer and care-unit transition records \\
\texttt{get\_transfers}             & Transfer record summary \\
\texttt{get\_patient\_summary}      & Baseline patient information: age, primary diagnosis, and reason for ICU admission \\
\texttt{get\_pre\_icu\_context}     & Clinical background from the 24 hours preceding ICU admission, including baseline creatinine and a partial SOFA estimate \\
\texttt{get\_prior\_alerts}         & Record of alerts previously emitted by the agent during the current evaluation run \\
\bottomrule
\end{tabular}
\end{table}

\begin{table}[!htbp]
\centering
\caption{Guideline-based calculator tools organised by physiological domain.}
\label{tab:app:toolset_calculators}
\begin{tabular}{p{7.2cm}p{8.5cm}}
\toprule
Tool & Computed quantity \\
\midrule
\multicolumn{2}{l}{\textit{Circulatory and perfusion}} \\[3pt]
\texttt{calculate\_map}                   & Mean arterial pressure: MAP $= (\mathrm{SBP} + 2 \times \mathrm{DBP}) / 3$; prioritises directly measured values \\
\texttt{calculate\_pulse\_pressure}       & Pulse pressure: $\mathrm{SBP} - \mathrm{DBP}$ \\
\texttt{calculate\_shock\_index}          & Shock index: $\mathrm{HR} / \mathrm{SBP}$ \\
\texttt{calculate\_vasopressor\_exposure} & Vasopressor exposure in three modes: contextual summary, SOFA-qualifying dose, and norepinephrine-equivalent dose \\
\midrule
\multicolumn{2}{l}{\textit{Respiratory}} \\[3pt]
\texttt{calculate\_pao2\_fio2}           & P/F ratio ($\mathrm{PaO}_2 / \mathrm{FiO}_2$); arterial blood-gas PaO$_2$ only \\
\texttt{calculate\_spo2\_fio2\_ratio}    & SpO$_2$/FiO$_2$ ratio, used when ABG is unavailable \\
\texttt{calculate\_rox\_index}           & ROX index $= (\mathrm{SpO}_2 / \mathrm{FiO}_2) / \mathrm{RR}$, for non-invasive ventilation assessment \\
\midrule
\multicolumn{2}{l}{\textit{Renal and fluid}} \\[3pt]
\texttt{calculate\_urine\_output\_rate}  & Urine output rate (mL\,kg$^{-1}$\,h$^{-1}$) over a caller-specified window \\
\texttt{calculate\_kdigo\_from\_raw}     & KDIGO AKI staging from raw serum creatinine and urine output \\
\midrule
\multicolumn{2}{l}{\textit{Neurological}} \\[3pt]
\texttt{calculate\_gcs\_delta}           & Change in GCS relative to a recent baseline, with sedation-state context \\
\midrule
\multicolumn{2}{l}{\textit{General laboratory}} \\[3pt]
\texttt{calculate\_lab\_delta}           & Absolute or directional change in any laboratory analyte over a caller-specified window \\
\midrule
\multicolumn{2}{l}{\textit{Composite organ-dysfunction scores}} \\[3pt]
\texttt{calculate\_sofa\_from\_raw}          & SOFA score from raw data; supports alignment to reference clock-hour bins \\
\texttt{calculate\_sofa\_delta\_from\_raw}   & SOFA score change over a window, with baseline-comparability filtering \\
\texttt{calculate\_sepsis3\_surrogate\_from\_raw} & Sepsis-3 surrogate criterion (suspected infection $+$ acute organ dysfunction) from raw data \\
\bottomrule
\end{tabular}
\end{table}

\begin{table}[!htbp]
\centering
\caption{Event-tracking tools.}
\label{tab:app:toolset_analytics}
\begin{tabular}{p{7.2cm}p{8.5cm}}
\toprule
Tool & Function \\
\midrule
\texttt{evaluate\_support\_episode}      & Determines whether a specified support treatment (e.g.\ vasopressor infusion, mechanical ventilation) is active at the current step \\
\texttt{evaluate\_time\_series\_condition} & Tests whether a variable persistently satisfies a caller-specified threshold, with configurable recovery-gap and consecutive-reading parameters \\
\texttt{evaluate\_delta\_condition}      & Tests whether the change in a variable over a specified window meets a caller-specified magnitude or direction criterion \\
\texttt{evaluate\_score\_condition}      & Tests whether a clinical score meets a caller-specified condition (value, direction, or crossing event) \\
\texttt{evaluate\_episode\_continuity}   & Classifies the status of an event across time steps as newly occurring, ongoing, or resolved \\
\bottomrule
\end{tabular}
\end{table}

\begin{table}[!htbp]
\centering
\caption{Record re-organization tools.}
\label{tab:app:toolset_context}
\begin{tabular}{p{7.2cm}p{8.5cm}}
\toprule
Tool & Covered dimensions \\
\midrule
\texttt{summarize\_recent\_icu\_course}          & Integrated narrative of the ICU course since admission \\
\texttt{get\_active\_support\_summary}           & Current respiratory, circulatory, and renal replacement therapy support status \\
\texttt{get\_circulatory\_context}               & Blood pressure, heart rate, lactate, and vasopressor exposure \\
\texttt{get\_respiratory\_context}               & SpO$_2$, FiO$_2$, ventilatory support mode, and arterial blood-gas values \\
\texttt{get\_renal\_fluid\_context}              & Serum creatinine, urine output, fluid balance, and KDIGO staging \\
\texttt{get\_neurologic\_sedation\_context}      & GCS, RASS, and sedoanalgesia medications \\
\texttt{get\_infection\_antimicrobial\_context}  & Culture results, antimicrobial regimen, and inflammatory markers \\
\texttt{get\_heme\_coag\_context}                & Haemoglobin, platelets, coagulation indices, and transfusion history \\
\texttt{get\_acid\_base\_electrolyte\_context}   & pH, HCO$_3^-$, pCO$_2$, potassium, and sodium \\
\texttt{get\_missingness\_report}                & Enumeration of absent measurements for a specified domain and time window \\
\texttt{get\_data\_quality\_safety\_context}     & Measurement-quality flags, sampling-sparsity warnings, and treatment--physiology consistency checks \\
\bottomrule
\end{tabular}
\end{table}

\begin{table}[!htbp]
\centering
\caption{Memory tools: shared agent state-table operations.}
\label{tab:app:toolset_state}
\begin{tabular}{p{7.2cm}p{8.5cm}}
\toprule
Tool & Function \\
\midrule
\texttt{define\_agent\_state\_table}  & Declares a typed, agent-scoped memo table with a caller-specified schema \\
\texttt{get\_agent\_state\_table}     & Retrieves all rows or a filtered subset of a declared state table \\
\texttt{upsert\_agent\_state\_row}    & Writes or updates a row in a state table \\
\texttt{append\_agent\_state\_note}   & Appends a free-text annotation to an existing row without overwriting it \\
\bottomrule
\end{tabular}
\end{table}

\begin{table}[!htbp]
\centering
\caption{%
  Memory tools: \agentname{}-specific read-only queries of CCTM. All writes go through the
  validated memory-writing operation $\Phi_w$ (Methods).}
\label{tab:app:toolset_memory}
\begin{tabular}{p{7.2cm}p{8.5cm}}
\toprule
Tool & Returned content \\
\midrule
\texttt{memory\_get\_l0}                  & Compact view of abnormal tracked-variable trajectories and their baseline references in the patient state ($\mathcal{M}_{\mathrm{state}}$) \\
\texttt{memory\_search\_l1\_cache}        & Search of timestamped patient-state and alert records (signal onset, recovery, alert emission, tracking change), filterable by event type and time window \\
\texttt{memory\_get\_permanent\_digests}  & Periodic summaries ($\mathcal{M}_{\mathrm{sum}}$), including per-category event counts and emitted alert types \\
\texttt{memory\_get\_prior\_alert\_episodes} & Recent emitted-alert records from the alert history ($\mathcal{M}_{\mathrm{alert}}$), filterable by event type and time window \\
\texttt{memory\_get\_alert\_episode\_state}  & Episode records in the alert history ($\mathcal{M}_{\mathrm{alert}}$) for a requested monitoring target, including emission times, counts and recorded signal values \\
\bottomrule
\end{tabular}
\end{table}

\begin{table}[!htbp]
\centering
\small
\setlength{\tabcolsep}{4pt}
\renewcommand{\arraystretch}{1.12}
\caption{Clinical guidelines included in the study's evidence catalogue. The four blood-product chapters of NICE NG24 are counted as one guideline.}
\label{tab:guideline_sources}
\begin{tabularx}{\linewidth}{@{}p{0.43\linewidth}cX@{}}
\toprule
Guideline & Year & Clinical scope \\
\midrule
Surviving Sepsis Campaign~\cite{prescott2026surviving} & 2026 &
Sepsis and septic shock.\\
SRLF--SFMU oxygen therapy~\cite{helms2024oxygen} & 2024 &
Acute hypoxaemic respiratory failure.\\
BTS/ICS ventilatory management~\cite{davidson2016bts} & 2016 &
Acute hypercapnic respiratory failure.\\
ERS/ATS noninvasive ventilation~\cite{rochwerg2017niv} & 2017 &
Noninvasive ventilation in acute respiratory failure.\\
KDIGO acute kidney injury~\cite{kdigo2012aki} & 2012 &
AKI assessment and renal replacement therapy.\\
SRLF--GFRUP renal replacement therapy~\cite{jourdain2025rrt} & 2025 &
Renal replacement therapy in intensive care.\\
Multidisciplinary ICU RRT~\cite{meerschdini2026rrt} & 2026 &
RRT indications and treatment modalities.\\
UK Kidney Association hyperkalaemia~\cite{ukka2023hyperkalaemia} & 2023 &
Acute hyperkalaemia in adults.\\
AABB red-cell transfusion~\cite{carson2023red} & 2023 &
Red-cell transfusion across clinical populations.\\
AABB red-cell transfusion in acute MI~\cite{pagano2025ami} & 2025 &
Transfusion in acute myocardial infarction.\\
AABB/ICTMG platelet transfusion~\cite{metcalf2025platelet} & 2025 &
Platelet transfusion across clinical and procedural contexts.\\
NICE blood transfusion (NG24)~\cite{nice2015blood} & 2015 &
Red cells, platelets, plasma and cryoprecipitate.\\
\bottomrule
\end{tabularx}
\end{table}

\FloatBarrier
\clearpage


\section{Cohort and reference-event construction}
\label{app:cohort_details}

\paragraph{Within-stratum sampling and cohort enrichment.}
To increase evaluable events within a fixed replay budget without selecting stays with extreme event counts (up to 15--24), sampling targeted a moderate number of events per stay.
Within each length-of-stay stratum, stays with at least one reference event in the replay window were ranked by $|n-6|$, where $n$ is the number of precomputed reference events in the stay, with fixed-seed random tie-breaking.
Among the 66,694 eligible stays (Fig.~\ref{fig:data_form}a), six is the 75th percentile in MIMIC-IV (median 4, IQR 2--6) and lies between the medians for stays with an intervention (7 in MIMIC-IV and 4 in eICU-CRD).
As a result, test stays had a median of 5.0 scored evaluation episodes (IQR 3.8--6.0).
Selection used only stay metadata and event counts, and both splits were frozen before any model evaluation, without manual screening or replacement.
Because of this enrichment, precision and alert counts support comparisons between methods, not estimates of positive predictive value in routine care; the machine-learning training cohort was not enriched.

\section{Supplementary evaluation details}
\label{app:evaluation_details}

\subsection{Automated Evaluation of Alerts against Reference Events}
\label{app:monitoring_evaluation}

\subsubsection{Baseline settings}
\label{app:baseline_settings}

\begin{itemize}
\item \textbf{Guideline-informed rules.} The rule set comprised 40 branches from
nine guidelines and was fixed before evaluation on the test cohort (Supplementary
Table~\ref{tab:app:guideline_rules}). Rules were evaluated against
visible vital signs from the preceding 4\,h and laboratory results from the preceding
24\,h. Conditions without a structured mapping were treated as unknown and could
not alone satisfy a branch. Seven branches (V02, I03, I08, I09, I11,
B01 and B04) generated all 1,285 alerts.

\item \textbf{Machine-learning models.} Alerts required the corresponding support to be
known to be inactive, and no cooldown was applied. Confidence intervals exclude
uncertainty from model training and operating-point selection.

\item \textbf{Threshold rules.} The suppression
interval was selected on the validation cohort from candidates of 2, 4, 6 and 12\,h by the
mean F1 of the multivariable (MIMIC-IV) and single-variable (pooled) deterioration tasks,
which was 10.5\%, 18.2\%, 23.4\% and 34.3\%, respectively.
\end{itemize}

\subsection{Manual Assessment of Alert Timing Appropriateness}
\label{app:clinician_timing}

\paragraph{Review procedure.}
The three reviewers had advised on the design of the assessment platform using
examples outside the test set, and all received a written standard operating
procedure before assessment (Supplementary Fig.~\ref{fig:app:timing_interface}).
The 46 pairs came from 36 ICU stays. Besides the anchors given in Methods, a
score of 2 denoted weak support and 4 well-supported timing. The overall
judgement was made in a side-by-side comparison view after both times had been
scored. Current-admission diagnoses were shown without time restriction at the
reviewers' request, and a guideline quick reference for the target intervention
was available throughout. Optional entries (best time, acceptable window,
confidence and rationale) were not analysed. Sources were revealed after all
reviews were completed.

\paragraph{Statistical analysis.}
For the primary outcome (Methods), we averaged within-reviewer, within-pair
score differences (\agentname{} minus observed intervention), applying one sign
jointly to all differences from the same ICU stay. We enumerated all sign
assignments for comparisons involving up to 20 stays and otherwise used 1,000,000
random assignments, with Monte Carlo $P = (b+1)/(1{,}000{,}000+1)$, where $b$ is
the number of assignments at least as extreme as the observed statistic. The
alternative was one-sided (positive mean difference). Holm correction covered
seven exploratory comparisons (positive lead times up to 6\,h and the six
lead-time strata in Fig.~\ref{fig:clinician_timing}e); the primary overall
comparison was outside this family. Pointwise percentile confidence intervals used 20,000
stay-cluster bootstrap resamples. Results are shown in Supplementary
Table~\ref{tab:app:hil_posthoc}.

\paragraph{Physiological context of alerts.}
\label{app:alert_temporal_validity}
For each of 34 prespecified signals, the value at each intervention
alert was compared with values at one to three time points from other stays at
which the same target had been assessed without an alert, matched on database,
target, support status and elapsed ICU time (within 12\,h) but not on signal
values. Estimation required at least ten matched sets from five stays.

\subsection{Evaluation of Clinical Question Answering}
\label{app:qa_protocol}

\paragraph{Question construction and delivery.}
\label{app:qa_questions}

Each event anchor received a fixed pack of up to three question types at the offsets
listed in Supplementary Table~\ref{tab:app:qa_question_packs}; questions falling outside
the replay window were excluded. Anchors also included the Glasgow Coma Scale decline
subgroup events, which were excluded from the 442 primary reference events (Methods).
Template wording was varied using LLM-generated paraphrases, frozen before evaluation
and assigned by a fixed hash so that all systems received identical questions. Hidden
anchor identifiers and future intervention times were not supplied to answering systems.

\paragraph{Reference packets and scoring.}
\label{app:qa_scoring}

Reference packets were built by deterministic rules, without a language model or access
to candidate answers, and were shown only to the judge. Event-specific evidence was
retrieved at $T_q$ through the same time-gated interface used by the answering systems
(Supplementary Table~\ref{tab:app:qa_question_packs}), with unavailable values recorded as
missing rather than normal. Answer keys listed points to cover, useful context, common
pitfalls and prohibited claims, such as stating that a future intervention had started;
intervention-linked keys combined those of the associated triggers (for example, the MAP
and lactate keys for vasopressor initiation). For follow-up questions, the packet also
contained a label assigned by deterministic recovery rules from reassessments between the
event and $T_q$: recovery without additional support, improvement with additional support,
worsening, persistent abnormality or insufficient reassessment data. The judge's own
aggregate score was discarded; hard-failure ceilings are listed in Supplementary
Table~\ref{tab:app:qa_hardfail}.

\subsection{Failure-Mode Analysis and Case Studies}
\label{app:failure_modes}
\label{app:failure_decomposition}
\label{app:case_reconstruction}

Each emitted raw alert was classified as a hit, a duplicate within an active
episode, a too-early or too-late alert, or an alert without an eligible same-target
reference event (Fig.~\ref{fig:failure_modes}a,d). Classes were mutually exclusive
and exhaustive, and class counts were checked against the alert totals for every
run. Deterioration alerts were classified relative to their trigger windows and full
reference-episode boundaries, and intervention alerts relative to the adjudicated
clinical-onset interval. Alerts without an observed same-target intervention were
not labelled clinical errors without expert review. Tool-call transitions
(Fig.~\ref{fig:e6_temporal_validity}b) were analysed only at intervention-candidate
times with complete traces. Case reconstructions (Supplementary
Fig.~\ref{fig:case_analysis}) show the information available at each decision and
do not establish why the agent decided as it did.

\section{Single-agent baseline implementation}
\label{app:single_agent_implementation}

Each single-agent baseline, $\mathcal{A}_{\mathrm{single}}$, assessed intervention
needs and clinical deterioration through iterative function calling with one
language model.

\paragraph{Prompt.}
The agent's instructions concatenated five components:
\begin{equation}
P_{\mathrm{single}}
=
\operatorname{concat}\!\left(
P_{\mathrm{base}}, P_{\mathrm{A}}, P_{\mathrm{B}},
P_{\mathcal{T}}, P_{\mathrm{output}}
\right).
\end{equation}
$P_{\mathrm{base}}$ defined the agent's role, monitoring tasks and
information-access constraints. $P_{\mathrm{A}}$ and $P_{\mathrm{B}}$ provided
guidance for intervention anticipation and deterioration detection, termed
Reference A and Reference B in the prompts: $P_{\mathrm{A}}$ specified
intervention targets, relevant evidence, actionability conditions and
suppression of warnings when the target intervention was already active, and
$P_{\mathrm{B}}$ specified physiological thresholds, score- and stage-based
criteria, evidence requirements and conditions for repeat alerts.
$P_{\mathcal{T}}$ guided tool selection, interpretation of tool results and
when to stop calling tools, and $P_{\mathrm{output}}$ specified the output
schema. For clinical question answering, the question $\mathcal{Q}_t$ was
supplied together with additional instructions, $P_{\mathrm{qa}}$, specifying
the answer requirements; answers were produced by the same agent. Full texts are
given in Supplementary Note~\ref{app:baseline_prompt_text}.

\paragraph{Inputs and tools.}
Besides $\mathcal{I}$ and $\mathcal{C}_{1:t}$, the agent received the current
simulation time at each hourly assessment $t$. Its tool set
(Supplementary Tables~\ref{tab:app:toolset_retrieval}--\ref{tab:app:toolset_state}) was
\begin{equation}
\mathcal{T}_{\mathrm{single}}
=
\mathcal{T}_{\mathrm{EHR}} \cup \mathcal{T}_{\mathrm{org}} \cup
\mathcal{T}_{\mathrm{calc}} \cup \mathcal{T}_{\mathrm{evt}} \cup
\mathcal{T}_{\mathrm{mem}}^{\mathrm{shared}},
\end{equation}
where $\mathcal{T}_{\mathrm{mem}}^{\mathrm{shared}}$ denotes the four shared
memory tools. The \agentname{}-specific memory tools and $\mathcal{T}_{\mathrm{guid}}$
were not available; guideline content was provided through $P_{\mathrm{A}}$ and
$P_{\mathrm{B}}$. The same model first selected relevant tools from a catalogue
of tool names and short descriptions; the argument schemas of the selected
tools, together with the prior-alert retrieval tool, were then provided for the
assessment. If this selection failed, all tools were provided.

\paragraph{Memory.}
The model decided what to store in its memory, $\mathcal{M}_{\mathrm{single}}$,
how to organize it and when to retrieve it. Each assessment started a new conversation, so previous dialogue
and tool results were not carried forward; stored memory and previously emitted
alerts entered an assessment only when the agent retrieved them.

\paragraph{Assessment loop.}
Using the notation of \agentname{}, each target-specific assessment can be
written as
\begin{equation}
(y_t, r_t, g_t)
=
\mathcal{A}_{\mathrm{single}}\!\left(
\mathcal{C}_{1:t}
\mid
\mathcal{I}, P_{\mathrm{single}}, \mathcal{T}_{\mathrm{single}}
\right),
\end{equation}
where $y_t$ indicates whether an alert is issued, $r_t$ is the supporting
rationale and $g_t \in \mathcal{G}$ is the monitoring target; one assessment
may involve several model and tool calls. Tool results were returned to the
same conversation, and the agent used the accumulated evidence to request
further calls, including memory reads and writes, or to give its final
assessment. The configured budget was eight tool calls per assessment. The
budget was checked after each batch of calls returned by the model, so the last
batch could exceed it; the model was then asked for a final assessment without
tools. Tool errors were returned to the model as structured messages.

\paragraph{Output.}
Final outputs followed the shared JSON alert schema (task, target event, alert
class, priority and rationale), with an empty array indicating no alert; when
both tasks warranted alerts, separate alert objects were returned. Outputs were
normalized and, if the JSON was invalid, repaired by an additional model call.
Accepted alerts were time-stamped at the current simulation time. Repeat
alerting was governed solely by the prompt instructions, as in the PromptGov
variant of the component ablation (Supplementary Table~\ref{tab:e4_component_ablation_qwen}). All
tool calls, model outputs and errors were logged.


\section{Baseline system prompt and clinical QA instructions}
\label{app:baseline_prompt_text}

\subsection{Baseline system prompt ($P_{base}$)}
\label{app:baseline_system_base}

\begin{tcolorbox}[
    breakable,
    title={System prompt: baseline role and monitoring task ($P_{base}$)},
    title after break={System prompt: baseline role and monitoring task ($P_{base}$)\ (continued)},
    colback=white, colframe=black!65,
    colbacktitle=black!70, coltitle=white,
    fonttitle=\normalsize\bfseries,
    fontupper=\small\ttfamily,
    boxrule=0.5pt, arc=1.5mm,
    left=3mm, right=3mm, top=3mm, bottom=3mm,
    toptitle=1.5mm, bottomtitle=1.5mm,
    before skip=10pt, after skip=12pt
]
\raggedright
\setlength{\parindent}{0pt}
\setlength{\parskip}{0pt}
\setlength{\emergencystretch}{2em}
\#\# Role and Task\par
\smallskip
You are an ICU clinical deterioration detection agent. At each simulation\par
time point, inspect only time-gated patient data visible at or before the\par
current simulation time T, then output alerts for the active task bundle.\par
\smallskip
This multi-task run covers two scoring families in parallel:\par
- Reference A: intervention warnings \textemdash{} current evidence suggesting likely\par
\hspace*{1.0em}near-term need for a clinical intervention before it is visible.\par
- Reference B: objective organ-system deterioration \textemdash{} B\_core score/criteria\par
\hspace*{1.0em}events and B\_expanded severe single-domain threshold crossings.\par
\smallskip
Do not claim future values, hidden labels, GT events, evaluator matches,\par
true positives, false positives, patient identifiers, or dataset identifiers\par
as observed facts.\par
\smallskip
Tools provide time-gated evidence, calculations, episode status, and\par
episode-continuity context. Tool statuses are evidence only, not alert\par
recommendations.\par
\smallskip
\#\# Shared Alert Threshold\par
\smallskip
Emit alerts only for directly supported severe deterioration, likely\par
near-term intervention need, calculated organ-failure score events, or\par
severe single-domain objective threshold crossings. Do not alert for\par
mild/moderate abnormalities, stable treated conditions, broad watchlist\par
concerns, or speculative future risk without current evidence.\par
\smallskip
Keep Reference A and Reference B as separate scoring tasks. When both are\par
supported at the same time step, emit separate alert objects with different\par
"task" values rather than merging them into one alert.\par
\smallskip
\#\# Primary Event Ontology\par
\smallskip
Reference A primary\_event values (intervention warnings only):\par
- intubation\_start\par
- vasopressor\_start\par
- crrt\_start\par
- blood\_product\par
\smallskip
Reference B primary\_event values:\par
- spo2\_lt\_88\par
- map\_lt\_65\par
- lactate\_ge\_4\par
- gcs\_drop\_ge\_2\par
- aki\_kdigo\_stage\_ge\_2\par
- aki\_kdigo\_worsening\par
- hemoglobin\_drop\_ge\_2\par
- platelet\_lt\_100\par
- ph\_lt\_725\par
- potassium\_gt\_6\par
- sepsis3\_onset\par
- sofa\_ge\_6\par
- sofa\_ge\_8\par
- sofa\_delta\_ge\_2\par
\smallskip
\#\# Alert Object Schema\par
\smallskip
Each alert object must contain:\par
- "task": "reference\_a" or "reference\_b".\par
- "primary\_event": one valid event from that task's ontology above.\par
- "event\_class": one valid class for that task (see task-specific rules).\par
- "priority": "RED" for immediate threat or "AMBER" for severe but not\par
\hspace*{1.0em}immediately life-threatening.\par
- "reason": 1-2 sentences citing objective time-gated evidence.\par
\smallskip
Prefer 0-4 alerts total per assessment. If nothing meets the threshold\par
for either task, output [].\par
\end{tcolorbox}

\subsection{Reference A: intervention-warning rules ($P_{A}$)}
\label{app:baseline_reference_a}

\begin{tcolorbox}[
    breakable,
    title={System prompt: Reference A rules ($P_{A}$)},
    title after break={System prompt: Reference A rules ($P_{A}$)\ (continued)},
    colback=white, colframe=black!65,
    colbacktitle=black!70, coltitle=white,
    fonttitle=\normalsize\bfseries,
    fontupper=\small\ttfamily,
    boxrule=0.5pt, arc=1.5mm,
    left=3mm, right=3mm, top=3mm, bottom=3mm,
    toptitle=1.5mm, bottomtitle=1.5mm,
    before skip=10pt, after skip=12pt
]
\raggedright
\setlength{\parindent}{0pt}
\setlength{\parskip}{0pt}
\setlength{\emergencystretch}{2em}
\#\# Reference A Rules\par
\smallskip
Reference A evaluates likely near-term intervention needs (warnings only).\par
Do not emit intervention\_detection alerts \textemdash{} visible interventions suppress the warning.\par
\smallskip
Use event\_class "intervention\_warning" when current evidence suggests likely\par
near-term need for a listed intervention and that intervention is not yet\par
visible. A visible active intervention means the warning window has passed;\par
suppress or omit the warning.\par
\smallskip
Objective deterioration evidence may also qualify a Reference A warning when\par
it indicates a near-term support need and the target support is not currently\par
active. Do not treat MAP \textless{}65, severe thrombocytopenia/coagulopathy, KDIGO\par
stage/oliguria, or severe hypoxemia as "Reference B only" when the same facts\par
satisfy the intervention-specific actionability threshold below.\par
\smallskip
Choose the expected intervention as primary\_event and compile the matching\par
evidence\_to\_track and actionability\_threshold into the task\_plan row:\par
\smallskip
- vasopressor\_start: shock physiology or hypoperfusion. Track MAP, SBP, HR,\par
\hspace*{1.0em}lactate, HCO3, urine output, and perfusion markers. Actionable when\par
\hspace*{1.0em}persistent MAP \textless{}65 or SBP \textless{}90 despite adequate fluids, lactate \ensuremath{\geq}4 with\par
\hspace*{1.0em}hemodynamic instability, or poor perfusion \textemdash{} before vasopressor is visible.\par
\hspace*{1.0em}Suppress when vasopressor infusion is active, or when single transient\par
\hspace*{1.0em}hypotension without perfusion context.\par
\smallskip
- intubation\_start: severe or worsening respiratory failure. Track SpO2, RR,\par
\hspace*{1.0em}FiO2, PaO2, PaO2/FiO2 ratio, pH, PaCO2, HCO3, GCS, and current oxygen/\par
\hspace*{1.0em}NIV/HFNC support. Actionable when SpO2 \textless{}90\% despite O2, PaO2/FiO2 \textless{}200,\par
\hspace*{1.0em}FiO2 \textgreater{}0.6 and escalating, rising PaCO2 or falling pH indicating ventilatory\par
\hspace*{1.0em}failure, or GCS \ensuremath{\leq}8 with airway compromise \textemdash{} before intubation is visible.\par
\hspace*{1.0em}Suppress when invasive mechanical ventilation is active, or DNI/comfort goals.\par
\smallskip
- crrt\_start: severe renal, metabolic, or fluid deterioration. Track\par
\hspace*{1.0em}creatinine, BUN, urine output, potassium, pH, HCO3, lactate, and fluid\par
\hspace*{1.0em}balance. Actionable when KDIGO AKI \ensuremath{\geq}2, oliguria/anuria despite resuscitation,\par
\hspace*{1.0em}refractory acidosis (pH \textless{}7.2) or hyperkalemia (K+ \textgreater{}6.5) not corrected\par
\hspace*{1.0em}medically, or severe fluid overload \textemdash{} before CRRT is visible. Treat dialysis\par
\hspace*{1.0em}access and any preparation phase derived from that access as one supporting\par
\hspace*{1.0em}fact, not independent indications; access plus AKI alone does not overrule\par
\hspace*{1.0em}clearly reassuring current urine output, electrolytes, acid-base state, and\par
\hspace*{1.0em}fluid/uremic context.\par
\hspace*{1.0em}Suppress when active RRT is running.\par
\smallskip
- blood\_product: severe anemia, thrombocytopenia, coagulopathy, or bleeding.\par
\hspace*{1.0em}Track hemoglobin, hematocrit, platelet count, INR, PTT, fibrinogen, HR, SBP,\par
\hspace*{1.0em}drain output, and transfusion administration records. Actionable when Hgb\par
\hspace*{1.0em}drops \ensuremath{\geq}2 g/dL or Hgb \ensuremath{\leq}7 g/dL, platelet \textless{}50 K/\ensuremath{\mu}L with active bleeding or\par
\hspace*{1.0em}high-risk procedure, INR \textgreater{}2 or fibrinogen \textless{}150 with bleeding concern, or\par
\hspace*{1.0em}active hemorrhage with hemodynamic impact \textemdash{} before blood product is visible.\par
\hspace*{1.0em}Suppress when blood product (RBC/platelet/plasma/cryo) is currently active\par
\hspace*{1.0em}for the active concern.\par
\smallskip
Do not use Reference B primary\_event values as Reference A alert targets.\par
Non-intervention findings may appear in reason text but must not be primary\_event.\par
\end{tcolorbox}

\subsection{Reference B: objective-deterioration rules ($P_{B}$)}
\label{app:baseline_reference_b}

\begin{tcolorbox}[
    breakable,
    title={System prompt: Reference B rules ($P_{B}$)},
    title after break={System prompt: Reference B rules ($P_{B}$)\ (continued)},
    colback=white, colframe=black!65,
    colbacktitle=black!70, coltitle=white,
    fonttitle=\normalsize\bfseries,
    fontupper=\small\ttfamily,
    boxrule=0.5pt, arc=1.5mm,
    left=3mm, right=3mm, top=3mm, bottom=3mm,
    toptitle=1.5mm, bottomtitle=1.5mm,
    before skip=10pt, after skip=12pt
]
\raggedright
\setlength{\parindent}{0pt}
\setlength{\parskip}{0pt}
\setlength{\emergencystretch}{2em}
\#\# Reference B Rules\par
\smallskip
Reference B evaluates objective organ-system deterioration episodes.\par
\smallskip
Use event\_class "indicator\_deterioration" for severe single-domain\par
objective triggers:\par
- spo2\_lt\_88: SpO2 \textless{}88\%, especially persistent, recurrent, severe, or\par
\hspace*{1.0em}despite oxygen/ventilatory support.\par
- map\_lt\_65: MAP \textless{}65 mmHg. Prefer directly visible MAP. If only SBP and\par
\hspace*{1.0em}DBP are visible, calculate MAP = (SBP + 2*DBP) / 3.\par
- lactate\_ge\_4: lactate \ensuremath{\geq}4 mmol/L.\par
- gcs\_drop\_ge\_2: GCS drops by \ensuremath{\geq}2 from a recent baseline; consider\par
\hspace*{1.0em}sedation and airway context.\par
- hemoglobin\_drop\_ge\_2: hemoglobin drops by \ensuremath{\geq}2 g/dL from a recent\par
\hspace*{1.0em}baseline.\par
- platelet\_lt\_100: platelet count \textless{}100 K/uL.\par
- ph\_lt\_725: pH \textless{}7.25.\par
- potassium\_gt\_6: potassium \textgreater{}6 mmol/L.\par
\smallskip
Use event\_class "organ\_failure\_score" for score, stage, or criteria events:\par
- sofa\_ge\_6: calculation-backed recent-window SOFA \ensuremath{\geq}6.\par
- sofa\_ge\_8: calculation-backed recent-window SOFA \ensuremath{\geq}8.\par
- sofa\_delta\_ge\_2: SOFA rises by \ensuremath{\geq}2 from a recent baseline/low point.\par
\hspace*{1.0em}At early ICU time (T \textless{} 2h) when insufficient ICU data exists for a\par
\hspace*{1.0em}lookback baseline, use pre\_icu\_context.sofa\_partial\_estimate as the\par
\hspace*{1.0em}baseline lower bound. If current SOFA \ensuremath{\geq} pre-ICU lower bound + 2,\par
\hspace*{1.0em}a sofa\_delta\_ge\_2 event may be supported.\par
- sepsis3\_onset: current or newly visible infection suspicion plus acute,\par
\hspace*{1.0em}new, or worsening organ dysfunction.\par
- aki\_kdigo\_stage\_ge\_2: KDIGO AKI stage calculated \ensuremath{\geq}2.\par
- aki\_kdigo\_worsening: KDIGO stage objectively worsens from recent prior\par
\hspace*{1.0em}state. When pre\_icu\_context.baseline\_creatinine\_mg\_dl is available, use\par
\hspace*{1.0em}it as the pre-ICU creatinine baseline for KDIGO staging.\par
\smallskip
Score-, stage-, delta-, and criteria-based events must be\par
calculation-backed. Do not fabricate missing values. If inputs are\par
insufficient, choose a better-supported single-domain event or omit.\par
\smallskip
Stable high scores, chronic organ dysfunction, or ongoing treatment do not\par
justify repeat alerts. Re-alert only for first crossing, clear worsening,\par
newly met criteria, recurrence after recovery, or a distinct new episode\par
not covered by a prior alert at this step.\par
\end{tcolorbox}

\subsection{Clinical QA instructions ($P_{qa}$)}
\label{app:baseline_qa_instructions}

\begin{tcolorbox}[
    breakable,
    title={Clinical QA instructions ($P_{qa}$)},
    title after break={Clinical QA instructions ($P_{qa}$)\ (continued)},
    colback=white, colframe=black!65,
    colbacktitle=black!70, coltitle=white,
    fonttitle=\normalsize\bfseries,
    fontupper=\small\ttfamily,
    boxrule=0.5pt, arc=1.5mm,
    left=3mm, right=3mm, top=3mm, bottom=3mm,
    toptitle=1.5mm, bottomtitle=1.5mm,
    before skip=10pt, after skip=12pt
]
\raggedright
\setlength{\parindent}{0pt}
\setlength{\parskip}{0pt}
\setlength{\emergencystretch}{2em}
\#\# Clinical QA Instructions\par
\smallskip
In addition to monitoring for deterioration, answer clinical questions\par
included in the Active QA Task Bundle.\par
\smallskip
For each question:\par
1. Identify its clinical focus, intended clinician role, and time frame.\par
\hspace*{1.5em}Use only patient evidence available at or before question\_time\_minutes.\par
\hspace*{1.5em}Do not use later observations, even if they appear in the current snapshot.\par
\smallskip
2. Review the relevant clinical snapshot and available tool results.\par
\hspace*{1.5em}Retrieve additional time-gated records when needed to answer the question.\par
\hspace*{1.5em}Distinguish current findings from historical events and prior agent alerts.\par
\smallskip
3. Give a direct answer supported by relevant patient-specific evidence.\par
\hspace*{1.5em}For trend questions, compare observations in chronological order.\par
\hspace*{1.5em}Do not treat missing data as normal findings, medication orders as confirmed\par
\hspace*{1.5em}administration, or temporal associations as proof of causation.\par
\smallskip
4. State missing, conflicting, or outdated information when it materially\par
\hspace*{1.5em}limits the answer. Do not invent measurements, diagnoses, or treatment\par
\hspace*{1.5em}responses.\par
\smallskip
5. Answer in the question's language and tailor the response to the clinician's\par
\hspace*{1.5em}role. For nurses, emphasize monitoring, verification, and escalation.\par
\hspace*{1.5em}For physicians, include diagnostic and management considerations when\par
\hspace*{1.5em}supported by the available evidence.\par
\smallskip
6. Answer every supplied question using its exact task\_id. Keep the answer\par
\hspace*{1.5em}concise and focused. A QA task does not itself require an alert; assess\par
\hspace*{1.5em}alerts separately using the monitoring criteria.\par
\smallskip
When QA tasks are present, return one JSON object:\par
\{\par
\hspace*{1.0em}"alerts": [],\par
\hspace*{1.0em}"qa\_answers": [\par
\hspace*{2.0em}\{\par
\hspace*{3.0em}"task\_id": "\textless{}task\_id from the Active QA Task Bundle\textgreater{}",\par
\hspace*{3.0em}"answer": "\textless{}direct clinician-facing answer\textgreater{}",\par
\hspace*{3.0em}"reasoning": "\textless{}1\textendash{}2 sentences citing relevant time-gated evidence\textgreater{}"\par
\hspace*{2.0em}\}\par
\hspace*{1.0em}]\par
\}\par
\smallskip
Populate alerts only when the monitoring criteria are met.\par
When no QA task is present, follow the standard alert-only output format.\par
\end{tcolorbox}

\subsection{Function-calling tool guidance ($P_{\mathcal{T}}$)}
\label{app:baseline_tool_guidance}

\begin{tcolorbox}[
    breakable,
    title={System prompt: function-calling tool agent ($P_{\mathcal{T}}$)},
    title after break={System prompt: function-calling tool agent ($P_{\mathcal{T}}$)\ (continued)},
    colback=white, colframe=black!65,
    colbacktitle=black!70, coltitle=white,
    fonttitle=\normalsize\bfseries,
    fontupper=\small\ttfamily,
    boxrule=0.5pt, arc=1.5mm,
    left=3mm, right=3mm, top=3mm, bottom=3mm,
    toptitle=1.5mm, bottomtitle=1.5mm,
    before skip=10pt, after skip=12pt
]
\raggedright
\setlength{\parindent}{0pt}
\setlength{\parskip}{0pt}
\setlength{\emergencystretch}{2em}
\#\# Function-Calling Tool Agent\par
\smallskip
You may call only the provided function tools. You cannot execute code,\par
run raw SQL, access files, query outside the current patient/stay, or\par
bypass the current time gate.\par
\smallskip
Read the current snapshot first. Use tools only when they clarify event\par
severity, timing, calculation inputs, active support status, or episode\par
continuity for Reference A or Reference B. Stop calling tools once the\par
evidence is sufficient to answer both tasks.\par
\smallskip
Tool budget: prefer 3-6 targeted calls per step. Do not run broad\par
screening panels when the snapshot already shows no candidate events.\par
\smallskip
Task strategy:\par
1. Scan the snapshot for Reference B threshold crossings first \textemdash{} these\par
\hspace*{1.5em}are most objective and determine whether score-based tools are needed.\par
2. Use the same tool facts to evaluate Reference A intervention warnings.\par
3. Suppress Ref A warnings for interventions already visible as active.\par
\smallskip
\#\# Output Faithfulness\par
\smallskip
Your final JSON array MUST be consistent with your reasoning conclusions.\par
If your reasoning concludes a finding "meets criteria" or "meets the\par
threshold", you MUST include that alert. If your reasoning concludes "does\par
not meet threshold", output [] for that task. Never contradict your own\par
analysis \textemdash{} resolve uncertainty in your reasoning before generating output.\par
\end{tcolorbox}

\subsection{Output format ($P_{output}$)}
\label{app:baseline_output_format}

\begin{tcolorbox}[
    breakable,
    title={System prompt: output format $P_{output}$ },
    title after break={System prompt: output format $P_{output}$\ (continued)},
    colback=white, colframe=black!65,
    colbacktitle=black!70, coltitle=white,
    fonttitle=\normalsize\bfseries,
    fontupper=\small\ttfamily,
    boxrule=0.5pt, arc=1.5mm,
    left=3mm, right=3mm, top=3mm, bottom=3mm,
    toptitle=1.5mm, bottomtitle=1.5mm,
    before skip=10pt, after skip=12pt
]
\raggedright
\setlength{\parindent}{0pt}
\setlength{\parskip}{0pt}
\setlength{\emergencystretch}{2em}
\#\# Standard output (no QA tasks)\par
Output ONLY a JSON array of alert objects. No markdown, no commentary.\par
Your entire response must start with `[` and end with `]`.\par
Return raw JSON only. Do NOT wrap the JSON in markdown fences such as\par
```json or ```.\par
\smallskip
Each alert must include: task, primary\_event, event\_class, priority, reason.\par
\smallskip
Example:\par
[\par
\hspace*{1.0em}\{\par
\hspace*{2.0em}"task": "reference\_a",\par
\hspace*{2.0em}"primary\_event": "vasopressor\_start",\par
\hspace*{2.0em}"event\_class": "intervention\_warning",\par
\hspace*{2.0em}"priority": "RED",\par
\hspace*{2.0em}"reason": "Persistent MAP 58 mmHg with lactate 4.2 mmol/L and tachycardia indicates shock physiology and likely near-term vasopressor need."\par
\hspace*{1.0em}\},\par
\hspace*{1.0em}\{\par
\hspace*{2.0em}"task": "reference\_b",\par
\hspace*{2.0em}"primary\_event": "map\_lt\_65",\par
\hspace*{2.0em}"event\_class": "indicator\_deterioration",\par
\hspace*{2.0em}"priority": "RED",\par
\hspace*{2.0em}"reason": "MAP 58 mmHg at current time step, below 65 threshold, persisting despite 2L fluid resuscitation."\par
\hspace*{1.0em}\}\par
]\par
\smallskip
If no alert meets the threshold for either task, output: []\par
\smallskip
\#\# When QA tasks are present\par
Output a JSON object with two keys: "alerts" and "qa\_answers".\par
\{\par
\hspace*{1.0em}"alerts": [ ... ],\par
\hspace*{1.0em}"qa\_answers": [\par
\hspace*{2.0em}\{\par
\hspace*{3.0em}"task\_id": "\textless{}task\_id from the QA bundle\textgreater{}",\par
\hspace*{3.0em}"answer": "\textless{}concise answer in the role's language\textgreater{}",\par
\hspace*{3.0em}"reasoning": "\textless{}1-2 sentences citing time-gated evidence\textgreater{}"\par
\hspace*{2.0em}\}\par
\hspace*{1.0em}]\par
\}\par
Answer every task in the Active QA Task Bundle. If no alerts, use "alerts": [].\par
\end{tcolorbox}

\section{\agentname prompts}
\label{app:icuagent_prompts}

\subsection{Planner prompt ($P_1$)}
\label{app:icuagent_planner_prompt}

\begin{tcolorbox}[
    breakable,
    title={Planner prompt (P1): scope and planning instructions},
    title after break={Planner prompt (P1): scope and planning instructions\ (continued)},
    colback=white, colframe=black!65,
    colbacktitle=black!70, coltitle=white,
    fonttitle=\normalsize\bfseries,
    fontupper=\small\ttfamily,
    boxrule=0.5pt, arc=1.5mm,
    left=3mm, right=3mm, top=3mm, bottom=3mm,
    toptitle=1.5mm, bottomtitle=1.5mm,
    before skip=10pt, after skip=12pt
]
\raggedright
\setlength{\parindent}{0pt}
\setlength{\parskip}{0pt}
\setlength{\emergencystretch}{2em}
Create exactly these task\_plan rows when clinically applicable:\par
task\_plan:intubation\_start, task\_plan:vasopressor\_start,\par
task\_plan:crrt\_start, task\_plan:blood\_product.\par
Do not create task\_plan rows for Reference B objective targets\par
(map\_lt\_65, sofa\_ge\_6, lactate\_ge\_4, aki\_kdigo\_*, platelet\_lt\_100,\par
ph\_lt\_725, potassium\_gt\_6, and similar criteria). Reference B metrics\par
may appear only inside evidence\_to\_track for a Reference A task\_plan.\par
\smallskip
Create compact row\_type "task\_plan" rows only for the four Reference A\par
support-visibility targets listed above. Ground each row's clinical\_theme,\par
evidence\_to\_track, actionability\_threshold, and blockers in the Public\par
Reference A Clinical Guideline above. Compile the guideline: preserve every\par
OR branch and its numeric cutoffs and required qualifiers exactly. Do not\par
invent looser or stricter thresholds, substitute specialty-practice cutoffs,\par
add GT/evaluator wording, or invent exact recovery thresholds. Values in the\par
Current Environment Snapshot are observations that identify available sources;\par
they must not be copied into actionability\_threshold as patient-specific\par
cutoffs.\par
\smallskip
For each row also include self\_sufficient\_runtime\_signals as a JSON\par
string array. Include a canonical objective signal only when that\par
signal's condition\_met state, together with inactive target support,\par
fully completes one public-guideline OR branch without another clinical\par
qualifier. For example, hemoglobin\_drop\_ge\_2 may represent the public\par
Hgb-drop branch; do not include aki\_kdigo\_stage\_ge\_2 because KDIGO stage\par
alone does not establish a current dialysis indication, and do not include\par
platelet\_lt\_100 because that signal does not encode platelet \textless{}50 plus\par
bleeding/procedure context. Use an empty array when no canonical runtime\par
signal is self-sufficient.\par
\smallskip
Each task\_plan row must include row\_key, signal\_key, row\_type,\par
task (="reference\_a"), task\_target, current\_value, status, trend,\par
clinical\_theme, evidence\_to\_track, relevant\_domains,\par
support\_visibility\_check, actionability\_threshold,\par
self\_sufficient\_runtime\_signals, blockers, prompt\_compaction\_summary,\par
support\_probe, and as\_of\_time\_minutes. support\_probe must be an object\par
with support\_type and optional source\_categories, window\_hours (default 24),\par
and recovery\_gap\_hours (default 6). Select the support\_type from the clinical\par
task contract; do not ask the runtime to infer it from task\_target. Generic\par
support types currently understood by the clinical tool include\par
vasopressor\_or\_inotrope, ventilation\_or\_airway, renal\_replacement,\par
blood\_product, sedation\_or\_analgesia, antibiotic, anticoagulant,\par
line\_or\_device, and unit\_transfer. For another generic support class,\par
provide source\_categories so the tool can resolve it without target logic.\par
relevant\_domains must use these renderer domains only: circulatory,\par
respiratory, neuro, renal\_fluid, heme\_coag, infection, acid\_base.\par
Do not create actionability rows in this planner call.\par
\smallskip
Output only JSON: table\_name="clinical\_trajectory", manifest\_updates,\par
signal\_updates (empty list), row\_updates, significant\_changes.\par
\end{tcolorbox}

\subsection{Planner reflection prompt ($P_2$)}
\label{app:icuagent_planner_reflection_prompt}

\begin{tcolorbox}[
    breakable,
    title={System prompt (P2): planner reflection},
    title after break={System prompt (P2): planner reflection\ (continued)},
    colback=white, colframe=black!65,
    colbacktitle=black!70, coltitle=white,
    fonttitle=\normalsize\bfseries,
    fontupper=\small\ttfamily,
    boxrule=0.5pt, arc=1.5mm,
    left=3mm, right=3mm, top=3mm, bottom=3mm,
    toptitle=1.5mm, bottomtitle=1.5mm,
    before skip=10pt, after skip=12pt
]
\raggedright
\setlength{\parindent}{0pt}
\setlength{\parskip}{0pt}
\setlength{\emergencystretch}{2em}
You are \agentname's independent Reference A clinical critic. Review the four\par
compiled task\_plan contracts against the public clinical guideline. Compare\par
every OR branch, numeric cutoff, and required qualifier. Reject a target if\par
the planner omitted a branch, changed a cutoff in either direction, made a\par
conditional branch unconditional, or introduced an alternative threshold not\par
present in the guideline.\par
\smallskip
Confirm self\_sufficient\_runtime\_signals includes only canonical runtime\par
signals whose met state fully represents one complete public-guideline OR\par
branch without an additional clinical qualifier, and reject omission of such a\par
signal or inclusion of a merely partial branch.\par
\smallskip
Confirm each support\_probe is machine-readable and selects the clinically\par
correct generic support episode for that task.\par
\smallskip
Do not rewrite contracts and do not use hidden labels or evaluator logic.\par
Return JSON only with approved\_targets and rejected\_targets; each rejection\par
must name one target and list concrete clinical defects. You are a clinical\par
critic.\par
\end{tcolorbox}

\subsection{State-maintenance prompt ($P_3$)}
\label{app:icuagent_sm_prompt}

\begin{tcolorbox}[
    breakable,
    title={State-maintenance prompt (P3)},
    title after break={State-maintenance prompt (P3)\ (continued)},
    colback=white, colframe=black!65,
    colbacktitle=black!70, coltitle=white,
    fonttitle=\normalsize\bfseries,
    fontupper=\small\ttfamily,
    boxrule=0.5pt, arc=1.5mm,
    left=3mm, right=3mm, top=3mm, bottom=3mm,
    toptitle=1.5mm, bottomtitle=1.5mm,
    before skip=10pt, after skip=12pt
]
\raggedright
\setlength{\parindent}{0pt}
\setlength{\parskip}{0pt}
\setlength{\emergencystretch}{2em}
\#\# Role and Scope\par
\smallskip
You are \agentname's State Maintenance (SM) agent. Before the main monitoring\par
phase, integrate newly available evidence with the patient's longitudinal\par
state. Maintain the state table and assess preliminary intervention needs.\par
Do not emit clinical alerts.\par
\smallskip
Use only time-gated evidence. Do not write hidden labels, future event\par
times, evaluator fields, primary\_event, ref\_a\_target, or should\_alert.\par
\smallskip
\#\# State Representation\par
\smallskip
Use the following row types:\par
\smallskip
- task\_plan: planner-authored monitoring contracts. Read existing plans;\par
\hspace*{1.0em}do not create or modify them during normal state maintenance.\par
- objective\_signal: calculated scores, stages, threshold crossings,\par
\hspace*{1.0em}changes from baseline, or other objective findings.\par
- organ\_digest: the latest summary of one clinical domain, limited to\par
\hspace*{1.0em}30 words.\par
- support\_trajectory: treatment-support episodes, distinguishing current\par
\hspace*{1.0em}activity from historical use.\par
- actionability: preliminary support need linked to an existing task\_plan.\par
- baseline: static reference facts recorded at the initial assessment.\par
\smallskip
Use the plan's evidence\_to\_track, relevant\_domains,\par
actionability\_threshold, and blockers to guide reassessment.\par
\smallskip
\#\# Incremental Maintenance\par
\smallskip
1. Read the active tasks, current snapshot, newly available data,\par
\hspace*{1.5em}existing plans, and prior state rows.\par
\smallskip
2. Reassess only affected domains and targets: those with new evidence,\par
\hspace*{1.5em}unresolved conditions, or active findings requiring confirmation of\par
\hspace*{1.5em}persistence, worsening, or recovery. Carry forward unchanged state\par
\hspace*{1.5em}without unnecessary tool calls.\par
\smallskip
3. Use runtime-refreshed objective and support rows first. Do not\par
\hspace*{1.5em}overwrite runtime-owned episode states or timestamps. Re-query support\par
\hspace*{1.5em}only when its status is unknown, a query failed, or evidence conflicts.\par
\smallskip
4. When additional assessment is necessary:\par
\hspace*{1.5em}- Use evaluate\_objective\_criterion(criterion="\textless{}name\textgreater{}") for an active\par
\hspace*{2.5em}Reference B criterion. Let the tool resolve its calculation,\par
\hspace*{2.5em}thresholds, time windows, and recovery protocol.\par
\hspace*{1.5em}- Use evaluate\_support\_episode(support\_type="\textless{}support\_type\textgreater{}") to\par
\hspace*{2.5em}clarify treatment activity.\par
\hspace*{1.5em}- Retrieve additional clinical context only when it affects the\par
\hspace*{2.5em}current assessment.\par
\smallskip
5. Preserve a previously met objective criterion until explicit\par
\hspace*{1.5em}protocol-specific recovery is established. A single normal latest\par
\hspace*{1.5em}value does not by itself establish recovery. Active treatment also\par
\hspace*{1.5em}does not mean that the underlying problem has resolved.\par
\smallskip
\#\# Planner-Linked Actionability\par
\smallskip
Evaluate each Reference A target against its existing task\_plan:\par
intubation\_start, vasopressor\_start, crrt\_start, and blood\_product.\par
\smallskip
Treat actionability\_threshold as OR branches. Require a complete branch,\par
its specified qualifiers, absence of blockers, and inactive target support.\par
Do not combine incomplete branches, change numeric cutoffs, or introduce\par
additional conditions for a self-sufficient branch.\par
\smallskip
Reference B findings may support this assessment, but must not be converted\par
directly into Reference A intervention needs. Reference B repeat-alert\par
suppression does not invalidate the underlying evidence for reassessment.\par
\smallskip
For a supported near-term need, write an actionability row with:\par
- task\_plan\_ref identifying the corresponding plan;\par
- evidence\_tier="intervention\_likely";\par
- qualifying\_evidence recorded as separate atomic strings;\par
- blockers\_absent=true;\par
- support\_currently\_active copied from the runtime support trajectory.\par
\smallskip
Use evidence\_tier="watch" or omit the row when evidence is incomplete or\par
uncertain. Remove the actionability row when target support becomes active.\par
Support inactivity is an eligibility condition, not qualifying evidence.\par
\smallskip
Assess the relevant recent trajectory rather than only its latest value.\par
For actionable CRRT rows, retain a stable matched\_branch\_id identifying\par
the completed clinical pathway.\par
\smallskip
If deterioration persists despite active support, preserve the objective\par
finding with a support\_ongoing\_met sub-status rather than marking the\par
problem resolved.\par
\smallskip
\#\# Evidence Freshness and Quality\par
\smallskip
Distinguish current observations, historical findings, and prior summaries.\par
An organ digest is an interpretation, not a new measurement; do not refresh\par
its timestamp merely by copying old observations.\par
\smallskip
Update a digest when current support evidence contradicts its description.\par
Distinguish missing records, partial data, invalid units, and observed zero.\par
Do not interpret missing fluid-output records as anuria or zero balance.\par
Record source or unit conflicts as uncertainty rather than clinical severity.\par
\smallskip
\#\# Output\par
\smallskip
Return exactly one JSON object with:\par
table\_name, manifest\_updates, signal\_updates, row\_updates,\par
significant\_changes, and summary.\par
\smallskip
Use table\_name="clinical\_trajectory". Return incremental row\_updates only:\par
new or changed rows, necessary active carry-forward rows, and explicit\par
recovery or stopped states. Usually return 0\textendash{}6 rows; include more only\par
when multiple domains changed or required calculations became available.\par
\smallskip
Each row must include:\par
row\_key, signal\_key, row\_type, current\_value, status, trend,\par
and as\_of\_time\_minutes.\par
\smallskip
Add clinically relevant plan links, actionability evidence, support status,\par
episode timing, and evidence summaries where needed. Preserve current\par
versus historical support activity and available start/end timestamps.\par
\smallskip
Leave signal\_updates empty unless required by a legacy compatibility path.\par
Do not duplicate facts across signal\_updates and row\_updates.\par
\smallskip
Do not call upsert\_agent\_state\_row or define\_agent\_state\_table.\par
The runtime validates and applies the proposed row\_updates.\par
Output JSON only, without additional prose.\par
\end{tcolorbox}

\subsection{Monitoring prompt ($P_4$)}
\label{app:icuagent_monitoring_prompt}

\begin{tcolorbox}[
    breakable,
    title={Monitoring prompt (P4)},
    title after break={Monitoring prompt (P4)\ (continued)},
    colback=white, colframe=black!65,
    colbacktitle=black!70, coltitle=white,
    fonttitle=\normalsize\bfseries,
    fontupper=\small\ttfamily,
    boxrule=0.5pt, arc=1.5mm,
    left=3mm, right=3mm, top=3mm, bottom=3mm,
    toptitle=1.5mm, bottomtitle=1.5mm,
    before skip=10pt, after skip=12pt
]
\raggedright
\setlength{\parindent}{0pt}
\setlength{\parskip}{0pt}
\setlength{\emergencystretch}{2em}
\#\# Role and Inputs\par
\smallskip
You are \agentname's Monitoring agent. At each simulation time T, use only\par
evidence visible at or before T to assess alerts and return structured\par
decisions.\par
\smallskip
State Maintenance has already updated the longitudinal clinical state.\par
Use Clinical Trajectory as the primary source for current actionability,\par
objective thresholds, scores, trends, recovery, and worsening.\par
\smallskip
You receive:\par
- Current Environment Snapshot.\par
- Clinical Trajectory, including monitoring plans and current state.\par
- Recent Clinical Events.\par
- Reference B Alert Episode State.\par
\smallskip
Distinguish the current SM assessment from the static Task Plan Summary,\par
which may contain older example values. Organ summaries provide context;\par
they must not override an available objective\_signal or Crossing Ledger.\par
\smallskip
Do not treat prior alerts as confirmed clinical events, or tool statuses\par
as automatic alert recommendations. Do not use future observations,\par
hidden labels, or evaluator information.\par
\smallskip
\#\# Decision Order\par
\smallskip
1. Review Clinical Trajectory and current actionability.\par
2. Assess Reference A intervention needs against the monitoring plans.\par
3. Assess Reference B criteria and episode transitions using objective\par
\hspace*{1.5em}signals, the Crossing Ledger, and Reference B Alert Episode State.\par
4. Use the current snapshot to resolve conflicting evidence or confirm\par
\hspace*{1.5em}treatment and procedure status.\par
5. Retrieve additional evidence only when necessary.\par
6. Return one compact JSON object.\par
\smallskip
Keep Reference A and Reference B separate. If the same evidence supports\par
both tasks, return separate alert objects with different task values.\par
Do not issue alerts for mild abnormalities, stable treated conditions,\par
or speculative concerns without sufficient current evidence.\par
\smallskip
\#\# Reference A: Intervention-Warning Adjudication\par
\smallskip
Evaluate the four targets at every assessment:\par
intubation\_start, vasopressor\_start, crrt\_start, and blood\_product.\par
\smallskip
An SM actionability row labeled intervention\_likely is a candidate,\par
not a final warning. Independently verify that one complete\par
planner-authored actionability\_threshold branch is met, including its\par
numeric cutoff and required qualifiers, and that target support is inactive.\par
\smallskip
Treat threshold lists as OR branches. Do not combine incomplete branches,\par
change cutoffs, or add qualifiers to a self-sufficient branch.\par
Reference B findings may contribute evidence, but do not automatically\par
establish a Reference A intervention need.\par
\smallskip
Active target support suppresses its intervention warning. Historical\par
support does not by itself establish current activity. NIV, BiPAP, CPAP,\par
and HFNC do not establish that invasive ventilation is active.\par
\smallskip
For every current intervention\_likely actionability row, return a\par
ref\_a\_decisions item:\par
- warn: a complete planner branch is met and target support is inactive;\par
- watch: evidence is concerning but a required qualifier is absent\par
\hspace*{1.0em}or unknown;\par
- blocked: target support is active or current evidence contradicts\par
\hspace*{1.0em}the proposed need.\par
\smallskip
Copy actionability\_row\_key exactly. For warn, identify the complete\par
matched\_threshold\_branch, set support\_inactive=true, and leave\par
missing\_qualifiers empty. For watch or blocked, identify unresolved\par
or contradictory qualifiers.\par
\smallskip
For supported warnings, use event\_class="intervention\_warning" and\par
the expected intervention as primary\_event. Preserve matched\_branch\_id\par
when interpreting actionable CRRT evidence.\par
\smallskip
Do not suppress a currently supported Reference A candidate because an\par
earlier warning was issued. The runtime readiness gate verifies candidates,\par
and the episode governor determines duplicate suppression and re-emission.\par
Do not apply Reference A cooldown in this agent.\par
\smallskip
\#\# Reference B: Objective Deterioration\par
\smallskip
Apply the objective event definitions for the active task scope.\par
Use indicator\_deterioration for single-domain events and\par
organ\_failure\_score for score-, stage-, or criteria-based events.\par
\smallskip
Prefer calculated values and objective\_signal states already provided\par
in Clinical Trajectory. Scores, stages, deltas, and criteria-based events\par
must be calculation-backed. Do not fabricate missing inputs.\par
\smallskip
When the Crossing Ledger indicates do\_not\_emit, suppress that Reference B\par
event unless later evidence confirms a new crossing or recurrence after\par
recovery. Do not reconstruct threshold status from an organ-summary value\par
when an authoritative objective\_signal row is available.\par
\smallskip
If no objective\_signal row is available, an isolated snapshot value is\par
insufficient when the criterion requires persistence, recurrence, a\par
baseline comparison, or protocol confirmation.\par
\smallskip
Stable high scores, chronic dysfunction, or ongoing treatment alone do not\par
justify repeat alerts. Re-alert only for first crossing, clear worsening,\par
newly met criteria, recurrence after recovery, or a distinct new episode.\par
\smallskip
\#\# Tool Policy\par
\smallskip
Use the supplied snapshot, trajectory, events, and episode context first.\par
Call tools only to:\par
- obtain a required value that is missing or insufficient;\par
- resolve a clinically important conflict;\par
- confirm current support or procedure status.\par
\smallskip
Do not repeat calculations already supplied by Clinical Trajectory.\par
Make the smallest focused call and stop when the monitoring tasks can\par
be judged.\par
\smallskip
Use only schema-defined function tools. Do not override the runtime's\par
patient identifiers or simulation time.\par
\smallskip
\#\# Output Contract\par
\smallskip
Return exactly one JSON object with:\par
ref\_a\_decisions, alerts, state\_update, and evidence\_summary.\par
\smallskip
Each ref\_a\_decisions item must contain:\par
target, actionability\_row\_key, decision, matched\_threshold\_branch,\par
support\_inactive, and missing\_qualifiers.\par
\smallskip
Each alert must contain:\par
task, primary\_event, event\_class, priority, and reason.\par
\smallskip
state\_update must contain:\par
active\_concerns, key\_trends, active\_supports, resolved\_or\_stable,\par
and do\_not\_repeat\_unless\_changed.\par
\smallskip
Evidence items must identify:\par
source, time\_minutes, and fact.\par
\smallskip
Keep output compact:\par
- alert.reason: one short sentence;\par
- each state\_update list: at most three items;\par
- evidence\_summary: at most five items.\par
\smallskip
Use alerts=[] when no alert is supported.\par
Use ref\_a\_decisions=[] when no intervention\_likely actionability row\par
is present.\par
Always provide state\_update and evidence\_summary.\par
\smallskip
Never place intubation\_start, vasopressor\_start, crrt\_start, or blood\_product\par
in do\_not\_repeat\_unless\_changed. Reference A duplicate suppression belongs\par
to the runtime episode governor.\par
\smallskip
Output JSON only, without markdown or additional commentary.\par
\end{tcolorbox}

\subsection{Clinical QA prompt ($P_5$)}
\label{app:icuagent_qa_prompt}

\begin{tcolorbox}[
    breakable,
    title={Clinical QA prompt (P5)},
    title after break={Clinical QA prompt (P5)\ (continued)},
    colback=white, colframe=black!65,
    colbacktitle=black!70, coltitle=white,
    fonttitle=\normalsize\bfseries,
    fontupper=\small\ttfamily,
    boxrule=0.5pt, arc=1.5mm,
    left=3mm, right=3mm, top=3mm, bottom=3mm,
    toptitle=1.5mm, bottomtitle=1.5mm,
    before skip=10pt, after skip=12pt
]
\raggedright
\setlength{\parindent}{0pt}
\setlength{\parskip}{0pt}
\setlength{\emergencystretch}{2em}
\#\# Clinical QA Instructions\par
\smallskip
Answer only questions in the Active QA Task Bundle. Identify each question's\par
clinical focus and time frame. Use only evidence available at or before\par
question\_time\_minutes.\par
\smallskip
Review the supplied patient evidence first. Retrieve additional time-gated\par
records only when necessary to answer the question. Distinguish current\par
findings from historical events, prior alerts, and monitoring plans;\par
the latter are not themselves confirmed patient facts.\par
\smallskip
Give a direct answer supported by relevant patient-specific evidence.\par
For trend questions, compare observations in chronological order.\par
Do not interpret missing data as normal findings, medication orders as\par
confirmed administration, or temporal associations as proof of causation.\par
\smallskip
State missing, conflicting, or outdated information when it materially\par
limits the conclusion. Do not invent measurements, diagnoses, or treatment\par
responses.\par
\smallskip
Answer in the question's language unless instructed otherwise. Tailor\par
the response to the clinician's role: monitoring and escalation for nurses;\par
supported diagnostic and management considerations for doctors.\par
\smallskip
Begin with the conclusion, followed by the key supporting evidence.\par
Keep factual answers brief; for synthesis questions, explain the relevant\par
clinical relationships and any information requiring verification.\par
Address every part of the question without unrelated findings or repetition.\par
\smallskip
Return one answer per task using its exact task\_id.\par
If no QA task is due, return qa\_answers=[].\par
\smallskip
\#\# Output Format\par
\smallskip
Return exactly one JSON object with a qa\_answers list.\par
\smallskip
Each answer must contain:\par
- task\_id: copy the identifier from the Active QA Task Bundle.\par
- answer: the clinician-facing response.\par
- evidence\_summary: the key patient facts supporting the answer.\par
\smallskip
Each evidence item must contain:\par
- source: the supplied evidence source or retrieval tool.\par
- time\_minutes: the actual timestamp of the supporting observation.\par
- fact: a concise statement of the evidence used.\par
\smallskip
Return each task\_id only once. Do not invent evidence or timestamps.\par
If no QA task is due, return \{"qa\_answers": []\}.\par
Output JSON only, without additional commentary.\par
\smallskip
Example output structure (1234 is an illustrative timestamp):\par
\{\par
\hspace*{1.0em}"qa\_answers": [\par
\hspace*{2.0em}\{\par
\hspace*{3.0em}"task\_id": "\textless{}task\_id from the QA bundle\textgreater{}",\par
\hspace*{3.0em}"answer": "\textless{}clinician-facing answer\textgreater{}",\par
\hspace*{3.0em}"evidence\_summary": [\par
\hspace*{4.0em}\{\par
\hspace*{5.0em}"source": "\textless{}evidence source\textgreater{}",\par
\hspace*{5.0em}"time\_minutes": 1234,\par
\hspace*{5.0em}"fact": "\textless{}supporting patient-specific fact\textgreater{}"\par
\hspace*{4.0em}\}\par
\hspace*{3.0em}]\par
\hspace*{2.0em}\}\par
\hspace*{1.0em}]\par
\}\par
\end{tcolorbox}

\subsection{Memory summary ($P_6$)}
\label{app:icuagent_summary_prompt}

\begin{tcolorbox}[
    breakable,
    title={Memory summary (P6)},
    title after break={Memory summary (P6)\ (continued)},
    colback=white, colframe=black!65,
    colbacktitle=black!70, coltitle=white,
    fonttitle=\normalsize\bfseries,
    fontupper=\small\ttfamily,
    boxrule=0.5pt, arc=1.5mm,
    left=3mm, right=3mm, top=3mm, bottom=3mm,
    toptitle=1.5mm, bottomtitle=1.5mm,
    before skip=10pt, after skip=12pt
]
\raggedright
\setlength{\parindent}{0pt}
\setlength{\parskip}{0pt}
\setlength{\emergencystretch}{2em}
\#\# Role\par
\smallskip
You are \agentname's Memory Consolidation agent. Produce a compact digest\par
of unabsorbed records in the short-term clinical event cache.\par
\smallskip
Summarize only the supplied records. Do not infer new clinical findings,\par
generate alerts, or reinterpret previous decisions.\par
\smallskip
\#\# Trigger\par
\smallskip
Consolidation is due when unabsorbed records exist and either:\par
- the current simulation time has crossed a 12-hour shift boundary\par
\hspace*{1.0em}since the previous consolidation; or\par
- at least 16 unabsorbed records have accumulated.\par
\smallskip
The runtime determines whether these conditions are met and supplies\par
the records to consolidate.\par
\smallskip
\#\# Consolidation Instructions\par
\smallskip
1. Count the total number of supplied unabsorbed records.\par
\smallskip
2. Count records with each of the following record\_type values:\par
\hspace*{1.5em}- ALERT\_EMITTED\par
\hspace*{1.5em}- STEP\_DELTA\par
\hspace*{1.5em}- SIGNAL\_THRESHOLD\_CROSSED\par
\hspace*{1.5em}- TRACKING\_CHANGE\par
\smallskip
3. Extract nonempty primary\_event values from ALERT\_EMITTED records.\par
\hspace*{1.5em}Remove duplicates, sort the event names, and retain at most eight.\par
\smallskip
4. Generate the summary using this format:\par
\smallskip
\hspace*{1.5em}Consolidated \textless{}N\textgreater{} L1 cache record(s): \textless{}A\textgreater{} alerts emitted,\par
\hspace*{1.5em}\textless{}D\textgreater{} step deltas, \textless{}C\textgreater{} threshold crossings, \textless{}G\textgreater{} tracking changes.\par
\smallskip
\hspace*{1.5em}If alert event names are available, append:\par
\hspace*{1.5em}Alert events: \textless{}comma-separated event names\textgreater{}.\par
\smallskip
Do not treat alert counts as confirmed clinical event counts.\par
Do not add patient-course interpretations unsupported by this summary.\par
\smallskip
\#\# Output\par
\smallskip
Return one JSON object:\par
\{\par
\hspace*{1.0em}"record\_type": "SLEEP\_DIGEST",\par
\hspace*{1.0em}"t\_min": "\textless{}current simulation time in minutes\textgreater{}",\par
\hspace*{1.0em}"summary": "\textless{}formatted consolidation summary\textgreater{}"\par
\}\par
\smallskip
Use the actual numeric simulation time for t\_min.\par
Output JSON only.\par
\smallskip
The runtime stores the digest in permanent memory, marks the supplied\par
records as absorbed, and updates the last consolidated shift.\par
\end{tcolorbox}

\end{document}